\documentclass[letterpaper]{article} 
\usepackage[submission]{aaai2026}  
\usepackage{times}  
\usepackage{helvet}  
\usepackage{courier}  
\usepackage[hyphens]{url}  
\usepackage{graphicx} 
\usepackage{natbib}  
\usepackage{caption} 
\usepackage{multirow}
\usepackage{multicol}
\usepackage{algorithm}
\usepackage{algorithmic}

\usepackage{listings}
\usepackage{multirow}
\usepackage{booktabs}
\usepackage{graphicx}
\usepackage{amsmath} 
\usepackage{booktabs}
\usepackage{amssymb}
\usepackage[switch]{lineno}
\usepackage{multirow}
\usepackage{array}
\usepackage{bbding}
\usepackage{longtable}
\usepackage{bibentry}
\usepackage{tcolorbox}
\usepackage{cleveref}
\usepackage{newfloat}
\usepackage{listings}

\usepackage{graphicx}    
\usepackage{fontawesome5}
\usepackage{amsmath}
\usepackage{xcolor} 
\usepackage{twemojis}

\definecolor{softgreen}{HTML}{8FBC8F}
\definecolor{softred}{HTML}{E9967A}

\usepackage{graphicx}   
\usepackage{fontawesome5}
\usepackage{amsmath}
\usepackage{xcolor} 
\usepackage{fontawesome5}
\usepackage{twemojis}

\usepackage{tcolorbox}
\usepackage{cleveref}
\tcbuselibrary{listings, breakable}

\definecolor{softgreen}{HTML}{8FBC8F}
\definecolor{softred}{HTML}{E9967A}

\newcommand{\mirage}{{\textbf{\textsc{Mirage}}}}

\newcommand{\model}{{\textbf{\textsc{Mirage-R1}}}}

\newcommand{\miragetitle}{\scalebox{2.5}{\twemoji{desert}} \textsf{MIRAGE}}

\DeclareCaptionStyle{ruled}{labelfont=normalfont,labelsep=colon,strut=off} 
\floatstyle{ruled}
\newfloat{listing}{tb}{lst}{}
\floatname{listing}{Listing}
\title{\miragetitle: Towards AI-Generated Image Detection in the Wild}

\author{
    Cheng Xia\textsuperscript{\rm 1}\equalcontrib, Manxi Lin\textsuperscript{\rm 1}\equalcontrib, Jiexiang Tan\textsuperscript{\rm 1}\equalcontrib, Xiaoxiong Du\textsuperscript{\rm 1}, Yang Qiu\textsuperscript{\rm 1}\\ 
    Junjun Zheng\textsuperscript{\rm 1}\thanks{Co-corresponding authors.}, Xiangheng Kong\textsuperscript{\rm 1}\footnotemark[2], Yuning Jiang\textsuperscript{\rm 1}, Bo Zheng\textsuperscript{\rm 1}
}
\affiliations{
    \textsuperscript{\rm 1}Algorithm Tech Team Taobao \& Tmall Group of Alibaba\\
}

\begin{document}

\maketitle

\begin{abstract}
The spreading of AI-generated images (AIGI), driven by advances in generative AI, poses a significant threat to information security and public trust. Existing AIGI detectors, while effective against images in clean laboratory settings, fail to generalize to \textbf{in-the-wild} scenarios. These real-world images are noisy, varying from ``obviously fake" images to realistic ones derived from multiple generative models and further edited for quality control. We address \textbf{in-the-wild AIGI detection} in this paper. We introduce \mirage, a challenging benchmark designed to emulate the complexity of in-the-wild AIGI. {\mirage} is constructed from two sources: (1) a large corpus of Internet-sourced AIGI verified by human experts, and (2) a synthesized dataset created through the collaboration between multiple expert generators, closely simulating the realistic AIGI in the wild. Building on this benchmark, we propose \model, a vision-language model with heuristic-to-analytic reasoning, a reflective reasoning mechanism for AIGI detection. {\model} is trained in two stages: a supervised-fine-tuning cold start, followed by a reinforcement learning stage. By further adopting an inference-time adaptive thinking strategy, {\model} is able to provide either a quick judgment or a more robust and accurate conclusion, effectively balancing inference speed and performance. Extensive experiments show that our model leads state-of-the-art detectors by $\textbf{5\%}$ and $\textbf{10\%}$ on {\mirage} and public benchmark, respectively. The benchmark and code will be made publicly available.
\end{abstract}

\section{Introduction}
The rapid development of generative AI has made the creation of AI-generated images (AIGI) widely accessible~\cite{rombach2022high,cai2025hidream}. While this technology benefits creative industry, it also raises serious concerns. On one hand, photorealistic forgeries can lead to problems like copyright issues~\cite{rombach2022high,ren2024copyright}; on the other hand, the mass spread of obviously-fake AI images can flood digital platforms, harming user experience and authenticity of the platform~\cite{yin2024enabling}. In response, recent works have developed AIGI detectors~\cite{xu2024fakeshield,huang2025so,gao2025fakereasoning} that shows impressive performance in separating real photos from even the most convincing synthetic ones.

In long-established computer vision tasks like object detection~\cite{chen2018domain,xiang2014beyond} and face recognition~\cite{liu2015deep}, ``in-the-wild" evaluation is standard to measure real-world performance. In this paper, we address the \textbf{in-the-wild AIGI detection} task. Inspired by similar tasks in other vision fields, we define this problem as identifying \textit{naturally-occurring} AI-generated images: those that have already circulated online or been altered in a real-world setting.

\begin{figure*}[ht]
    \centering
    \includegraphics[width=0.9\linewidth]{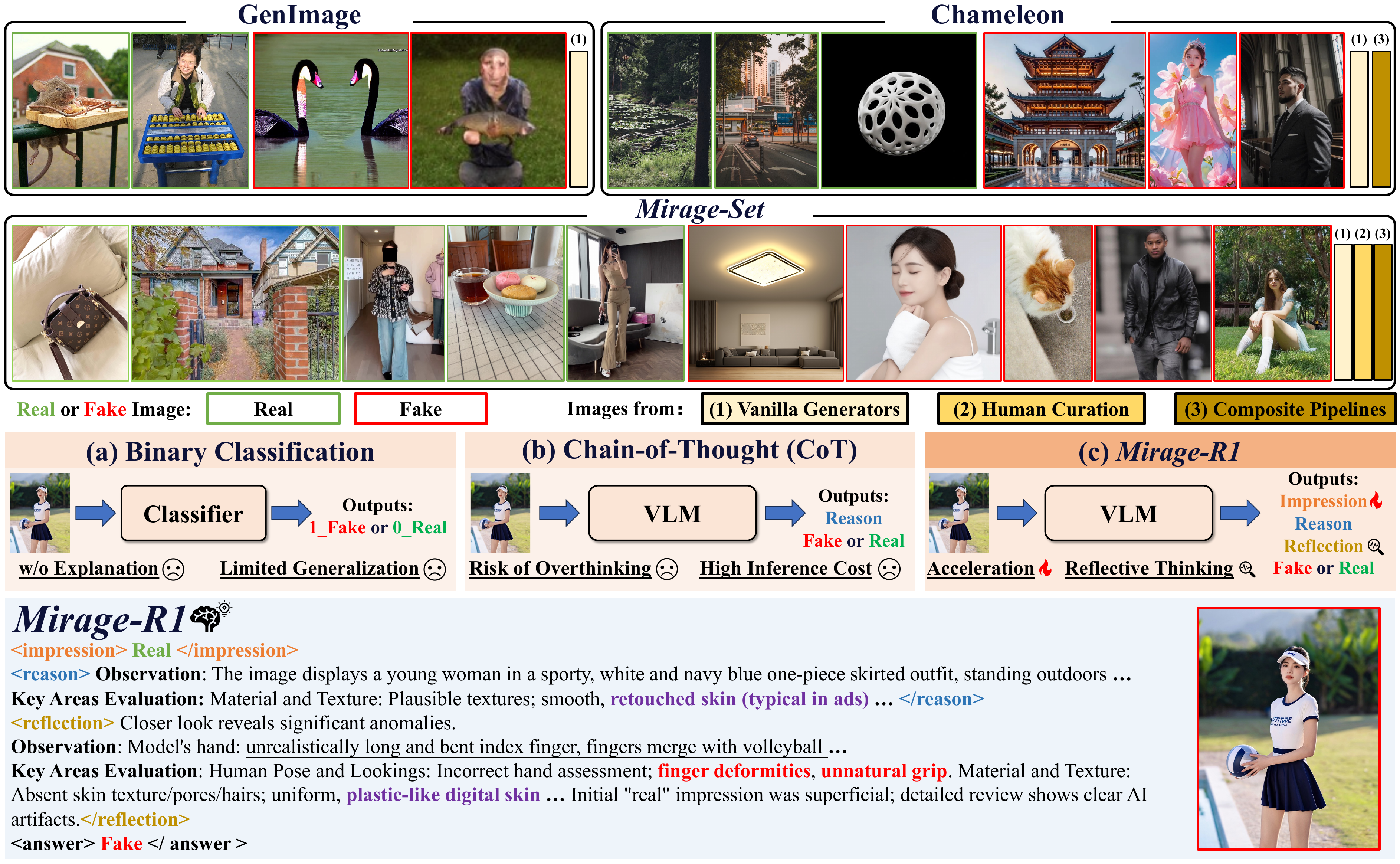}
    \caption{Overall review of our {\mirage} and \model.}
    \label{fig:teaser}
\end{figure*}

\noindent\textbf{Benchmark.}
Most existing benchmarks~\cite{cnn-spot,zhu2023genimage} confine AIGI to images from certain types of generative models in a controlled environment, without capturing the complexity and messiness of real-life scenarios. In contrast, recent studies~\cite{yan2024sanity,zhang2025ivy} have emphasized the importance of using ``real-world" fake images sourced from online AIGI communities, which are often realistic enough to deceive human experts~\cite{yan2024sanity}. However, the ``in-the-wild" scenario is not limited to these two types of images. It also encompasses a large volume of fake images that, while potentially recognizable by humans, still exhibit a data distribution distinct from that of single-model outputs due to real-world factors like post-processing and quality control.

To address these gaps, we introduce {\mirage} — the first dataset focused on in-the-wild AIGI detection. As shown in Fig.~\ref{fig:teaser}, besides the images from the controlled environment, i.e., \textit{vanilla generators} without further editing, our dataset also includes two types of positive examples: (1) \textit{Human Curation:} We scrawled and annotated images, typically fake images, from different sources on the Internet. Many of these images clearly show signs of AI generation and are ``obviously fake" to most viewers. (2) \textit{AIGI from Composite Pipelines:} We also generate photorealistic AIGI by applying multiple pipelines, involving the collaboration between many models, as well as a range of post-processing steps (such as face restoration). These images serve as more challenging synthetic examples, aiming to capture the variety seen in uncontrolled, real-world settings.

\noindent\textbf{Method.} Recent works have revealed two critical shortcomings of existing AIGI detectors: limited generalization and a lack of explainability~\cite{gao2025fakereasoning}. Prior works~\cite{zhang2025ivy} suggest that Vision-Language Models (VLMs), particularly those employing Chain-of-Thought (CoT) reasoning, hold promise in addressing these challenges due to their strong prior knowledge from large-scale pretraining. In CoT, complex problems are broken down into well-defined steps, such as image captioning and criterion-based judgment, which the VLM completes sequentially. The final decision is made after this explicit reasoning process, and the reasoning process serves as an explanation for the answer. However, CoT-based approaches usually increase inference costs and can be prone to ``overthinking": on ``easy" AIGI samples with obvious flaws, the model may make unnecessary mistakes during the long reasoning process. This issue is especially prominent in noisy, in-the-wild AIGI data, which often contains such ``obviously fake" images.

To address these challenges, we introduce {\model}, a VLM specifically tailored for robust AIGI detection in the wild, building upon {\mirage}. Fig.~\ref{fig:teaser} shows an example of our model reasoning. {\model} is progressively trained in two stages to think in a way, namely heuristic to analytic reasoning. As a consequence, the model is able to generate both a rapid initial prediction (\verb|<impression>| in the figure) and a more deliberate final answer (\verb|<answer>|), inspecting the first one. As shown in the figure, our model is able to perform reflection (\verb|<reflection>|), that is, correcting the issues in its initial reasoning ((\verb|<reason>|)). In inference, {\model} adaptively selects its response according to its confidence in the fast answer. This allows {\model} to efficiently balance accuracy and computational cost in the noisy in-the-wild scenarios. 

Our main contributions are as follows:
\begin{itemize}
\item We define the task of in-the-wild AIGI detection and introduce {\mirage}, a comprehensive benchmark for real-world evaluation.
\item We propose {\model}, a VLM capable of adaptive and reflective reasoning for AIGI detection.
\item Extensive experiments demonstrate that {\model} significantly outperforms existing methods on both the {\mirage} benchmark and other public datasets.
\end{itemize}


\begin{figure*}[ht]
    \centering
    \includegraphics[width=0.95\linewidth]{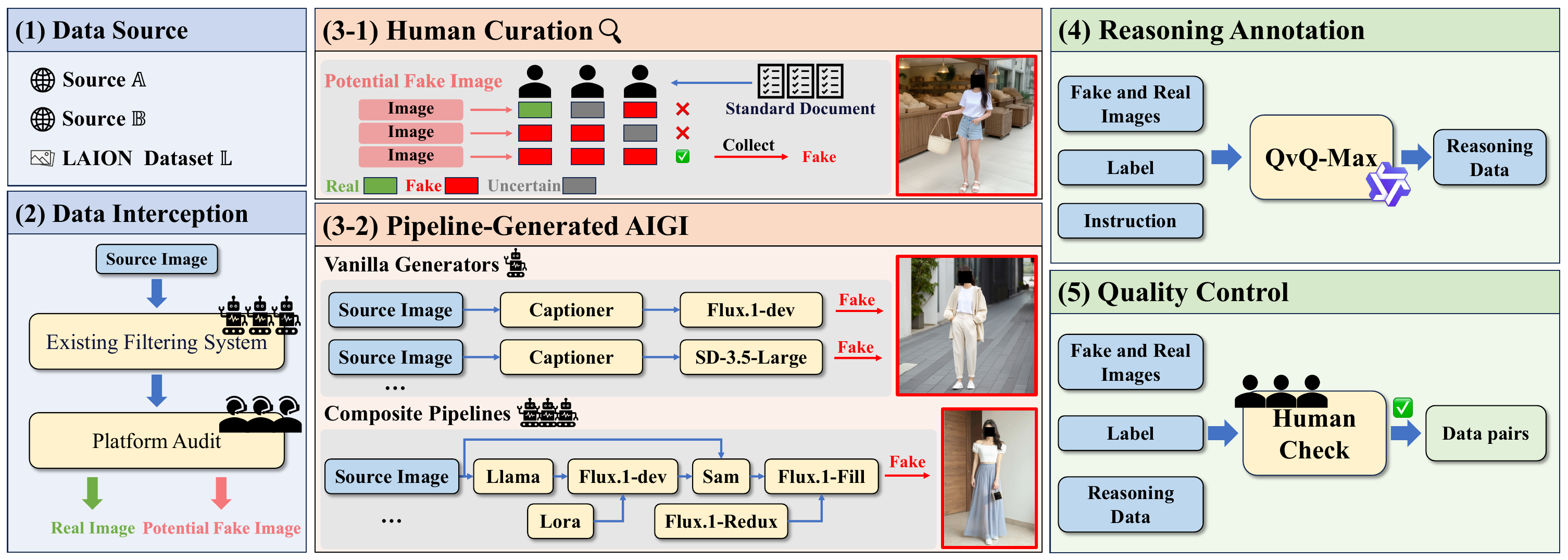}
    \caption{Illustration of the dataset construction of our \mirage. We have anonymized the faces. }
    \label{fig:data}
\end{figure*}
\section{Related Works}
Due to the page limits, we put a detailed related works section in the appendix.

\noindent\textbf{Benchmarks in AIGI Detection.}
Existing AIGI detection benchmarks~\cite{cnn-spot,zhu2023genimage,hong2024wildfake} often fail to model real-world complexity. Even recent benchmarks that simulate real-world AIGI~\cite{yan2024sanity, zhang2025ivy} suffer from two critical limitations: 1) domain bias, introduced by sourcing real and AI-generated images from disparate online contexts, and 2) a lack of examples from complex pipelines involving fine-tuned models and post-processing, which is common in in-the-wild AIGI. 

To bridge these gaps, we introduce {\mirage}, a benchmark designed for more rigorous generalization assessment. It addresses these issues by including: (a) real and AI-generated images curated from the same online communities to eliminate domain bias, and (b) challenging forgeries created through advanced generation and post-processing pipelines, thus providing a more realistic evaluation.

\noindent\textbf{AIGI Detection.}
AIGI detection has evolved from early, low-level artifact detectors~\cite{cnn-spot, npr} to modern approaches leveraging large pre-trained models like CLIP~\cite{univfd, effort,yan2024sanity} for improved generalization. However, these advanced methods suffer from two critical limitations: 1) they are often ``black-box" models, lacking \textit{explainability}~\cite{gao2025fakereasoning}, and 2) their \textit{generalization} is rarely tested on ``in-the-wild" images that undergo unknown post-processing. To address this, recent work has shifted towards VLMs, yet these have introduced their own fundamental trade-off. They are split between fast but potentially unfaithful \textit{heuristic (answer-first)} reasoning~\cite{li2024fakebench, wen2025spot}, and robust but slow \textit{analytic (reason-first)} reasoning~\cite{gao2025fakereasoning, zhang2025ivy}.

To address these challenges, we introduce {\model}, a VLM that pioneers adaptive and reflective reasoning. It first provides a fast, heuristic answer, then adaptively decides whether to engage in deeper, reflective thinking by estimating its confidence in the fast answer. This allows {\model} to achieve generalizable and explainable detection while dynamically allocating computational resources based on task difficulty, offering a unified solution to in-the-wild AIGI detection.

\section{MIRAGE Dataset}
\begin{table}[t]  
    \centering 
    \resizebox{\columnwidth}{!}{\begin{tabular}{c|c|c|c}
    \toprule
    
    \textbf{Dataset} & Vanilla Generators & Human Curation & Composite Pipelines\\ 
    
    \midrule
    CNN-Detection & \textcolor{softgreen}{\faCheckSquare} & \textcolor{softred}{\faWindowClose} &\textcolor{softred}{\faWindowClose}\\
    GenImage& \textcolor{softgreen}{\faCheckSquare} & \textcolor{softred}{\faWindowClose} &\textcolor{softred}{\faWindowClose}\\
    LOKI& \textcolor{softgreen}{\faCheckSquare} & \textcolor{softred}{\faWindowClose} &\textcolor{softred}{\faWindowClose}\\
    Chameleon& \textcolor{softgreen}{\faCheckSquare} & \textcolor{softred}{\faWindowClose} &\textcolor{softgreen}{\faCheckSquare}\\

    \midrule
    \textbf{\mirage}&\textcolor{softgreen}{\faCheckSquare}&\textcolor{softgreen}{\faCheckSquare}&\textcolor{softgreen}{\faCheckSquare}\\         
    
    \bottomrule
    
    \end{tabular}}
    \caption{Data distribution across various benchmarks. Unlike existing benchmarks, our {\mirage} dataset is distinguished by its inclusion of AIGI sourced from different provenances. The columns correspond to these provenances: direct generation from vanilla generators without further post-processing, human-curated examples sourced online, and complex composite pipelines.}
    \label{table:benchmark}
    
\end{table}

\noindent\textbf{Problem Formulation. }
We define in-the-wild AIGI to be naturally-occurring AI-generated images. Compared to conventional benchmark data, in-the-wild AIGI have unique characteristics: (1) \textit{Mixed Presence with Real Photos:} In-the-wild AIGI appear in the same settings where real photos are expected, such as social media and daily news, instead of AI art communities. (2) \textit{Generation from Pipeline:} These images often come from a pipeline: a combination of diverse fine-tuned generative models and potential further manual modification through, e.g., editing, compositing, or retouching for use in social media, advertisements etc. (3) \textit{Wide Realism Range:} Real-world AIGI can span a wide range in realism, from obviously synthetic or heavily edited to highly convincing forgeries that are difficult for humans to identify.

Based on these features, we build the dataset of \textbf{M}ixed \textbf{I}mages from \textbf{R}eal-world \textbf{A}I \textbf{Ge}neration (\mirage). As shown in Tab.~\ref{table:benchmark}, our {\mirage} has the most complete coverage of in-the-wild AIGI compared to existing benchmarks.

\begin{figure*}[ht]
    \centering
    \includegraphics[width=\linewidth]{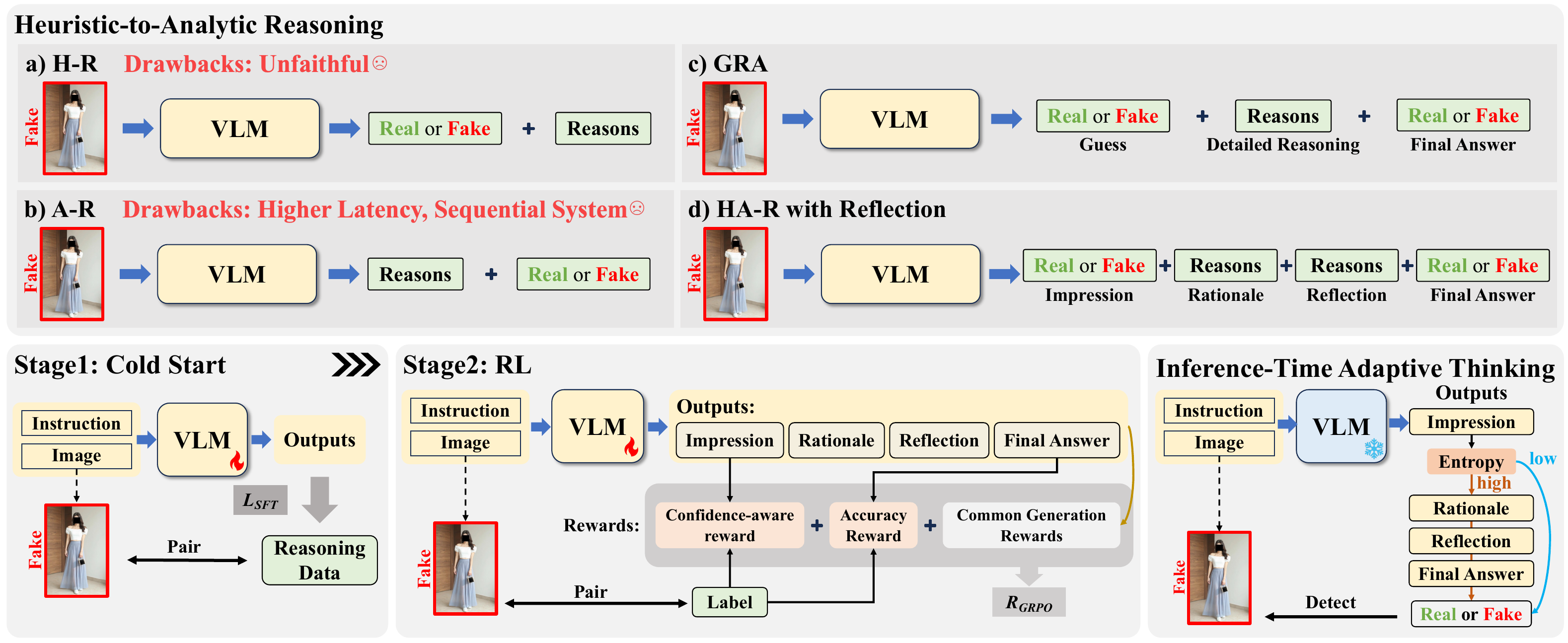}
    \caption{Training framework of our \model.}
    \label{fig:model}
\end{figure*}

\subsection{Dataset Construction. }
The construction of {\mirage} is a multi-stage process encompassing data sourcing, expert annotation of real-world AIGI, advanced pipeline-based generation, and rigorous quality control. An overview is illustrated in Fig.~\ref{fig:data}.

\noindent\textbf{Data Sourcing and Human Curation. }
We sourced our initial data from three diverse origins: two large-scale proprietary websites, $\mathbb{A}$ and $\mathbb{B}$~\footnote{To preserve anonymity, the sources are not named but represent major online platforms with a mix of user-generated content.}, and the public LAION DATASET $\mathbb{L}$~\cite{Laion}. This collection contains a mix of real and potentially AI-generated images from scenarios such as social media, e-commerce, news, etc. A portion of this data, totaling 15,000 images with preliminary ``real" or ``AI-generated" flags from the source platforms, underwent a rigorous expert re-annotation process. We employed 29 trained annotators, with each image independently assessed by three different experts. They could label an image as ``Real" ``AI-generated", or ``Uncertain". To ensure the highest data quality, we applied a strict filtering protocol: (1) Only images where all three annotators and the original platform label reached a unanimous decision were kept; (2) Images unanimously labeled as ``Uncertain" were discarded.
This process yielded 11,559 high-confidence images (8,465 Real, 3,094 AI-generated), forming our human-curated dataset.

\noindent\textbf{AIGI Generation via Vanilla Generators and Composed Pipelines. }
Human annotation can only capture existing, identifiable forgeries. To cover emerging threats and complex manipulations, we identified 8 common real-world generation patterns: Text-to-Image (\textit{T2I}), Inpainting/Outpainting (\textit{IP\&OP}), Instruction-based Editing (\textit{IE}), Face Swapping (\textit{FW}), Background Change (\textit{CB}), Virtual Try-on (\textit{VTO}), Realistic Model Generation (\textit{RMG}), and Pose-Consistent Model Generation (\textit{PCMG}). Here, \textit{T2I} refers to the AIGI construction from vanilla generators (similar to most existing benchmarks), and the rest 7 patterns are all complicated composite pipelines that simulate the real-world application. 

Using ComfyUI~\footnote{\url{https://github.com/comfyanonymous/ComfyUI}} and the Python interpreter, we built 64 automated pipelines integrating 53 different models (including text-to-image generators, LoRAs~\cite{hu2022lora}, ControlNets~\cite{zhang2023adding}, and other image processing models) to realize these generation patterns. For example, our \textit{PCMG} pipeline first uses a VLM to generate a detailed prompt from a real photograph while extracting the subject's pose. Then, a combination of a fine-tuned LoRA, a T2I model, and ControlNet generates a new character in the same pose. As a final step, segmentation models (SAM~\cite{kirillov2023segment} and GroundingDINO~\cite{liu2024grounding}) and inpainting are used to transfer the original clothing onto the generated character, creating a highly convincing forgery. Further details are available in the Appendix.

\noindent\textbf{Reasoning Annotation and Quality Control. }
To train our VLM, we generated textual reasoning for each image using QvQ-Max~\cite{QwenBlog2024QVQMax}, providing it with the ground-truth label and an in-context example, similar to~\cite{zhang2025ivy, huang2025so}. Prompts are given in the appendix. 

Finally, both curated and generated images underwent a final quality control step. Five AIGI experts filtered out noisy or low-quality examples, ensuring that the final benchmark contains only challenging and representative samples.

\subsection{Dataset Partition}
The final {\mirage} benchmark is partitioned into two primary splits for comprehensive evaluation: In-Distribution (ID) and Out-of-Distribution (OOD).

\noindent\textbf{ID Split. }
The ID split is designed for training and standard ID evaluation. It combines human-curated and T2I images from sources $\mathbb{A}$ and $\mathbb{L}$, maintaining a 1:1 real-to-fake ratio. This split contains 21,685 images, divided into a training set (20,000 images) and an ID test set (1,685 images).

\noindent\textbf{OOD Split. }
The OOD split is crafted to rigorously test model generalization against unseen data sources. It contains 12,087 images and is composed of images from the unseen source $\mathbb{B}$ and images generated with the full set of 8 generation patterns from all sources. This split is further divided into multiple test subsets for granular analysis. This includes an OOD human-curated set to test generalization to new real image distributions. Furthermore, each of the 8 generation patterns forms an OOD pipeline-generated test set, composed of the real images from $\mathbb{B}$, and the generated images from the real images from $\mathbb{B}$, across all the 8 generation types. This allows for a fine-grained evaluation of a model's robustness to each manipulation type. Note that the T2I models for this split are different from the ones used in the ID split to keep the generated images OOD.

\section{Methodology}
The detection of AIGI in the wild demands a delicate balance of: (1) \textit{Generalization:} As generative models rapidly evolve, detectors must perform reliably on OOD images from unseen models and diverse scenarios. (2) \textit{Explainability:} In sensitive applications like content moderation, providing clear, human-understandable rationales for decisions is crucial for user trust. (3) \textit{Efficiency:} Many practical scenarios, such as real-time media analysis, impose strict low-latency requirements.

To address these challenges, we introduce {\model}, a VLM architected for in-the-wild AIGI detection. As illustrated in Fig.~\ref{fig:model}, our core innovation is a novel reasoning framework, Heuristic-to-Analytic Reasoning, which emulates the multi-stage cognitive process of human experts. We operationalize this framework through a two-stage progressive training paradigm. The resulting {\model} is uniquely capable of delivering a fast initial verdict for efficiency, a detailed rationale for explainability, and a final, more accurate answer derived from reflective thinking for superior generalization.

\subsection{Heuristic-to-Analytic Reasoning (HA-R)}
VLM-based AIGI detectors typically adopt one of two reasoning paradigms. Let $I$ be the input, $A$ be the binary verdict, and $R$ be the textual rationale. The key distinction lies in how they model the joint probability $p(A,R\mid I)$:
\begin{itemize}
\item \textbf{Heuristic Reasoning (H-R)} first generates an answer, modeling the probability as $p(A,R\mid I)=p(A\mid I) \cdot p(R\mid I,A)$. This approach is fast, as inference can be terminated after $A$ is generated. However, the rationale $R$ risks being a post-hoc rationalization rather than a cause for the verdict.
\item \textbf{Analytic Reasoning (A-R)} first formulates a rationale, modeling the probability as $p(A,R\mid I)=p(R\mid I) \cdot p(A\mid I,R)$. This promotes causal thinking and often improves generalization, but at the cost of higher latency since the entire rationale $R$ must be generated.
\end{itemize}

We aim to fuse these paradigms to harness their respective strengths. We develop our HA-R framework through an iterative design process.

\noindent\textbf{Initial Formulation: Guess-Reason-Answer (G-R-A). }
Our first design was a direct combination of the two paradigms, inspired by human cognition: forming an initial ``guess" ($G$), followed by deliberate reasoning ($R$) to reach a final answer ($A$). This sequence, $G\rightarrow R \rightarrow A$, is modeled probabilistically as: $p(G,R,A\mid I) = p(G\mid I) \cdot p(R\mid I,G) \cdot p(A\mid I,G,R)$. While this structure begins to emulate human thought, the connection between $G$ and $A$ remains implicit.

\noindent\textbf{Final Formulation: HA-R with Reflection. }
To create a more structured and powerful reasoning process, we introduced an explicit self-correction step, inspired by the concept of reflective thinking in models like DeepSeek-R1~\cite{guo2025deepseek}. This led to our final HA-R framework, which follows a four-stage sequence: $A_i\rightarrow R_1\rightarrow R_2\rightarrow A_f$. (1) \textbf{Impression ($A_i$):} The model generates a fast, heuristic answer. (2) \textbf{Reasons ($R_1$):} It provides an explanation for its initial impression. (3) \textbf{Reflection ($R_2$):} The model critically re-evaluates its own rationale ($R_1$), explicitly modeling a self-correction mechanism. (4) \textbf{Answer ($A_f$):} Based on all preceding steps, the model outputs its final, more robust analytic answer.

This process is modeled as $p(A_i, R_1, R_2, A_f \mid I) = p(A_i\mid I) \cdot p(R_1\mid I,A_i) \cdot p(R_2\mid I,A_i,R_1) \cdot p(A_f\mid I,A_i,R_1,R_2)$. By conditioning the final answer $A_f$ on a dedicated reflection step, we enable the model to achieve higher accuracy and robustness. The HA-R structure inherently provides two verdicts, a Fast Answer ($A_i$) and a Robust Answer ($A_f$), laying the groundwork for an adaptive inference strategy.

\subsection{Progressive Training for Adaptive Reasoning. }
We instill the HA-R capability into a VLM via a progressive two-stage training paradigm, combining Supervised Fine-Tuning (SFT) and Reinforcement Learning with Verifiable Reward (RLVR)~\cite{guo2025deepseek}.

\noindent\textbf{Stage 1: Cold-Start via Supervised Fine-Tuning. }
We first ``cold-start" a pre-trained VLM using parameter-efficient SFT with LoRA~\cite{hu2022lora}. The objective is to teach the model the HA-R format by maximizing the log-likelihood of the ground-truth sequence:
\begin{equation}
    \mathcal{L}_{SFT} = - \log p(A_i, R_1, R_2, A_f \mid I).
\end{equation}
However, while SFT effectively teaches the desired output structure, it primarily encourages mimicry of annotated reasoning rather than fostering genuine, ``free-form" analytical skills, which can damage model generalization.

\noindent\textbf{Stage 2: RLVR for Adaptive and Free-Form Thinking. }
We use GRPO~\cite{shao2024deepseekmath} to perform RLVR to encourage the model to generate more diverse and higher-quality reasoning. RLVR's key advantage is its ability to use programmatic, rule-based rewards without requiring a separate value model or human preference data. This allows us to guide the model towards our desired reasoning principles while empowering it to ``think freely" beyond the SFT data.

\begin{table*}[ht]  
    \centering
    \resizebox{\linewidth}{!}{\begin{tabular}{c|c|ccc|ccc|c|ccccccc|c}
    \toprule
   \multirow{2}{*}{\textbf{Method}} & \multirow{2}{*}{\textbf{Venues}}& 
     \multicolumn{3}{c|}{\textbf{ID}}&
     \multicolumn{3}{c|}{\textbf{OOD-C}}&\textbf{T2I}&\textbf{IP\&OP}&\textbf{IE}&\textbf{FS}&\textbf{CB}&\textbf{VTO}&\textbf{RMG}&\textbf{PCMG}&\textbf{\textit{Mean}}             \\            
     &&\small ACC.&\small P.&\small R.&\small ACC.&\small P.&\small R.&\small ACC.&\small ACC.&\small ACC.&\small ACC.&\small ACC.&\small ACC.&\small ACC.&\small ACC.&\small ACC.\\
    \midrule
    Qwen-VL-Max & preprint 2025 & 66.21 & 61.35 & 95.88 & 75.35 & 71.08 & 86.51 & 76.71 & 58.17 & \underline{62.33} & 64.61 & 60.64 & 66.08 & \underline{74.12} & \underline{67.01} & 67.12\\
    QvQ-Max & preprint 2025 & 47.66 & 51.02 & 2.86 & 53.21 & 94.23 & 8.06 & 24.40 & 39.04 & 46.34 & 73.00 & 68.15 & 60.10 & 21.87 & 52.33 & 48.61 \\
    Gemini2.5-Pro-0617 & preprint 2025 & 71.45 & 90.67 & 50.74 & \textbf{82.30} & 98.29 & 66.01 & 66.48 & 50.57 & 56.93 & \underline{77.51} & \textbf{74.60} & 63.88 & 61.89 & 66.95 & \underline{67.26}\\
    \midrule
    CNNSpot &CVPR 2020& 77.57&94.85&60.61 &57.29&89.92&17.60
    &69.08&42.14&57.64&71.89&66.33&60.29&35.80&52.76&59.08\\
    
    UnivFD&CVPR 2023 &68.26&74.00&61.07&55.45&75.89&17.60&
    72.31&52.37&\textbf{66.31}&73.86&70.53&65.20&63.07&64.16&65.15\\
    
    NPR&CVPR 2024&73.21&91.86&53.80&50.29&60.00&5.43
    &79.81&36.61&39.96&71.12&64.00&55.69&27.18&51.93&54.63\\
    
    AIDE &ICLR 2025& 89.98&94.29&86.15 & 62.95&88.26&64.34
    &84.66&\textbf{44.22}&56.72&73.24&67.80&\underline{67.80}&59.90&44.47&59.41 \\
    
    Effort &ICML 2025 & 91.23&97.05&85.93 &57.04&96.00&15.79&
\underline{86.40}&43.71&45.42&\textbf{77.68}& 71.10 &63.92&44.40&59.67 &64.06\\
    
    \midrule
    Qwen-VL-2.5-3B & preprint 2025 & 96.44 & 98.02 & 95.14 & 65.92 & 96.68 & 33.50 & 68.24 & 39.80 & 45.26 & 75.31 & 69.98 & 62.51 & 56.24 & 63.43 & 64.31 \\
    Qwen-VL-2.5-7B & preprint 2025 & \underline{96.67} & 97.59 & 96.04 & 69.07 & 95.72 & 40.39 & 86.15 & 40.05 & 45.26 & 74.45 & 68.85 & 62.22 & 52.47 & 59.66 & 65.48\\
    LLaMA3.2-Vision-11B & preprint 2024 & 96.08 & 98.69 & 93.78 & 67.66 & 98.23 & 36.45 & 82.79 & 40.05 & 49.08 & 74.33 & 68.28 & 59.69 & 54.57 & 64.81 & 65.73\\
    \midrule
    \model & - & \textbf{97.26} & 96.45 & 98.41 & \underline{78.46} & 96.30 & 59.87 & \textbf{88.82} & \underline{43.96} & 55.63 & 77.10 & \underline{71.12} & \textbf{67.94} & \textbf{75.79} & \textbf{72.03} & \textbf{72.81} \\
    \bottomrule
    
    \end{tabular}}
    \caption{The comparision of model performance on our proposed {\mirage} benchmark cross various distributions: In-distribution test set (\textit{ID}), human-curated OOD set (\textit{OOD-C}), and 8 subsets from different image generation types. We report the {\model} performance with inference-time adaptive thinking in the table. All the models below ``Gemini2.5-Pro-0617" were trained/finetuned on our \mirage. 
      }
    \label{table:cmp_mirage}
    
\end{table*}

Traditional accuracy reward in RLVR gives an reward of 1 when the answer is correct other wise 0, which encourages a form of brittle overconfidence, where the model is incentivized to make a high-risk guess rather than honestly assessing its own uncertainty. For our reflective reasoning, the model must learn to produce a well-calibrated confidence score in its initial assessment; otherwise, it has no reliable internal signal to trigger deeper, more costly analytic reasoning when it is truly uncertain. We use a ``soft" reward function for $A_i$ as follows:
\begin{equation}
\mathcal{R}_{\text{conf}} = 1 - {cos}(\frac{\pi}{2}p(A_i=c | I)),
\end{equation}
where $c$ is the ground truth label of the sample. The final reward is a weighted sum of the GRPO component, the confidence reward $R_{conf}$, and an accuracy reward $\mathcal{R}_{acc}$ for the final answer $A_f$, a length reward $\mathcal{R}_{len}$ for longer response, a repetition reward $\mathcal{R}_{rep}$ penalizing the repetition, along with a format reward $\mathcal{R}_{fmt}$ for correct formatting. To sum, our RLVR reward is as follows: 
\begin{equation}
\mathcal{R}_{GRPO}=\mathcal{R}_{acc} + \mathcal{R}_{conf} + \mathcal{R}_{len} + \mathcal{R}_{rep} + \mathcal{R}_{fmt}.
\end{equation}

\subsection{Inference-Time Adaptive Thinking}
We perform adaptive thinking at inference time. Given $A_i$, we quantify its confidence by its probability entropy:
\begin{equation}
    H(A_i) = -\sum_{c\in\{{real, fake}\}} p(A_i=c \mid I)\log_2 p(A_i=c\mid I).
\end{equation}
Here, $H(A_i)$ values between 0 and 1, and the high value represents a high uncertainty, i.e., a low confidence. Using a empirical threshold $\tau$, the model dynamically chooses its reasoning path. If confidence is high ($h(A_i) \leq \tau$), it terminates early and returns the fast answer $A_i$. If confidence is low (($h(A_i) > \tau$)), it proceeds with the full HA-R process to generate the more robust answer $A_f$. This mechanism allows {\model} to efficiently allocate computational resources, performing deep reasoning only when the task is sufficiently challenging.

\section{Experiments}
\subsection{Experiment Settings}
\begin{table}[ht] 
    \centering
    \scalebox{0.8}{\begin{tabular}{c|c|ccc}
    \toprule
    \multirow{2}{*}{\textbf{Method}} & \multirow{2}{*}{\textbf{Training Set}}  & \multicolumn{3}{c}{\textbf{Chameleon}}\\
    & &\small ALL&\small FAKE&\small REAL\\
     
    \midrule
    CNNSpot\textsuperscript{*} & GenImage & 68.89&9.86&99.25 \\
    UnivFD\textsuperscript{*}& GenImage & 60.42&\textbf{85.52}&41.56 \\
    NPR\textsuperscript{*}& GenImage & 57.81&1.68&\textbf{100.00}\\
    AIDE\textsuperscript{*} & GenImage & 65.77&26.80&95.06 \\
    \midrule
    CNNSpot& {\mirage} & 64.21&38.08&83.85 \\
    UnivFD& {\mirage} & 50.29&63.19&40.60 \\
    NPR& {\mirage} & 62.95&15.49&98.62\\
    AIDE & {\mirage} & 60.69&45.32&72.25 \\
    Effort& {\mirage} & 64.36&17.07&99.91\\
    \midrule
    \model & {\mirage} & \textbf{78.31} & 50.29 & 98.37\\
    \bottomrule
    
    \end{tabular}}
    \caption{Model performance on the Chameleon benchmark. The numbers in the table refer to the accuracy on different subsets of the benchmark. The result of methods with \textsuperscript{*} is taken from~\cite{yan2024sanity}.}\label{table:cross_benchmark}
    \vspace{-1.25mm}
\end{table}

\begin{table}[ht]
    \centering
    \resizebox{\columnwidth}{!}{\begin{tabular}{c|cccccc}
    \toprule
    \multirow{2}{*}{\textbf{Method}}& \multicolumn{3}{c}{\textbf{ID}}&\multicolumn{3}{c}{\textbf{OOD-C}} \\
    &\small ACC.&\small P.&\small R.&\small ACC.&\small P.&\small R.\\
    \midrule
    w/o Reasoning & 96.67 &\textbf{97.59} & 96.04 & 69.07 & 95.72 & 40.39 \\
    H-R & 96.12&95.53& 97.16 & 72.36 & 94.23 & 48.36 \\
    A-R & 93.90&93.39&95.10 & 73.13&89.29 & 53.90 \\
    \midrule
    G-R-A & 96.54 & 96.49 & 96.93 & 75.25 & \textbf{96.99} & 52.87 \\
    \midrule
    HA-R-$A_i$ & 97.26 & 97.18 & 97.62 & 76.63 & 96.33 & 56.09 \\
    HA-R-$A_f$ & 97.14 & 96.34 & 98.30 & 78.13 & 96.01 & 59.38 \\
    HA-R & \textbf{97.26} & 96.45 & \textbf{98.41} & \textbf{78.46} & 96.30 & \textbf{59.87} \\
    \bottomrule    
    \end{tabular}}
     \caption{Ablations on the reasoning format. The model with the best performance is in \textbf{bold}. Here, HA-R-$A_i$, HA-R-$A_f$, HA-R  refer to our model prediction from the ``impression", ``final answer", and inference-time adaptive thinking. }
    \label{table:ablation_cot}
\end{table}

\noindent\textbf{Baseline.} We evaluated 5 specialized state-of-the-art (SOTA) AIGI detection models, including CNN-Spot~\cite{cnn-spot}, UnivFD~\cite{univfd}, NPR~\cite{npr}, AIDE~\cite{yan2024sanity}, and Effort~\cite{effort}. Besides, we selected 3 state-of-the-art (SOTA) closed-source models for zero-shot evaluation: Qwen-VL-Max~\cite{bai2023qwenvlversatilevisionlanguagemodel}, QvQ-Max~\cite{QwenBlog2024QVQMax}, and Gemini2.5-Pro~\cite{comanici2025gemini}. We also finetuned 3 open-source models, Qwen-VL-2.5-3B~\cite{bai2025qwen2}, Qwen-VL-2.5-7B~\cite{bai2025qwen2}, and LLaMA3.2-Vision-11B~\cite{grattafiori2024llama3herdmodels}, as foundational models to evaluate the effectiveness of our approach.

\noindent\textbf{Dataset.} To comprehensively evaluate the capability of existing methods in ID and OOD scenarios, we conducted experiments according to the setting in the dataset section. Additionally, we used a well-known public benchmark for real-world AIGI detection, Chameleon~\cite{yan2024sanity}, for cross-benchmark evaluation.

\noindent\textbf{Implementation Details.} We selected hyperparameters on a sit-alone development set sampled from the training set. We employed a pre-trained Qwen2.5-VL-7B~\cite{bai2025qwen2} as the backbone of \model. The LoRA rank was set to 256 in the SFT stage for 4000 steps on 64 NVIDIA H20 GPUs, and to 8 in the RLVR stage for 50 steps on 32 NVIDIA H20 GPUs. In both stages, the batch size was set to 8. We empirically selected $\tau=0.96$, where $10\%$ samples go to ``deep thinking" in our development set. In the cold-start stage, we constructed the training set to be with $70\%$ $A_i$ the wrong answer and $30\%$ the correct answer, so as to teach the model both to correct or enhance $A_i$ in $A_f$. Other implementation details and hyperparameters are included in the appendix.

\noindent\textbf{Metrics.} Following previous works~\cite{yan2024sanity,effort,cnn-spot}, we report the classification accuracy (ACC.), precision (P.), and recall (R.) in our experiments, where ACC. is our primary evaluation metric.

\subsection{Comparison with State-of-the-Arts}
\noindent\textbf{On benchmark \mirage. }We evaluated {\model} against SOTA specialized detectors and VLMs on our {\mirage} benchmark, with results summarized in Tab.~\ref{table:cmp_mirage}. The findings underscore the effectiveness of our approach. Overall, {\model} not only achieves the highest accuracy on the ID test set but also establishes a new SOTA on the OOD splits, surpassing the next-best competitor, Gemini-2.5 Pro, by a significant margin of over $5\%$. This robust generalization extends to a fine-grained level: when analyzed across the eight distinct generative attack patterns, {\model} secures the top rank in four categories and places in the top three for seven out of eight, proving its consistency against a wide spectrum of sophisticated manipulation techniques. Notably, its performance on the T2I subset, where it leads by $2\%$, specifically evaluates generalization to unseen vanilla generative models — a common setting in prior work. The comprehensive generalization across diverse in-the-wild scenarios validates the advantages of our adaptive and reflective reasoning framework.

\noindent\textbf{On benchmark Chameleon. }
We conducted a rigorous cross-benchmark evaluation on the public Chameleon dataset~\cite{yan2024sanity} to further assess the OOD generalization of {\model}. The results, presented in Tab.~\ref{table:cross_benchmark}, demonstrate the generalization of our approach. Our model surpasses all other methods by a significant margin, establishing a new state-of-the-art on this challenging benchmark. This is a critical finding, as Chameleon's images are sourced from online AIGI communities and are known to be difficult even for human experts to identify. Remarkably, without any exposure to Chameleon's data during training, {\model} achieves a balanced performance, correctly identifying half of the sophisticated forgeries while maintaining a near-perfect accuracy ($98.37\%$) on real images. This showcases its ability to learn generalizable forgery cues rather than overfitting to artifacts from a specific training distribution. Moreover, this experiment highlights an equally important contribution of our work: the {\mirage} dataset itself. We observed that when baseline models are trained on {\mirage} training set instead of a conventional dataset like GenImage, their accuracy on Chameleon's FAKE subset consistently improves. This provides strong empirical evidence that our {\mirage} benchmark, with its diverse and challenging pipeline-generated examples, better prepares models for the complexities of in-the-wild AIGI detection.

\subsection{Quantitative Study}
     

    
    
    
    
    

\begin{table}[t] 
    \centering
    \resizebox{\linewidth}{!}{\begin{tabular}{cc|cccc|cc}
    \toprule
    \multirow{2}{*}{\textbf{Method}} &\multirow{2}{*}{Original}& \multicolumn{4}{c|}{\textbf{JPEG Compression}}& \multicolumn{2}{c}{\textbf{Gaussian Blur}}\\
    && \small QF$=$50 & \small QF$=$70 &\small QF$=$90&\small QF$=$95&$\sigma=$1.0&$\sigma=$2.0\\
    \midrule

    CNNSpot&77.57&77.13&77.85&78.45&77.97&75.10&73.19 \\
    
    UnivFD&68.26&69.07&68.00&67.94&68.06 &69.13&67.94\\
    
    NPR&73.21&58.69&66.33&71.52&71.82&71.88&62.39\\
    
    AIDE &89.98&87.04&85.97&89.13&89.97&89.85&87.94 \\
    
    Effort&91.23&84.54&88.30&88.60&88.78&80.12&62.57\\
    \midrule
    \model & \textbf{97.26} & \textbf{95.43} & \textbf{96.14} & \textbf{96.08} & \textbf{95.19} & \textbf{91.33} & \textbf{88.30} \\
    \bottomrule
    
    \end{tabular}}
    \caption{Robustness of Classification Accuracy on different levels of JPEG compression and Gaussian blurring attack. The accuracy is calculated over the ID subset of the \textbf{\mirage} Benchmark. }
    \label{table:robust}
\end{table}

\noindent\textbf{Ablation studies.} 
We conducted a comprehensive ablation study, with results presented in Tab.~\ref{table:ablation_cot}. The findings confirm our hypotheses. First, any form of reasoning (H-R, A-R) significantly enhances OOD generalization compared to a non-reasoning baseline. Second, introducing an explicit reflection step in our HA-R-$A_f$ model yields a substantial OOD accuracy gain over our initial G-R-A design ($78.13\%$ vs. $75.25\%$), demonstrating the critical role of self-correction. Most importantly, our final adaptive HA-R model achieves the best overall performance ($78.46\%$ OOD Acc.), even surpassing the strategy of always using the final answer. This indicates that by dynamically choosing when to engage in deep, reflective thinking, our model strikes an optimal balance between accuracy and efficiency, preventing ``over-thinking" and ``wrong impression".

\noindent\textbf{Robustness evaluation.} 
The robustness of an AIGI detector against real-world corruptions is critical for its practical application, as perturbations like compression and noise can erase forgery artifacts. We therefore evaluated {\model} and baselines on our ID test set subjected to standard JPEG compression and Gaussian noise, following~\cite{cnn-spot}. The results in Tab.~\ref{table:robust} highlight our model's robustness. Under JPEG compression, {\model} maintains consistently high accuracy while baseline performance drops significantly. It also outperforms competitors against Gaussian noise. Crucially, {\model} achieves this robustness without being trained on corresponding data augmentations (e.g., Gaussian noise), which were commonly used to develop the baseline models.

\section{Conclusion}
In this work, we take a step towards addressing the challenge of in-the-wild AIGI detection by introducing {\mirage}, a benchmark designed to better reflect real-world complexity, and proposing {\model}, a VLM with a novel adaptive reasoning framework. Our experiments show that {\model} demonstrates strong performance across our benchmark and public datasets, with notable improvements in both generalization and robustness. Future work can focus on scaling data annotation and mitigating potential biases inherited from the backbone VLMs and VLMs for annotation. To conclude, we believe this work offers a useful step toward developing more practical and reliable AIGI detection systems for our increasingly complex digital world.
\clearpage
\bibliography{aaai2026}

\clearpage
\appendix
\maketitle
\section{Elaboration of Concepts}
In this paper, we address the task of in-the-wild AIGI detection. This section formally defines this task and its associated core concepts.

\paragraph{Defining In-the-Wild AIGI Detection.}
Distinct from detection in controlled, laboratory settings, we define in-the-wild AIGI detection as the identification of AI-generated images as they naturally occur and circulate on the internet. The term ``naturally-occurring" implies a set of challenging characteristics that existing benchmarks often overlook:

\begin{itemize}
\item \textbf{Contextual Co-existence:} In-the-wild AIGI are not found in isolation but are interspersed with authentic images within the same context (e.g., a single social media feed or online community). This necessitates detection models that can operate without domain-specific shortcuts, distinguishing real from fake in a shared environment.
\item \textbf{Wide Quality Spectrum:} The images exhibit a vast range of perceptual quality, from ``obviously fake" artifacts resulting from novice attempts or poor editing, to very realistic outputs. The source of these images is uncontrolled, leading to extreme variability and ``noise."
\item \textbf{Sophisticated Generation Workflows:} To create highly deceptive forgeries, malicious actors often employ complex, multi-stage processes. This goes far beyond using a single generative model and includes chaining multiple models, extensive prompt engineering, and significant manual post-processing (e.g., color grading, inpainting, manual edits).
\end{itemize}

\paragraph{Key Terminology.}
Building on the characteristics above, we introduce the following terminology to categorize AIGI origins:

\begin{itemize}
\item \textbf{Generation from Vanilla Generators:} This refers to images produced directly from a single, off-the-shelf generative model without significant modification. While such images exist online, they represent only a fraction of the in-the-wild landscape and are the primary focus of many previous benchmarks.
\item \textbf{Generation from Composite Pipelines:} This describes the creation of AIGI through the sophisticated, multi-step workflows mentioned earlier. These images represent a more challenging and realistic forgery scenario.
\item \textbf{Human-Curated AIGI:} This term refers to a specific subset of our benchmark sourced from the internet. These are images unanimously identified as ``obviously fake" by multiple human experts. Their artificiality may stem from various causes, including conspicuous and poorly executed post-processing (e.g., bad retouching). They are distinct from ``vanilla" outputs due to the presence of manual editing, and from high-quality ``pipeline" outputs due to their blatant and easily recognizable flaws.
\end{itemize}

The unique challenges posed by in-the-wild AIGI demand that detectors possess the following key attributes:

\begin{itemize}
\item \textbf{Generalization.} As generative models rapidly evolve, detectors must perform reliably on OOD images from unseen models and diverse scenarios. Traditionally, generalization in AIGI detection is defined as the ability to identify images from unseen generators~\cite{zhu2023genimage}. In this work, we argue for a more comprehensive, two-fold definition:
\begin{enumerate}
\item \textit{Generalization to Unseen Models:} Aligning with the traditional view, this evaluates performance on images from generators not encountered during training. Our T2I subset is explicitly designed for this purpose.
\item \textit{Generalization to Unseen Scenarios and Forgery Types:} This more challenging dimension requires a model to be robust across different online contexts and a wide spectrum of forgery quality and techniques. Our OOD-C subset tests this by introducing a new scenario with human-curated fakes, while other splits evaluate generalization to various high-quality, pipeline-generated images.
\end{enumerate}

Experiments confirm that {\model} excels in both dimensions, demonstrating strong generalization in both the traditional and our more comprehensive setting.

\item \textbf{Explainability.} In high-stakes applications like content moderation, providing clear, human-understandable rationales for decisions is crucial for user trust and transparency. We define explainability in this context as the cogent, textual reasoning the VLM provides with its prediction, distinguishing it from a more formal definition in the ante-hoc or post-hoc methods common in the broader XAI field~\cite{selvaraju2016grad,lin2022saw,gong2024structured}. Due to page constraints, a detailed evaluation of our model's explanations is provided later in this appendix instead of the main paper. 

\item \textbf{Efficiency.} Practical deployment, especially in real-time applications like media analysis, imposes strict low-latency requirements. An ideal detector must therefore be not only accurate but also computationally efficient. This necessity motivated the design of {\model}'s adaptive reasoning mechanism, which dynamically balances inference speed with performance. A full analysis of our model's efficiency is deferred to later sections in this appendix due to page limitations of the main paper.

\end{itemize}

\section{Related Works}
We present a detailed version of related works in this section, as a complement to the brief one in the main paper. 

\paragraph{Benchmarks in AIGI Detection.}
The development of AIGI detection has been closely tied to the evolution of its evaluation benchmarks, progressing from early datasets focused on GANs, such as CNN-Detection~\cite{cnn-spot}, to more contemporary collections like GenImage~\cite{zhu2023genimage} and WildFake~\cite{hong2024wildfake} that incorporate outputs from advanced diffusion models. While foundational, a critical limitation of these benchmarks is their failure to capture the true complexity of AIGI encountered in-the-wild, as they often consist of ``clean" images from specific, known models, which hinders a detector's ability to generalize. Recognizing this gap, recent efforts like DRCT-2M~\cite{chen2024drct}, Chameleon~\cite{yan2024sanity} and Ivy-Fake~\cite{zhang2025ivy} have made significant strides by sourcing sophisticated AIGI from online communities, which better reflect the results of extensive manual parameter tuning. 

Despite this progress, two crucial challenges remain unaddressed. First, by sourcing real and AI-generated images from disparate contexts — such as photography websites versus AIGI communities — these benchmarks risk introducing domain-specific biases that a model can exploit, rather than learning the intrinsic artifacts of generation. Second, they do not fully account for the complex creation pipelines of high-quality forgeries, which often involve significant post-processing applied to images generated from fine-tuned models. To bridge these gaps, we introduce {\mirage}, a new benchmark designed to rigorously emulate in-the-wild scenarios. Besides clean AIGI from vanilla generation models, {\mirage} also involve human-collected and curated AIGI that co-exist within the same online contexts, and challenging AIGI from composed pipelines of state-of-the-art generation and post-processing techniques. This dual-source approach ensures {\mirage} provides a more realistic and challenging testbed for evaluating the true generalization capabilities of AIGI detectors.

\paragraph{AIGI Detection as Classification.}
The detection of AIGI is predominantly framed as a binary classification task. Initial approaches focused on identifying low-level artifacts using specialized architectures like Convolutional Neural Networks in CNNSpot~\cite{cnn-spot}. Other methods, such as NPR~\cite{npr}, concentrated on specific generation footprints, like those introduced by up-sampling operations. Despite their initial success, a critical limitation of these early methods is their poor generalization to images from unseen generative models. To address this challenge, research has evolved in two main directions. One line of work aims to combine different feature types, such as the frequency and semantic information used in AIDE's~\cite{yan2024sanity} dual-stream framework. A more recent and dominant trend involves harnessing the power of large pre-trained models. For example, UnivFD~\cite{univfd} and Effort~\cite{effort} leverage features from CLIP to build more universal detectors, demonstrating that high-level semantic traces are crucial for generalization. However, even these advanced methods face significant hurdles. A notable drawback is their ``black-box" nature, as they often struggle to provide explanations for their decisions~\cite{gao2025fakereasoning}. More importantly for practical applications, their generalization is rarely tested in the wild. The performance of these models on images that have undergone real-world post-processing, such as compression, resizing, and filtering, remains largely unevaluated. 

Our work directly confronts this gap. We argue that true generalization must encompass not only unseen models but also unseen real-world perturbations. We introduce {\mirage}, a novel framework that extends the frontier of generalization to images ``in the wild." By strategically leveraging the rich knowledge within a pre-trained VLM, our {\model} achieves superior performance in these highly challenging and realistic scenarios, setting a new benchmark for robust AIGI detection.

\paragraph{VLMs in AIGI detection. }
The advent of VLMs has catalyzed a shift in AIGI detection towards more generalizable and explainable solutions. These models are typically prompted to produce not only a classification verdict but also a textual rationale. Existing works in this domain can be broadly classified into two distinct reasoning paradigms. One dominant paradigm, which we term \textbf{heuristic reasoning} (H-R), involves generating a verdict first, followed by a justification. This answer-first approach is employed by FakeBench~\cite{li2024fakebench}, which evaluates the verdict and explanations independently, and FakeClue~\cite{wen2025spot}, which generates both in a single forward pass. 
The primary advantage of this paradigm is speed, but its main drawback is that the explanations can be mere post-hoc rationalizations, disconnected from the actual inference process, which compromises trustworthiness and generalization. An alternative paradigm, \textbf{analytic reasoning} (A-R), reverses this sequence by prioritizing a step-by-step thought process before delivering a final verdict. Drawing inspiration from CoT methods, FakeReasoning~\cite{gao2025fakereasoning} makes the model explicitly reason about image content first. This structured thinking is further advanced by IVY-FAKE~\cite{zhang2025ivy} and So-Fake~\cite{huang2025so}, which use dedicated tags (i.e., \verb|<think>| from DeepSeek-R1~\cite{guo2025deepseek}) to encapsulate the reasoning process, sometimes coupled with advanced training strategies like Group Relative Policy Optimization (GRPO). While analytic reasoning generally yields superior generalization, it suffers from higher latency and the risk of ``over-thinking" simple cases, where excessive analysis can paradoxically introduce errors~\cite{chen2024not,tu2025learning}\footnote{Avoiding over-thinking, i.e., Long-to-Short on VLMs, is a very active field in VLM research. Since we are focusing on the AIGI detection task in this paper, we save sentences of a detailed introduction to this field, but only involve related works in the AIGI detection problem. \url{https://github.com/hongcheng-gao/awesome-long2short-on-lrms} is useful survey and collection of state-of-the-art Long-to-Short VLMs used for general purpose.}. 

This dichotomy presents a fundamental trade-off: the speed of heuristic reasoning versus the generalization of analytic reasoning. We contend that neither approach alone is optimal for the diverse challenges of in-the-wild detection. An ideal agent should be swift for straightforward cases but deliberate for complex ones. In this paper, we introduce {\model}, a framework reasoning with our proposed Heuristic-to-Analytic Reasoning. Our model provide a fast answer and a more robust answer. It can estimate the confidence of the first answer, adaptively choose whether to reflectively think or not. To our knowledge, {\model} is the first AIGI detection framework to integrate both reflective and adaptive thinking, enabling a dynamic allocation of computational resources based on task difficulty.
\begin{figure*}[ht]
\centering
\begin{tcolorbox}[listing only, 
                  listing options={basicstyle=\ttfamily\footnotesize},
                  colback=gray!10, 
                  title=Annotation Protocols]
Label whether the image is real or AI-generated. There are three options: ``Real", ``AI-generated", and ``Uncertain".

The principle is: If the image gives you a feeling that it is AI-generated or if there are obvious flaws, annotate it as ``AI-generated fake". If there are no obvious flaws and it looks like a normal photo that you would expect to see on the Internet everyday, annotate it as ``Real". If you are unsure, annotate it as ``Uncertain". (Only label an image as ``Real" if you are certain it is real, and only label it as ``AI-generated" if you are certain it is AI-generated. If you are unsure, choose ``Uncertain".)

According to the previous annotation rules, if you cannot distinguish whether the graininess or blurriness is caused by AI or not, it should be classified as ``Uncertain".
\begin{itemize}
    \item If the image has relatively obvious flaws or an ``feeling of AI generation" (such as excessively smooth skin on the humans in the image, evidence of cutout edges around the objects, or limb deformities), it should be labeled as ``AI-generated".
    \item Do not focus excessively on minor details (but be attentive to hands and hair of humans). Focus on obvious flaws and unnatural features.
    \item The ``Uncertain" option should be used when there are unnatural areas in the image, but you are not sure if they were caused by AI generation or by other factors such as image compression or post-processing design.
\end{itemize}
\end{tcolorbox}
\caption{Annotation protocol of the human-curated images. }
\label{list:curation_proc}
\end{figure*}

\begin{figure*}[ht]
\centering
\begin{tcolorbox}[listing only, 
                  listing options={basicstyle=\ttfamily\footnotesize},
                  colback=gray!10, 
                  title=Common Editing Instructions]

\begin{itemize}
    \item Change the background, keep the subject in the exact same position and pose.
    \item Add ASCII style keyword about this image, no additional letters.
    \item Convert to pencil sketch with natural graphite lines, cross-hatching, and visible paper texture.
    \item Reduce brightness while keeping color palette and composition unchanged.
    \item Remove all background people while preserving the subject's position and lighting.
    \item Replace the subject, maintain the same style.
    \item Rotate the camera 90 degrees to view from behind.
    \item Switch to side profile while keeping facial features unchanged.
    \item Turn the subject to face forward completely, including both head and body.
    \item Transform to 1960s pop art style with bright colors and commercial graphics.
   
\end{itemize}
\end{tcolorbox}
\caption{The set of common editing instructions.}
\label{list:edit-instructions}
\end{figure*}

\section{Dataset Details}
\paragraph{Human Curation Details.} To acquire our human-curated images, we followed a multi-stage collection and annotation protocol. First, we sourced an initial pool of 15,000 images from real-world application scenarios on social media and e-commerce platforms (source $\mathcal{A}$ and $\mathcal{B}$ in the paper). The date of these images ranges from 2022 to 2025. This pool consisted of two groups: (1) images that had been flagged as ``suspected AI" by the platforms' internal systems (human or model-based), and (2) a randomly sampled set of images without such flags, which were predominantly authentic.

To ensure the accuracy of these preliminary labels, we recruited 29 annotators in AI image analysis for a comprehensive re-annotation task. We offered the an annotation protocol to the annotators to train them to be experts. The protocol is shown in Fig.~\ref{list:curation_proc}.

\begin{figure*}[ht]
    \centering
    \includegraphics[width=0.85\linewidth]{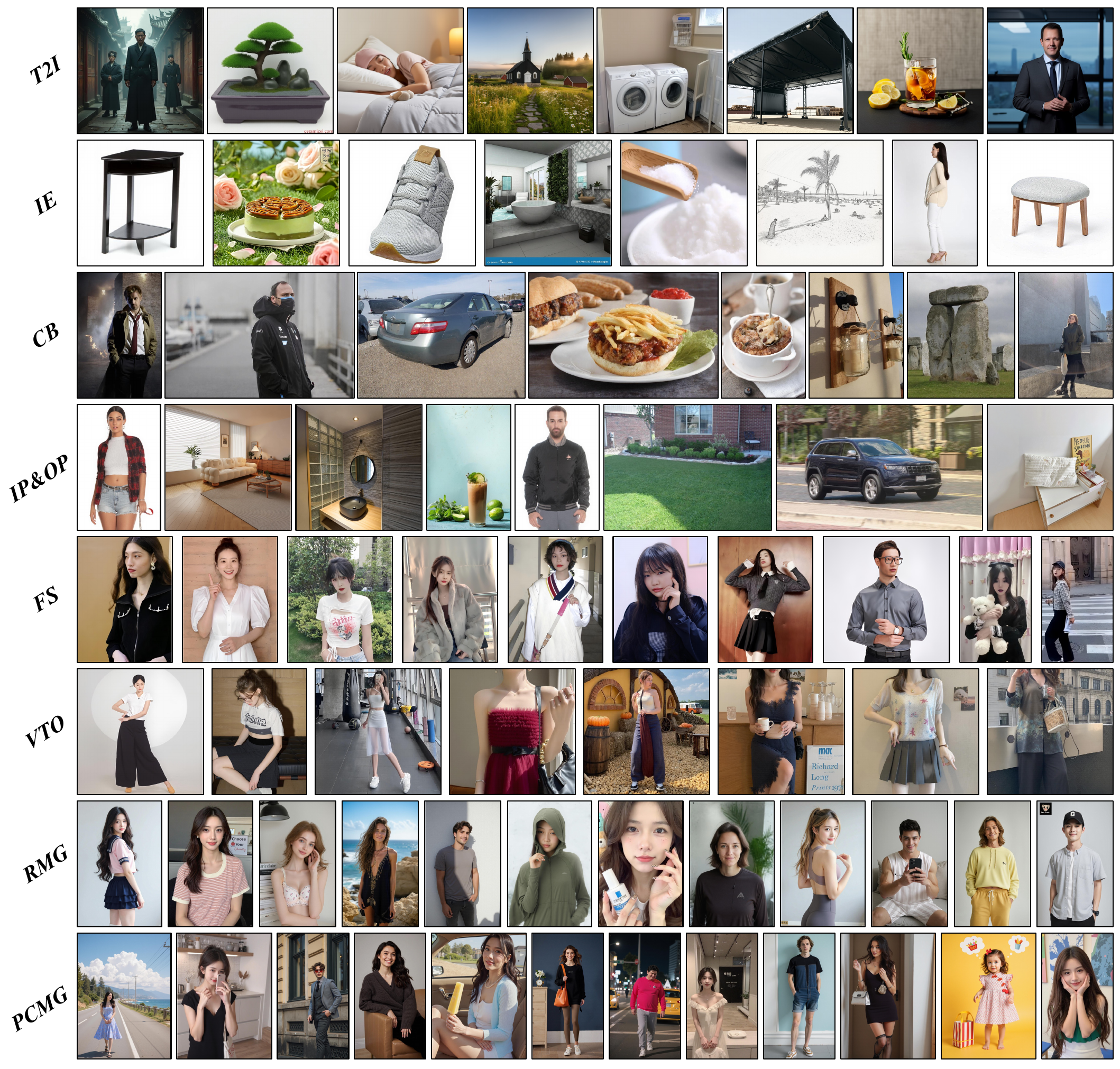}
    \caption{The visualization of images corresponding to eight generation patterns in \mirage.}
    \label{fig:vs_images}
\end{figure*}

We evaluated the annotator performance by checking their annotation consistency and accuracy on ``pilot annotation" with a batch of sampled images. The training persists until the annotator is able to give a consistent and precise annotation. We call these trained annotators experts. 

Then, each image was independently assigned to three different experts. The annotation options were ``Real", ``AI-generated", or ``Uncertain". To minimize subjective bias, annotators were explicitly instructed to select ``Real" or ``AI-generated" only when they were highly confident, and to use the ``Uncertain" label for any ambiguous cases.

Finally, upon collecting the annotations, we employed a stringent three-step filtering process to derive our final high-quality dataset:
\begin{itemize}
    \item \textit{Consensus Check}: We retained only those images for which all three independent annotators provided a unanimous label.
    \item \textit{Uncertainty Removal}: We then discarded all images from this consensus set that were unanimously labeled as "Uncertain."
    \item \textit{Consistency Verification}: As a final quality control measure, we removed any remaining images whose expert-verified label contradicted their original platform-provided tag.
\end{itemize}
This process yielded a dataset of 11,559 images, comprising 8,465 verified real images and 3,094 verified AI-generated images.

\begin{figure}[ht]
    \centering
    \includegraphics[width=1\linewidth]{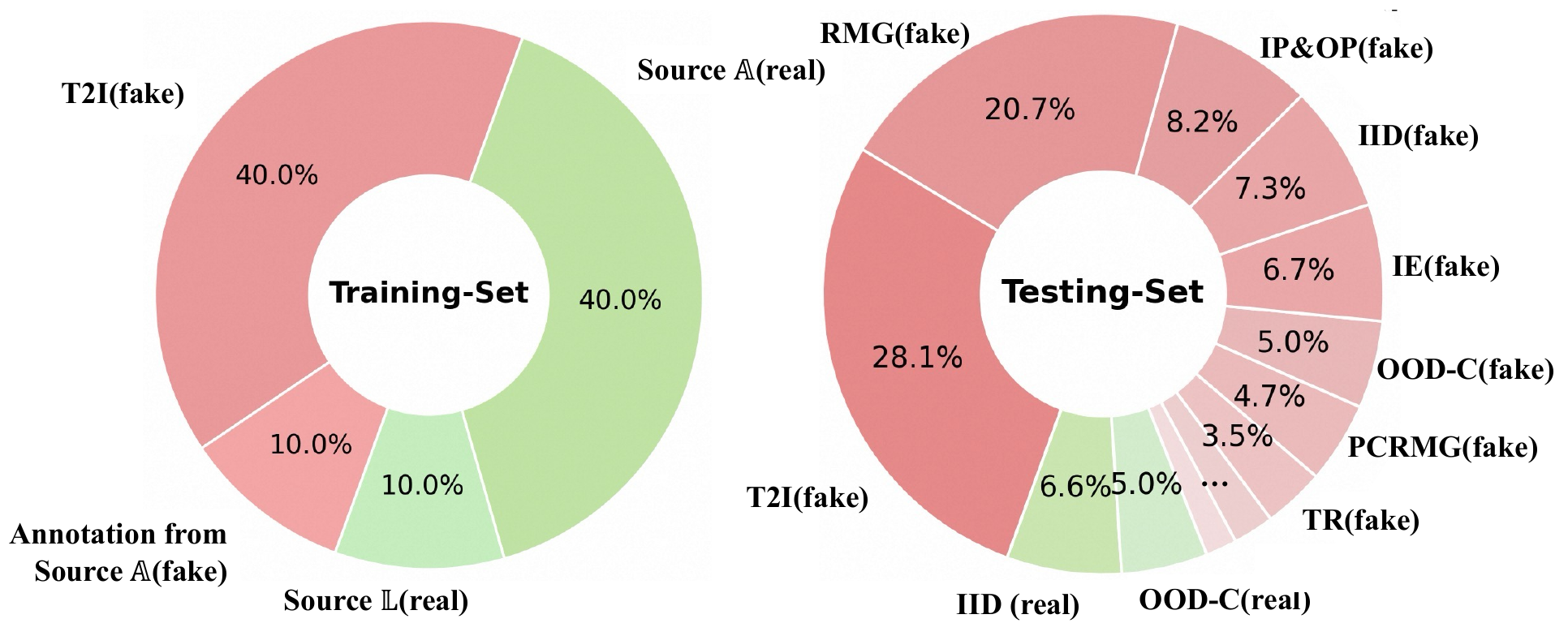}
    \caption{Data distribution of our \mirage. The left figure represents the proportion of fake and real images in training-set, while right represents the proportion in benchmark}
    \label{fig:mirage}
\end{figure}

\begin{table*}[ht] 
    \centering
    \scalebox{0.7}{\begin{tabular}{c|p{12cm}|c}
    \toprule
    Generation Patterns &\multicolumn{1}{c|}{Models}& Pipelines\\
    \hline

    Text-to-Image (T2I)&Flux.1-dev~\cite{flux2024}, SD-3.5-Large~\cite{sd3}, SD-XL~\cite{sdxl}, Infinity~\cite{Infinity}, CogView4-6B~\cite{cogview}, UniDiffuser~\cite{unidiffuser}, Bagel~\cite{bagel}, HunyuanDiT~\cite{hunyuandit}, PixArt-alpha~\cite{pixartalpha}, HiDream~\cite{cai2025hidream}, and Wanx2.1-t2i-turbo~\cite{wan2.1-t2i}.& 11\\ 
    \hline
    Instruction-based Editting (IE)&LLama-3.1-8B-Instruct~\cite{grattafiori2024llama3herdmodels}, Flux.1-Kontext-dev~\cite{Flux.1-Kontext}, Bagel~\cite{bagel}, Wanx2.1-imageedit~\cite{wan2.1-edit}, and Instruct-pix2pix~\cite{brooks2022instructpix2pix}.&4\\
    \hline
    Change background (CB)&Flux.1-Redux-dev~\cite{flux2024}, Flux.1-Fill-dev~\cite{flux2024}, BirefNet~\cite{zheng2024birefnet}, LLama-3.1-8B-Instruct~\cite{grattafiori2024llama3herdmodels}, U2-Net~\cite{U2-Net}, and 
    InSPyReNet~\cite{InSPyReNet}.&3\\
    
    \hline
    Inpainting/Outpainting (IP\&OP)&Wanx-x-painting~\cite{wanx-x-painting}, Wanx-image-out-painting~\cite{image-out-painting}, BrushNet~\cite{brushnet}, Flux.1-Fill-dev~\cite{flux2024}, stable-diffusion-xl-inpainting~\cite{sdxl}, SD-3.5-Medium~\cite{sd3}, YOLO~\cite{yolo}, SAM~\cite{kirillov2023segment}, and GroundingDino~\cite{liu2024grounding}.&8\\
    \hline
    Face Swapping (FS)&Retinaface~\cite{retinaface}, Codeformer~\cite{codeformer}, Inswapper~\cite{inswapper}, YOLO~\cite{yolo}, GFPGAN~\cite{GFPGAN}, and Real-ESRGAN~\cite{wang2021realesrgan}.&3\\
    \hline
    Virtual Try-On (VTO)&SAM~\cite{kirillov2023segment}, GroundingDino~\cite{liu2024grounding}, Flux.1-Redux-dev~\cite{flux2024}, Flux.1-Fill-dev~\cite{flux2024}, Real-ESRGAN~\cite{wang2021realesrgan}, BirefNet~\cite{zheng2024birefnet}, FLUX.1-Turbo-Alpha~\cite{flux2024}, SD-1.5-inpaint~\cite{sd15}, and CatVTON~\cite{catvton}.&3\\
    \hline
    Realistic Model Generation (RMG)&Flux.1-dev~\cite{flux2024},  Flux.1-Redux-dev~\cite{flux2024}, Flux.1-Fill-dev~\cite{flux2024}, SD-3.5-Large~\cite{sd3}, SD-XL~\cite{sdxl}, 16 different loras~\cite{hu2022lora}, SAM~\cite{kirillov2023segment}, GroundingDino~\cite{liu2024grounding}, GFPGAN~\cite{GFPGAN}, and BirefNet~\cite{zheng2024birefnet}&16\\
    \hline              
    Pose-consistent Model Generation (PCMG)&Flux.1-dev~\cite{flux2024},  Flux.1-Redux-dev~\cite{flux2024}, Flux.1-Fill-dev~\cite{flux2024}, SD-3.5-Large~\cite{sd3}, SD-XL~\cite{sdxl}, 16 different loras~\cite{hu2022lora}, SAM~\cite{kirillov2023segment}, GroundingDino~\cite{liu2024grounding}, and BirefNet~\cite{zheng2024birefnet}, Controlnet~\cite{zhang2023adding}, and DWPose~\cite{DWPose}&16\\

    \bottomrule
    
    \end{tabular}}
    \caption{Overview of the proposed eight generation patterns. For each generation pattern, we enumerate the integrated models and report the number of pipelines implemented using these models.}
    \label{table:appendix_models}
\end{table*}

\paragraph{Bias in the Dataset. }
We collected real and fake images from Source $\mathbb{A}$ and Source $\mathbb{B}$, mainly spanning the period from 2022 to 2025. During the human curation process, we identified a data bias: the number of AIGC images in 2022 is relatively small, while the number in 2025 is significantly larger. This data bias aligns with the actual progression of AIGC technology in the real world, and is consistent with the in-the-wild nature of our \mirage~dataset.

\paragraph{AIGI Generation via Vanilla Generators.}
As stated before, in this paper, we define ``AIGI Generation via Vanilla Generators" as generating fake images using single text-to-image models. Similar to previous works, we utilize several state-of-the-art text-to-image or text-video-image models, including Flux.1-dev~\cite{flux2024}, SD-3.5-Large~\cite{sd3}, SD-XL~\cite{sdxl}, Infinity~\cite{Infinity}, CogView4-6B~\cite{cogview}, UniDiffuser~\cite{unidiffuser}, Bagel~\cite{bagel}, HunyuanDiT~\cite{hunyuandit}, PixArt-alpha~\cite{pixartalpha}, HiDream~\cite{cai2025hidream}, and Wan2.1~\cite{wan2.1-t2i}. In the process of ``AIGI Generation via Vanilla Generators", we employ Qwen2.5-VL-32B-Instruct~\cite{bai2025qwen2} to caption images from Source $\mathbb{A}$, Source $\mathbb{B}$, and Laion Dataset $\mathbb{L }$. For text-to-image models, we simply use prompts to generate images. For the text-to-video models, we user prompts or origin images to generate videos. Then, we randomly sample the frames from the generated video. Similar to previous methods, we evaluate the model’s OOD detection ability on images generated by different text-to-image models. Specifically, in the \mirage~training set, we utilize models such as Flux.1-dev, SD-3.5-Large, SD-XL, Infinity, PixArt-alpha, and HunyuanDIT. For the T2I split in \mirage~benchmark , we employ models including CogView4-6B, Bagel, Wan2.1, HiDream, and UniDiffusers to assess the OOD generalization capability.

\paragraph{AIGI Generation via Composite Pipelines.} As stated before, in this paper, we define the seven generation patterns other than T2I as ``AIGI Generation via Composite Pipelines". These are constructed by assembling multiple modules (including image generators, LoRAs~\cite{hu2022lora}, ControlNets~\cite{zhang2023adding}, Upsampling and downsampling modules, Gaussian blur module, and other image processing models and modules) using either the Python Interpreter or the ComfyUI platform. The main types include Intrusion-based Editing (IE) and Change Background (CB), Inpainting/Outpainting (IP\&OP), Face Swapping (FS), Virtual Try-On (VTO), Realistic Model Generation (RMG), and Pose-consistent Model Generation (PCMG).

For \textbf{Instruction-based Editing (IE)}, we manually compile a set of common editing instructions, as shown in Fig.~\ref{list:edit-instructions}. IE is primarily implemented as an image-to-image process by using the Python Interpreter: we randomly sample an instruction from the set. Then, the origin image is processed by an instruction-based editing model, such as Bagel~\cite{bagel} (a state-of-the-art unified understanding and generation model).

\begin{figure*}[ht]
    \centering
    \includegraphics[width=1\linewidth]{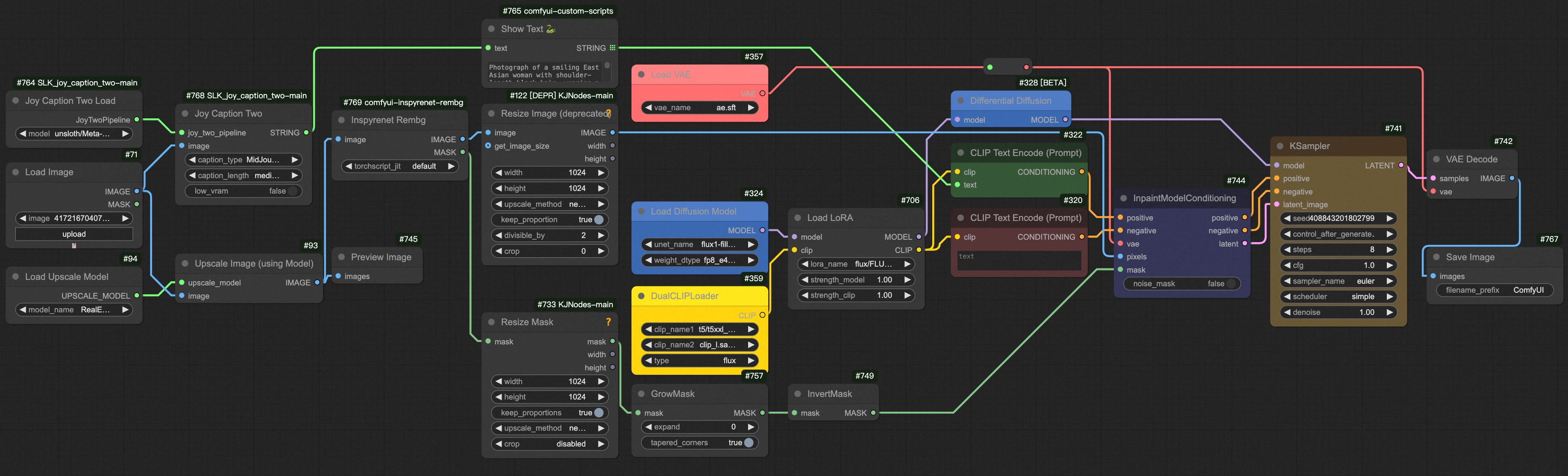}
    \caption{The ComfyUI workflow example of Change Background (CB).}
    \label{fig:cb}
\end{figure*}

In the \textbf{Change Background (CB)} generation pattern, the process mainly consists of two steps: removing the original background and generating a new one. We construct this image-to-image process using the ComfyUI platform. During the new background generation stage, we use the caption of the original image as a condition for generation, for two main reasons: (1) it makes the generated images more natural, thereby reducing the need for manual quality control; and (2) the generated images are closer to real images, providing more challenging hard samples for model evaluation. As shown in Fig.~\ref{fig:cb}, we present an ComfyUI workflow example of Change Background.

\begin{figure*}[ht]
    \centering
    \includegraphics[width=1\linewidth]{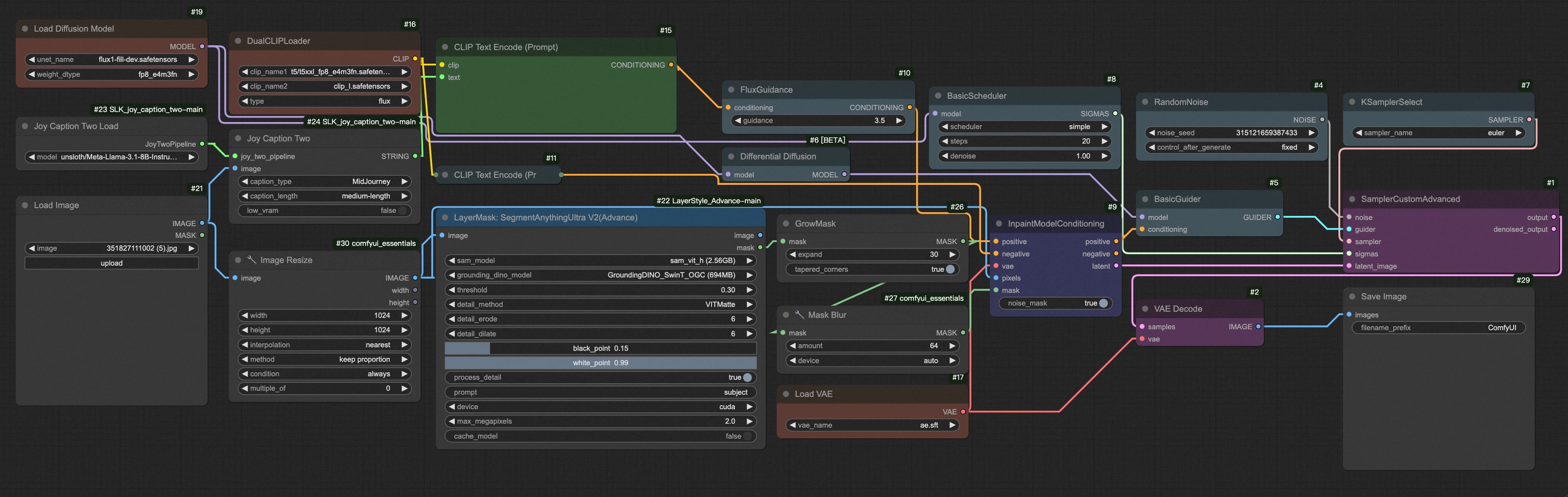}
    \caption{The ComfyUI workflow example of Segmentation Inpainting, a sub-pattern of Inpainting/Outpainting (IP\&OP).}
    \label{fig:ip}
\end{figure*}

For \textbf{Inpainting/Outpainting (IP\&OP)}, we design three sub-patterns: 1) Segmentation Inpainting: A segmentation model is used to locate the main object in the image and apply a mask. The masked region is then filled using an inpainting model. 2) Random Inpainting: A random size box is randomly selected from the central part of the image as mask area (It is does not exceed 30\% of the total image area). The masked region is subsequently restored using an inpainting model. 3) Outpainting: Taking the original image as the center, a region 1 to 1.5 times larger than the original is randomly extended and set as the mask. An outpainting model is then used to generate content for the extended area. For closed-source systems such as Wan21, we assemble the various modules using a Python interpreter. For open-source models like Flux.1-dev, we build the pipelines using the ComfyUI platform. Note that during the inpainting or outpainting process, the caption of the original image is used as a condition for generation. As shown in Fig.~\ref{fig:ip}, we present an ComfyUI workflow example of Segmentation Inpainting.

\begin{figure*}[ht]
    \centering
    \includegraphics[width=1\linewidth]{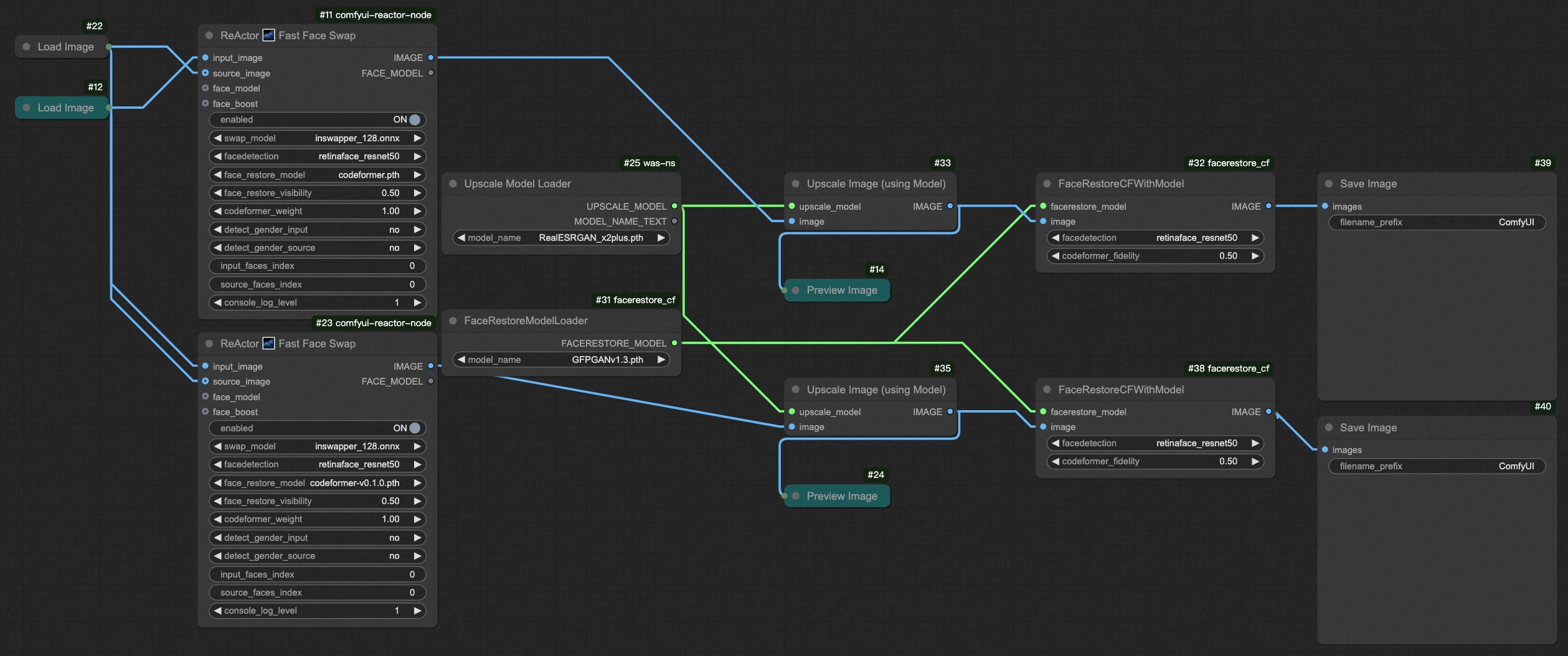}
    \caption{The ComfyUI workflow example of Face Swapping (FS).}
    \label{fig:fs}
\end{figure*}

For \textbf{Face Swapping (FS)}, we use the ComfyUI platform to construct an image-to-image process. Firstly, we use Qwen2.5-VL-32B-Instruct~\cite{bai2025qwen2} to filter the images from the Source $\mathbb{B}$, selecting those that contain human faces. The instruction is simply set as: ``If there are faces in the image, please return 1; otherwise, return 0. Please return the result in JSON format, for example: \{``contains\_face": 1\} or \{``contains\_face": 0\}." Then, in each process, the facial information between two original images is swapped, resulting in two new images. This pipeline mainly involves three steps: face detection, face swapping, and face restoration. As shown in Fig.~\ref{fig:fs}, we present an ComfyUI workflow example of Face Swapping.

\begin{figure*}[ht]
    \centering
    \includegraphics[width=1\linewidth]{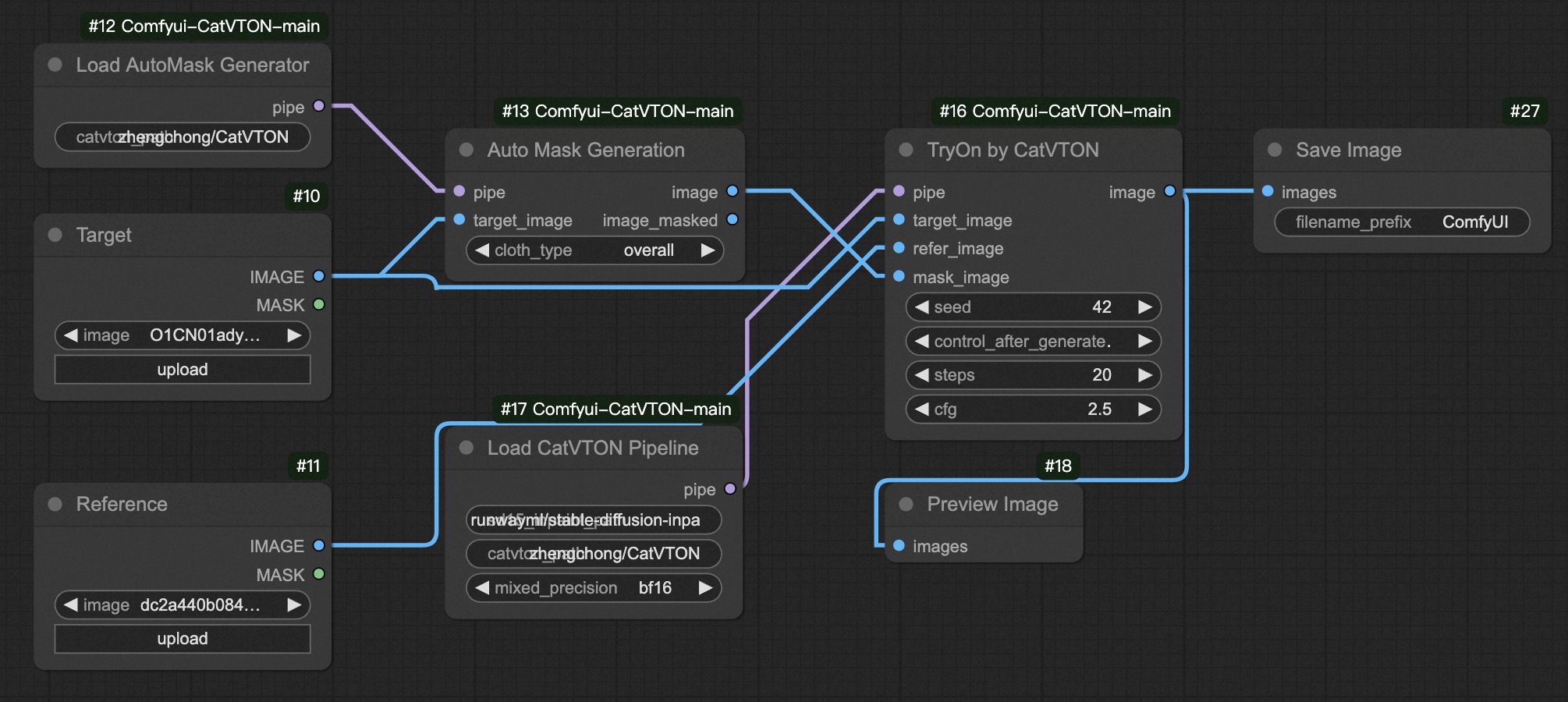}
    \caption{The ComfyUI workflow example of Virtual Try-ON (VTO).}
    \label{fig:vto}
\end{figure*}

\textbf{Virtual Try-On (VTO)} is defined as transferring clothing information from one image onto the model in another image. Similar to Face Swapping, we first use Qwen2.5-VL-32B-Instruct to select images containing human models from Source $\mathbb{B}$. Then, we build an image-to-image pipeline by using the ComfyUI platform. In each process, clothing information is exchanged between two model images, resulting in two new images. . This workflow mainly involves three steps: clothing segmentation, mask overlay, and local inpainting. As shown in Fig.~\ref{fig:vto}, we present an ComfyUI workflow example of Virtual Try-ON.

\begin{figure*}[ht]
    \centering
    \includegraphics[width=1\linewidth]{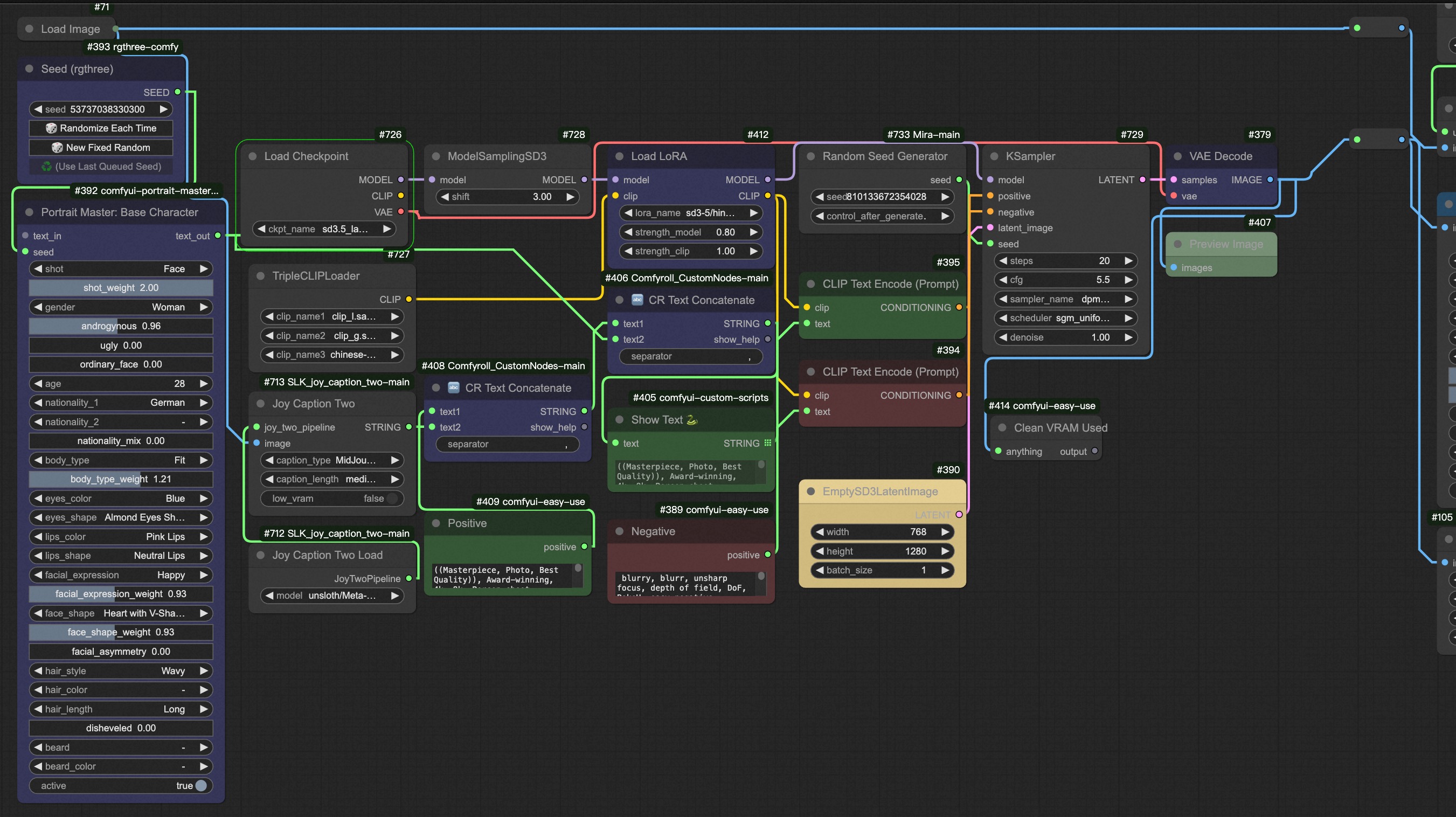}
    \includegraphics[width=1\linewidth]{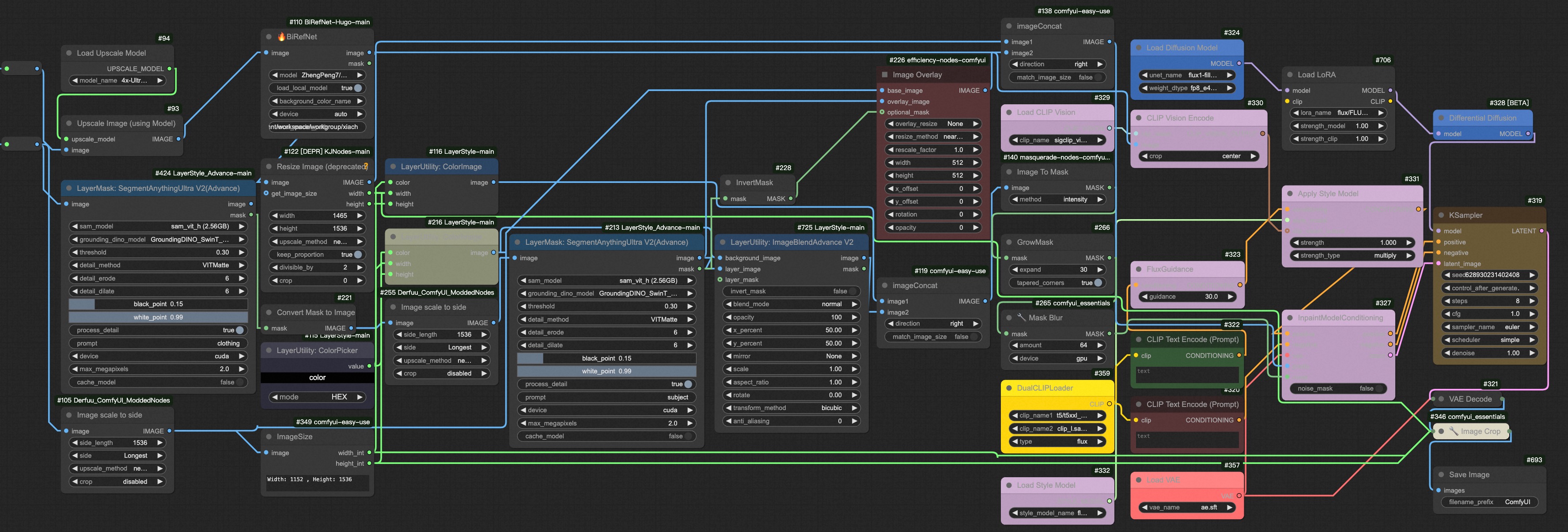}
    \caption{The ComfyUI workflow example of Realistic Model Generation (RMG).}
    \label{fig:rmg}
\end{figure*}

\textbf{Realistic Model Generation (RMG)} is defined as generating model images that closely resemble real-world scenarios. Unlike Virtual Try-On, which performs local clothing replacement on real images by using ComfyUI plat, RMG generates entirely new, full-body model images. Thus, the image distributions and application scenarios of the two generation patterns are quite different. The realism of RMG mainly involves two aspects: the realistic of faces and clothing. The LoRA~\cite{hu2022lora} model generally serves as an additional trainable lightweight network for specific base models (such as SD-XL and Flux.1-dev) to control certain attributes of the generated images. We select 30 LoRA models from the LibLib website that are highly faithful to real human faces, and after experimental screening, we retain 16 LoRA models. By combining the captions of origin image, the randomly chosen LoRA, and text-to-image generators, we are able to create model images with realistic faces. Secondly, similar to VTO, we use real clothing information from images to replace and generate the models’ clothing, thereby ensuring that the generated apparel more closely resembles real-world scenarios. As shown in Fig.~\ref{fig:rmg}, we present an ComfyUI workflow example of Realistic Model Generation.

\begin{figure*}[ht]
    \centering
    \includegraphics[width=1\linewidth]{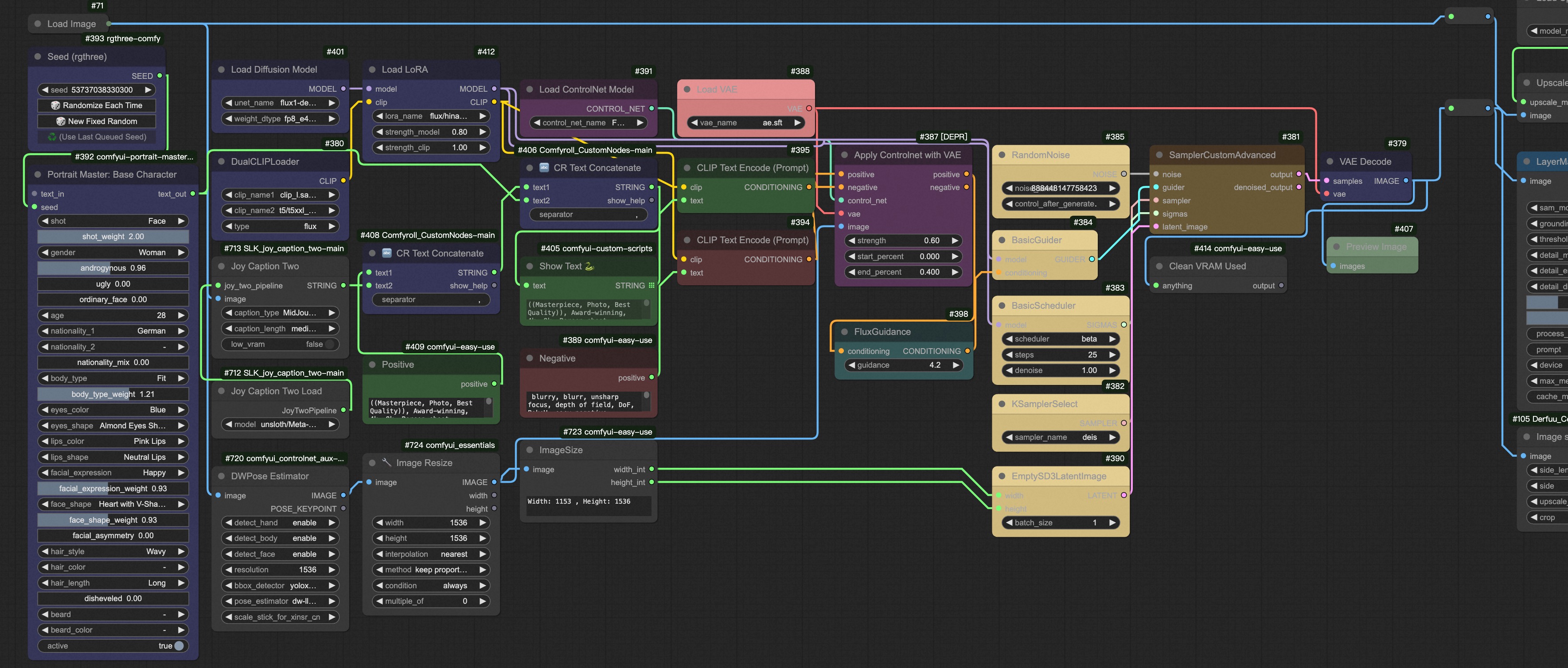}
    \includegraphics[width=1\linewidth]{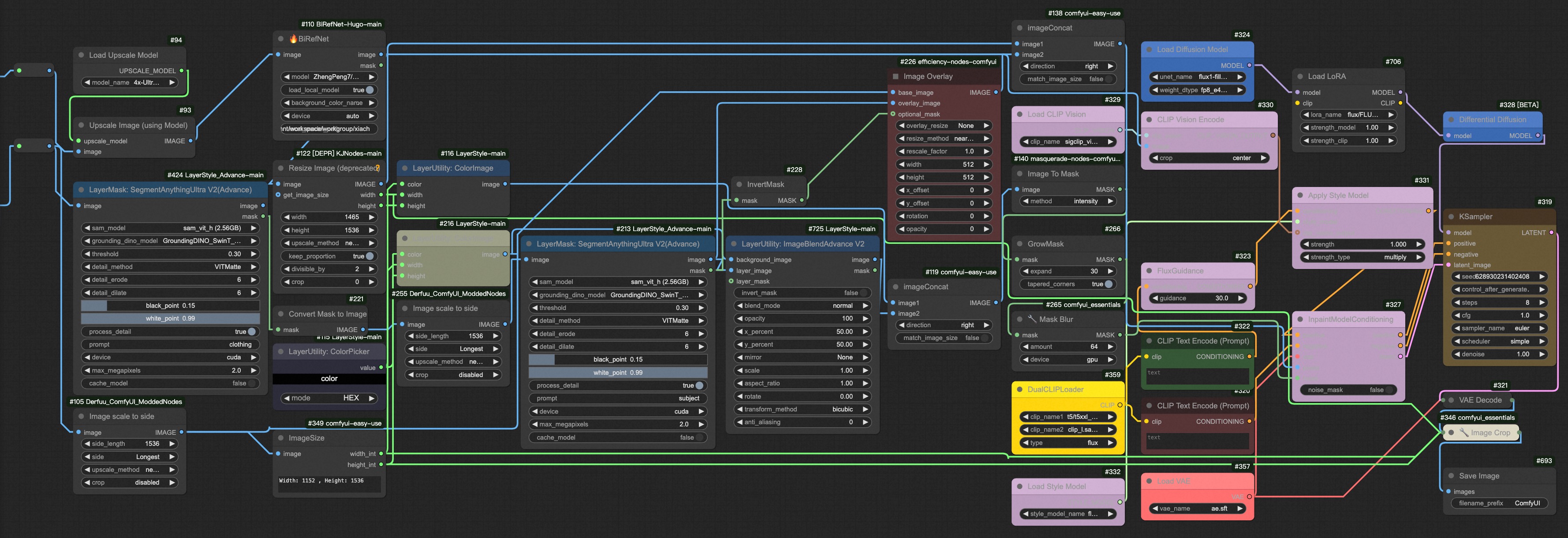}
    \caption{The ComfyUI workflow example of Pose-consistent Model Generation (PCMG).}
    \label{fig:pcmg}
\end{figure*}

\textbf{Pose-consistent Model Generation (PCMG)} introduces an additional pose control module based on image-to-image process in RMG. Specifically, pose information is first extracted from the original image using pose estimation, and then ControlNet~\cite{zhang2023adding}is utilized to generate images with pose-consistent models. As shown in Fig.~\ref{fig:pcmg}, we present an ComfyUI workflow example of Pose-consistent Model Generation.

In general, as shown in Tab.~\ref{table:appendix_models},we propose eight generation patterns in \mirage, which totally use 64 different models. In addition, multiple image processing modules are incorporated, such as upsampling modules, downsampling modules, and mask overlay modules. Besides, as shown in Fig.~\ref{fig:vs_images}, we present fake images corresponding to eight generation patterns in \mirage.

\paragraph{Data Distribution.} As detailed in the \mirage~dataset section of the main paper, we divide the \mirage~ dataset into a training set and a benchmark set. We show the ditribution of our {\mirage} dataset in Fig.~\ref{fig:mirage}. The left of the figure shows the distribution of the training set, and the right illustrates that of the test set (out benchmark).  As shown in Fig.~\ref{fig:mirage} (left), the \mirage~training set contains 2,000 samples in total, and we keep the ratio of real to fake images at 1:1. As shown in Fig.~\ref{fig:mirage} (right), our proposed benchmark contains 12087 images, covering a wide variety of generation patterns.

\section{Implementation Details}
\paragraph{Annotating Reasoning With VLMs.}
Similar to prior works~\cite{gao2025fakereasoning,zhang2025ivy}, we utilized powerful commercial VLMs to annotate the reasoning process of our training set. 

For the annotation VLM, we tried two well-known models, Gemini-2.5-Pro~\cite{comanici2025gemini} and QvQ-Max~\cite{QwenBlog2024QVQMax}. We save the widely-used GPT-4o~\footnote{\url{https://openai.com/index/hello-gpt-4o/}} for the VLM judge evaluating the explainability of our model. 

To guide the annotation process, we engineered a comprehensive instruction prompt, detailed in Fig.~\ref{list:instruction}. This prompt establishes the model's role as an expert, defines our criteria for in-the-wild AIGI, and outlines key areas for analysis inspired by~\cite{zhang2025ivy}. Crucially, for each analytical dimension (e.g., Lighting, Texture), we provide indicators for both ``real" and ``fake" signs to encourage a balanced and neutral perspective from the annotator VLM. This instruction serves as the preamble for all subsequent annotation prompts.

For the chain-of-thoughts steps, we refer to that in FakeReasoning~\cite{gao2025fakereasoning}, that is, observing the image content first, then check each applicable dimension defined in the instruction. 

Our primary goal was to generate data that emulates a reflective reasoning process, wherein a model assesses an initial judgment and then refines or corrects it. To achieve this, we explored two distinct annotation strategies:

\begin{itemize}
    \item \textbf{Strategy 1: Two-Turn Conversational Annotation.} Our initial approach involved a two-round dialogue. In the first turn, the VLM provided an initial judgment on whether the image was real or fake (Fig.~\ref{list:round1-1}-\ref{list:round1-2}). In the second turn, we supplied the ground-truth label and asked the model to reflect on its initial answer (Fig.~\ref{list:round2}). A key limitation of this method is that the quality of the reflection is highly dependent on the VLM's unconstrained and sometimes inconsistent initial judgment.
    
    \item \textbf{Strategy 2: Single-Turn Simulated Reflection.} To gain more control and generate higher-quality, targeted data, we developed a more effective single-turn strategy. In this approach, we generate two distinct annotations for each training sample. For both annotations, the model is provided with the ground-truth label. However, we prompt it to simulate two different initial states: one where its ``initial guess" was correct, and another where it was incorrect. The prompt for this strategy is shown in Fig.~\ref{list:strategy2-1-1}-\ref{list:strategy2-4-2}. This method offers a significant advantage: it grants us programmatic control over the training data, allowing us to explicitly create examples of self-correction from flawed initial heuristics. During training, we can then precisely tune the proportion of correct-to-incorrect initial pathways, systematically teaching our model this crucial reflective capability.
\end{itemize}

It is crucial to note that we employed a rejection sampling protocol to ensure the fidelity of the generated reasoning. Specifically, for annotations generated under Strategy 2 and in the second round of Strategy 1, we validated the model's final prediction against the ground-truth label. Any annotation where the VLM's final judgment did not match the ground truth was discarded, and the generation process was repeated for that sample until a correct prediction was achieved. This stringent quality control measure guarantees that all reasoning paths in our final training dataset culminate in a $100\%$ accurate conclusion.

\paragraph{SFT Training Details.} The cold-start Supervised Fine-Tuning stage of our model was implemented using the LLaMA-Factory framework\footnote{\url{https://github.com/hiyouga/LLaMA-Factory}}. The training data was prepared as follows: annotations from Strategy 1 were converted into a standard conversational format. For data from Strategy 2, we generated multiple dataset versions by systematically varying the proportion of samples with an ``incorrect" initial heuristic, ranging from $0\%$ (always correct) to $100\%$ (always incorrect). In proportions such as $30\%$, we generated a random number between $0$ and $1$ uniformly. When it is less than $0.3$, the wrong impression would be taken, otherwise the correct impression would be used. Note that to avoid bias introduced by the random sampling, we performed these sampling three times, and reported the results that ranked the second out of three. For our initial try, G-R-A, we kept the initial guess always correct. Therefore, the guess is always the same with the final answer. 

Training was conducted on a system with 8 NVIDIA H20 GPUs, running Python 3.10 on a Fedora OS. We configured a per-device batch size of 2 with a gradient accumulation factor of 4, resulting in an effective global batch size of 64. The learning rate was managed by a cosine annealing scheduler with an initial value of ${10}^{-4}$. Parameter-efficient fine-tuning was performed using LoRA with a rank of 256, an $\alpha$ of 512, and a dropout rate of 0.1. We only applied LoRA to the language models, and kept the vision encoder of the VLM frozen in all experiments. We limited the max new token from the model to be 4092 to avoid the out-of-memory problem. A repetition penalty of 1.2 was set. In addition, we kept the temperature of the VLM to be 0.3 to encourage the model to have a more certain answer. We used an Adam optimizer. The VLMs were quantized to bfloat16 to save memory.  

All experiments were initialized with a random seed of 42. The model was trained for 4,000 steps, and the final checkpoint was used for evaluation. Hyperparameters were determined using a held-out development set sampled from the training data. After finalizing the configuration, this development set was merged back into the training set for the final model training run.

To mitigate potential biases from resolution artifacts and to rigorously test the model's intrinsic robustness, we adopted a specific image handling strategy. During training, images larger than 448x448 were downsampled to this resolution, while smaller images retained their original size. Critically, no resizing was applied during inference. No other data augmentation techniques were applied. All ablation studies reported in this paper adhere to these same experimental settings to ensure a fair comparison.

In this stage, we used a prompt with the instruction in Fig.~\ref{list:instruction}. The prompt is presented in Fig.~\ref{list:sft}.

In our ablations in the main paper, we compared our model performance with other common types of reasoning in existing works. In Fig.~\ref{list:woreasoning} to \ref{list:gra}, we present the their prompts. 

\begin{figure*}[ht]
\centering
\begin{tcolorbox}[listing only, 
                  listing options={basicstyle=\ttfamily\footnotesize},
                  colback=gray!10, 
                  title=SFT-Prompt]
\verb|{Instruction}|

\# Task

\#\# Question

\verb|<image>| Is this image real or fake (AI-generated)? Think step by step following the \textbf{Thinking Steps} below:

\#\# Thinking Steps

a. \verb|<impression>|: Give a quick answer to the question (real/fake), which is your impression on the image at the first glance.

b. \verb|<reason>|: Output the reasons supporting your impression. The reasons should involve:
\begin{enumerate}
    \item Observation: The assistant describes the image content, especially unnatural details.
    \item Key Areas Evaluation: Analyze the image content from the key areas given by the user.
\end{enumerate}

c. \verb|<reflection>|: Reflective review your analysis above. Identify and correct the issues in \textbf{Observation} and \textbf{Key Areas Evaluation}, respectively. 

d. \verb|<answer>|: According to your reflective thinking, conclude with a final answer (real/fake). 

\#\# Output Format

\verb|<impression>| real/fake \verb|</impression>| \verb|<reason>|...\verb|</reason>| \verb|<reflection>|...\verb|</reflection>| \verb|<answer>| real/fake \verb|</answer>|    
\end{tcolorbox}

\caption{Model prompt in the SFT stage. }
\label{list:sft}
\end{figure*}

\begin{figure*}[ht]
\centering
\begin{tcolorbox}[listing only, 
                  listing options={basicstyle=\ttfamily\footnotesize},
                  colback=gray!10, 
                  title=01-Prompt]
\verb|{Instruction}|

\# Task

\#\# Question

\verb|<image>| Is this image real or fake (AI-generated)? Answer with ``real" or ``fake".
\end{tcolorbox}

\caption{Model prompt of ``w/o reasoning" in the ablations. }
\label{list:woreasoning}
\end{figure*}

\begin{figure*}[ht]
\centering
\begin{tcolorbox}[listing only, 
                  listing options={basicstyle=\ttfamily\footnotesize},
                  colback=gray!10, 
                  title=H-R-Prompt]
\verb|{Instruction}|

\# Task

\#\# Question

\verb|<image>|Is this image real or fake (AI-generated)? Answer with ``real" or ``fake", and give the reasons.

\#\# Output Requirements

- The assistant should give the reasons step-by-step, breaking down the task according to the \textbf{Thinking Steps}.

- The output format should be:\verb|<answer>|real/fake\verb|</answer>|\verb|<reason>|reasons\verb|</reason>|. 

\#\# Thinking Steps

The assistant first outlines the problem and the thinking steps.
\begin{itemize}
    \item \textbf{Observation}: the assistant describes the image content, especially unnatural details.
    \item \textbf{Analysis from different dimensions}: analyze the image content from the key areas given by the user.
    \item \textbf{Conclusion}: Real or Fake. 
\end{itemize}  
\end{tcolorbox}

\caption{Model prompt of ``H-R" in the ablations. }
\label{list:hr}
\end{figure*}

\begin{figure*}[ht]
\centering
\begin{tcolorbox}[listing only, 
                  listing options={basicstyle=\ttfamily\footnotesize},
                  colback=gray!10, 
                  title=A-R-Prompt]
\verb|{Instruction}| 

\# Task

\#\# Question

\verb|<image>|Is this image real or fake (AI-generated)? Think step by step and conclude with ``real" or ``fake". 

\#\# Output Requirements

- The assistant should reason through the question step-by-step, breaking down the task according to the \textbf{Thinking Steps} below to arrive at the final answer.

- The output format should be: \verb|<think>|thinking process\verb|</think>|
\verb|<answer>|real/fake\verb|</answer>|.

\#\# Thinking Steps

The assistant first outlines the problem and the thinking steps.
\begin{itemize}
    \item \textbf{Observation}: the assistant describes the image content, especially unnatural details.
    \item \textbf{Analysis from different dimensions}: analyze the image content from the key areas given by the user.
    \item \textbf{Conclusion}: Real or Fake. 
\end{itemize}  

\end{tcolorbox}

\caption{Model prompt of ``A-R" in the ablations. }
\label{list:ar}
\end{figure*}

\begin{figure*}[ht]
\centering
\begin{tcolorbox}[listing only, 
                  listing options={basicstyle=\ttfamily\footnotesize},
                  colback=gray!10, 
                  title=GRA-Prompt]
\verb|{Instruction}| 

\# Task

\#\# Question
\verb|<image>|Is this image real or fake (AI-generated)? First output your impression at the first glance (real/fake). Then, think step by step and conclude with a final answer (real/fake).

\#\# Output Requirements
- The assistant should examine the first impression, reason through the question step-by-step, breaking down the task according to the \textbf{Thinking Steps} below to arrive at the final answer.

- The output format should be: \verb|<impression>|real/fake\verb|</impression>|
\verb|<think>|thinking process\verb|</think>|
\verb|<answer>|real/fake\verb|</answer>|.  

\#\# Thinking Steps

The assistant first outlines the problem and the thinking steps.
\begin{itemize}
    \item \textbf{Observation}: the assistant describes the image content, especially unnatural details.
    \item \textbf{Analysis from different dimensions}: analyze the image content from the key areas given by the user.
    \item \textbf{Conclusion}: Real or Fake. 
\end{itemize}  

\end{tcolorbox}

\caption{Model prompt of G-R-A in the ablations. }
\label{list:gra}
\end{figure*}

\paragraph{RLVR Training Details. }For the RLVR stage, we employed the VLM-R1 framework\footnote{https://github.com/om-ai-lab/VLM-R1/tree/main/src}, initializing the model with the weights from the SFT stage. Our training leverages GRPO, an efficient policy optimization algorithm that circumvents the need for a separate, computationally expensive value function often required by methods like PPO.

The core idea of GRPO is to establish a dynamic baseline by using the average reward of multiple outputs sampled from the policy for the same input. For each training step, the model generates a group of $G$ candidate responses (where we set $G=4$ in our experiments). The advantage for each token is then calculated relative to the other responses within this group, obviating the need for an external value model. This approach is particularly well-suited for refining nuanced reasoning paths.

The training was conducted on 32 NVIDIA H20 GPUs, an increase in resources reflecting the higher computational intensity of RL training compared to SFT. The training set in this stage is the same with that in the SFT stage, but without reasoning annotations. The only annotation we used in this stage is the ``real/fake" annotation of the image. We trained the model for 50 steps with an effective global batch size of 128 (8 per-device with a gradient accumulation factor of 2). We used an Adam optimizer and a cosine annealing scheduler with an initial learning rate of ${10}^{-5}$. The LoRA rank was set to 8, with an alpha of 16 and a dropout rate of 0.1. We only applied LoRA to the language models, and kept the vision encoder of the VLM frozen in all experiments. The maximum number of new tokens for each generated response was capped at 2048. Random seed was set to 42 in all experiments. During training, the maximum resolution of the image was limited to 448x448, while in the inference time we used the original resolution. For the RLVR stage, we quantized the VLM to bfloat16 and used a sampling temperature of 0.95 to encourage a more calibrated token probability distribution. As our confidence-aware training adjusts the model's output probabilities, we re-calibrated the decision threshold for the initial answer, $A_i$. A new threshold of 0.58 for the ``fake" class was determined based on our development set. Furthermore, we implemented the adaptive reasoning trigger using prediction entropy. An entropy threshold of $\tau=0.96$, also tuned on the development set, was established. If the entropy of $A_i$ exceeds this value, the model engages in ``deep thinking" to produce a final answer $A_f$; otherwise, for high-confidence (low-entropy) predictions, the process is truncated, and $A_i$ is returned as the final output (and the explanations if needed).

In this stage, we employed the same prompt with the SFT stage, which is already given in Fig.~\ref{list:sft}.

As introduced in the main paper, our RLVR stage employs a composite reward function designed to cultivate specific desirable behaviors. In addition to the novel confidence-aware reward ($\mathcal{R}_{conf}$), our total reward signal integrates several other components:

\begin{itemize}
\item \textbf{Accuracy Reward ($\mathcal{R}_{acc}$):} This is a standard sparse reward that incentivizes correctness in the final answer ($A_f$). It is defined as:
\begin{equation}
\mathcal{R}_{acc} = \begin{cases} 1 & \text{if } A_f \text{ is correct} \\ 
0 & \text{otherwise} \end{cases}
\end{equation}

\item \textbf{Length Reward ($\mathcal{R}_{len}$):} To encourage the model to produce more detailed and comprehensive reasoning, we introduce a length-based reward. This reward is only granted for correct final answers to avoid rewarding verbose but incorrect responses. It is calculated using a cosine function to gently reward longer responses up to a soft cap:
\begin{equation}
    \mathcal{R}_{len} = \begin{cases} \frac{1 - \cos(\frac{l\pi}{512})}{2} & \text{if } A_f \text{ is correct} \\ 0 & \text{otherwise} \end{cases}
\end{equation}
where $l$ is the number of tokens in the generated response. 512 in the equation is the expected response length set by us. 

\item \textbf{Format Reward ($\mathcal{R}_{fmt}$):} To ensure the model adheres to the specified output structure (as shown in Fig.~\ref{list:sft}), we use a binary format reward. A reward of 1 is given for correctly formatted responses, and 0 otherwise.

\item \textbf{Repetition Penalty ($\mathcal{R}_{rep}$):} To discourage repetitive and uninformative text, we apply a penalty based on the number of n-gram repetitions. Specifically, we count the occurrences ($n$) of repeated 16-grams and apply a penalty:
\begin{equation}
    \mathcal{R}_{rep} = -\frac{n}{20}
\end{equation}
\end{itemize}

These components are combined to form the final reward signal used in the GRPO algorithm. It is important to note that GRPO's objective function also includes a KL divergence term for regularization against a reference policy. For a detailed formulation of this term, we refer readers to the original GRPO paper~\cite{shao2024deepseekmath}.

In experiments, we simply added all these rewards together. We acknowledge that setting proper weights on each component might lead to a better performance. In the present setting, our model already outperforms all the state-of-the-art methods. 

\paragraph{Training Details of Baselines.} For all the baselines in our paper, i.e., the specialized AIGI detectors in Table 2 in the main paper, we use a same setting with our model for a fair comparison. That is, all the baselines were trained for 13 epochs with image resized to 448x448. Some methods, such as CNNSpot~\cite{cnn-spot} and NPR~\cite{npr}, support images with a resolution of 448×448. Others, such as UnivFD~\cite{univfd}, which is based on CLIP models pre-trained on images of size 224×224, employ center cropping on images. For the VLMs in Table 2 in the main paper, we trained them with the same hyperparameters with our model on prompt in Fig.~\ref{list:woreasoning}. During the inference, threshold of 0.5 for ``fake" class was determined based on previous works.

\section{Additional Experiments}
We involve more experiment results in this section, especially the evaluation of explainability and efficiency that we do not have space to put in the main paper. We also show more ablations to validate the effectiveness of our propose method.

\paragraph{Selection on Backbone Models. }In Table 2. in the main paper, we already compared the performance of Qwen-VL-2.5-3B, Qwen-VL-2.5-7B, and LLaMA3.2-Vision-11B on our benchmark. We simply chose the model with the best performance, Qwen-VL-2.5-7B, as our backbone model. 

\paragraph{Evaluation of Explainability. }As previously established, explainability is a vital attribute for in-the-wild AIGI detectors. In this section, we present a quantitative evaluation of the reasoning quality generated by our model. Specifically, we define the ``explanation" as the textual content produced within the \verb|<reason>| tags of our model's output.

We benchmarked the explanation quality of our model against that of Gemini-2.5 Pro, QvQ-Max and GPT-o3~\footnote{https://openai.com/index/introducing-o3-and-o4-mini/}. 
We employed GPT-4o as an impartial, automated judge. For each explanation, GPT-4o was prompted to assign a quality score on a scale from 0 to 10, based on criteria such as correctness, detail, and relevance, as detailed in the prompt shown in Fig.~\ref{listing:explain}. This evaluation was conducted on the ID subset of our {\mirage} benchmark to assess performance in diverse scenarios.

The results, summarized in Table~\ref{table:explain}, demonstrate that the explanations generated by {\model} are of significantly higher quality than those from the baseline model. To provide further qualitative insight, we include several illustrative examples of our model's explanations at the end of the appendix.

\begin{figure*}[ht]
\centering
\begin{tcolorbox}[listing only, 
                  listing options={basicstyle=\ttfamily\footnotesize},
                  colback=gray!10, 
                  title=LLM As Judge]
\verb|{Instruction}|

\# Task

\#\# Question

\verb|<image>|This is a \verb|{Ground Truth}| image. And here is an explanation for why this image is \verb|{Ground Truth}|. Please review the image as well as the explanation, and give it a score from 0 to 10. The score should be rated according to the validity, clarity, completeness, and briefness of the explanation.  

\#\# Output Requirements

- The assistant should think step by step, and give a score between 0 and 10.

- The output format should only be a score from 0 to 10. 

- If the answer is not correct (e.g., the model predict real for the fake image), the score should be 0. 

\end{tcolorbox}
\caption{Prompts of the VLM judge (gpt-4o) in the explainability evaluation. }
\label{listing:explain}
\end{figure*}

\begin{table}[ht]
    \centering\begin{tabular}{c|cc}
    \toprule
    Method & Score\\
    \midrule
    Gemini2.5-Pro & 6.56\\
    QvQ-Max & 4.39 \\
    GPT-o3 & 8.46 \\
    \midrule
    \model & 8.81 \\
    \bottomrule    
    \end{tabular}
     \caption{Explanations scores of different models on the ID subset of {\mirage}. }
    \label{table:explain}
\end{table}

\paragraph{Evaluation of Efficiency. }As previously established, efficiency is a critical attribute for the practical deployment of in-the-wild AIGI detectors. We conducted a comparative analysis of inference latency across several reasoning strategies: ``without reasoning," heuristic-reasoning (H-R), analytic-reasoning (A-R), our initial try guess-reasoning-answer (G-R-A), and our adaptive Heuristic-to-Analytic (HA-R) approach. The evaluation was performed on our OOD-C test set using a single NVIDIA H20 GPU with a batch size of 6.

The results, summarized in Table~\ref{table:inference}, highlight the superior trade-off achieved by our model. For multi-stage models, we report latency for different operational modes. Our adaptive HA-R approach achieves the highest accuracy (78.46\%) among all variants while maintaining a lower latency than fully analytic methods like A-R.

Notably, a comparison between our adaptive HA-R and a non-adaptive variant that always performs deep analysis (denoted as HA-R-$A_f$) reveals a compelling finding: our adaptive model is not only significantly faster but also more accurate. This suggests that by judiciously trusting its high-confidence initial judgments, our model mitigates the risk of ``over-thinking" — a phenomenon where forcing deeper, more complex reasoning on straightforward cases can paradoxically introduce errors. Our adaptive mechanism thus provides a dual benefit of improved speed and enhanced robustness.

\begin{table}[ht]
    \centering\begin{tabular}{c|cc}
    \toprule
    Method & Inference Time (s) & Acc. (\%) \\
    \midrule
    w/o Reasoning & 1.35 & 69.07\\
    H-R & 3.28 & 72.36 \\
    A-R & 65.65 & 73.13 \\
    \midrule
    G-R-A-Guess & 3.39 & 75.25 \\
    G-R-A-Answer & 70.12 & 75.25 \\
    \midrule
    HA-R-$A_i$ & 3.57 & 76.63 \\
    HA-R-$A_f$ & 77.46 & 78.13\\
    HA-R & 10.77 & \textbf{78.46} \\
    \bottomrule    
    \end{tabular}
     \caption{Average inference time of a batch of images from the OOD-C subset of {\mirage}. The time was counted in second(s). }
    \label{table:inference}
\end{table}

\begin{figure}[ht]
    \centering
    \includegraphics[width=\linewidth]{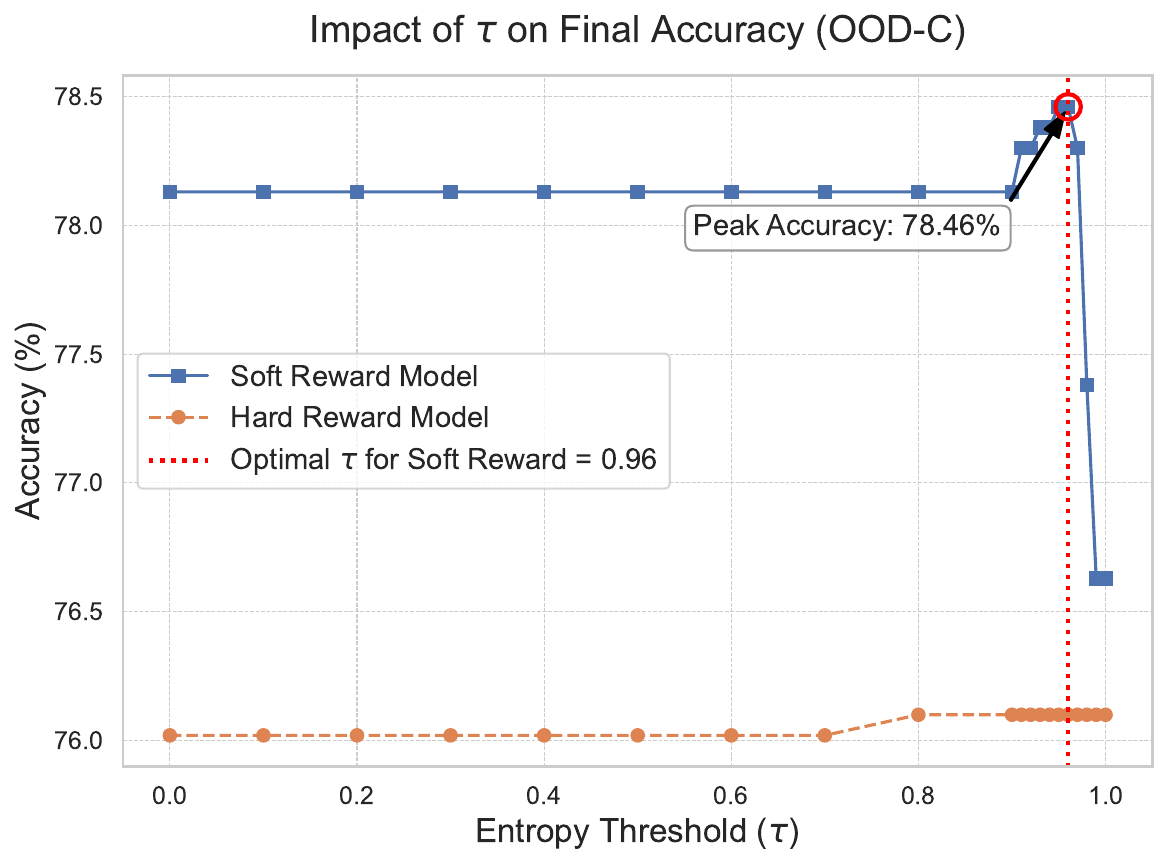}
    \caption{Ablations on $\tau$. The x-axis shows different values of $\tau$, the left y-axis shows the model accuracy on the OOD-C subset of {\mirage}. Here, ``soft reward" means the confidence-aware reward used in our \model. ``hard reward" refers to the original binary reward. They were applied to $A_i$ during the GRPO training. }
    \label{fig:tau1}
\end{figure}

\begin{figure}[ht]
    \centering
    \includegraphics[width=\linewidth]{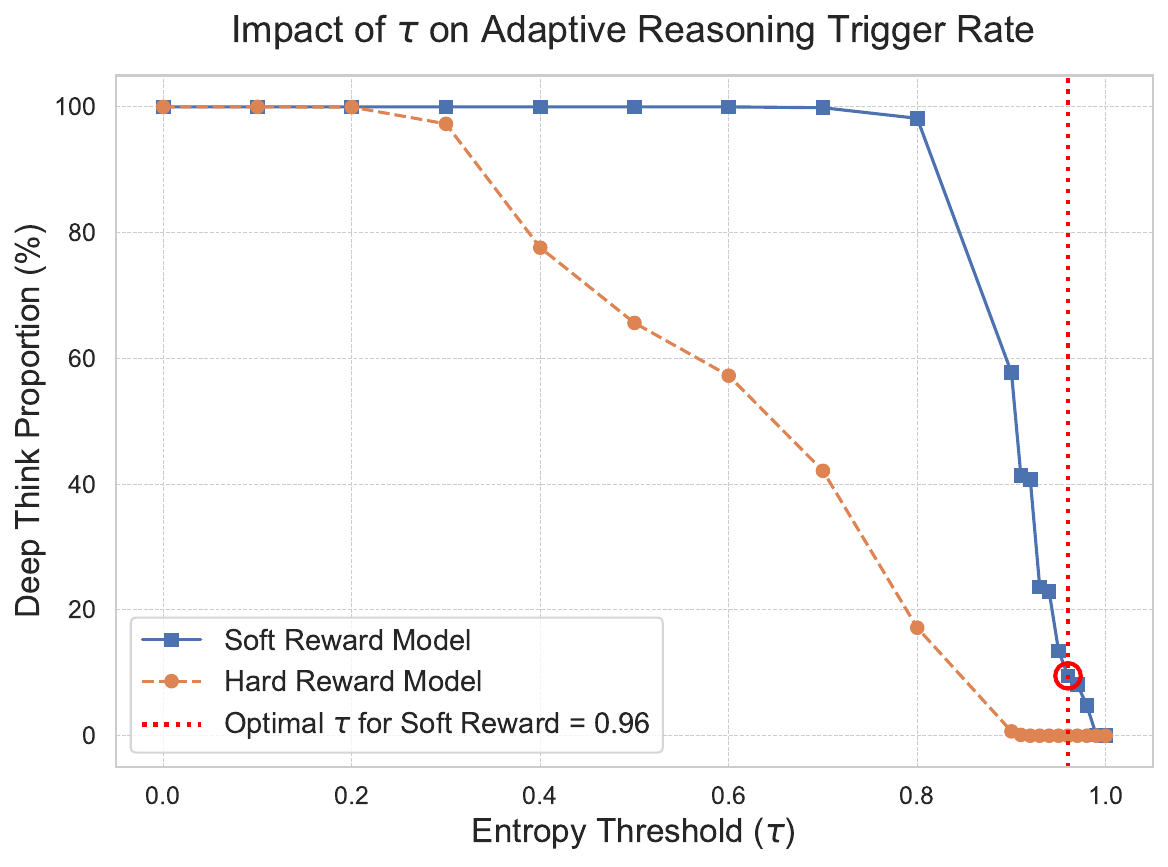}
    \caption{Ablations on $\tau$. The x-axis shows different values of $\tau$, the left y-axis shows the proportion of samples activating ``deep think" on the OOD-C subset of {\mirage}. }
    \label{fig:tau2}
\end{figure}

\paragraph{Ablations on $\tau$. }We conducted an ablation study to determine the optimal entropy threshold, $\tau$, which governs our model's adaptive reasoning mechanism. Figures~\ref{fig:tau1} and \ref{fig:tau2} illustrate the trade-off between inference efficiency and accuracy on the OOD-C set for both our proposed ``Soft Reward" model and a ``Hard Reward" (the original binary reward in RLVR) baseline.

For our primary model (``Soft Reward"), the results demonstrate a well-defined optimal point. As shown in Fig.~\ref{fig:tau1}, increasing $\tau$ reduces the proportion of samples that trigger ``deep think," which serves as a direct proxy for computational cost and latency. Concurrently, Fig.~\ref{fig:tau2} reveals that accuracy peaks at a specific threshold before declining. The optimal value is identified as $\tau=0.96$, where the model achieves its highest accuracy of 78.46\%. Remarkably, at this peak performance, the model only needs to activate its costly deep reasoning path for a small fraction of challenging cases (around 9.5\%).

This behavior contrasts sharply with the ``Hard Reward" baseline. The baseline's accuracy remains flat and significantly lower, indicating its inability to leverage the adaptive mechanism to improve performance. Its ``deep think" rate also plummets rapidly, suggesting poorly calibrated confidence.

In conclusion, this study validates that our confidence-aware training successfully endows the model with a nuanced understanding of its own uncertainty. This allows it to judiciously balance effectiveness and efficiency, achieving state-of-the-art accuracy while minimizing computational overhead — a critical capability for practical, in-the-wild deployment.

\paragraph{Ablations on Model Size. }To investigate the impact of model scale on performance, we compared our primary 7B backbone model (Qwen-VL-2.5-7B) with its smaller 3B counterpart. As shown in Table~\ref{table:ablation_size}, the 7B model demonstrates significantly stronger generalization capabilities, particularly on the more challenging OOD-C subset.

We attribute this performance gap primarily to the difference in the models' effective context lengths. Our training methodology, encompassing both SFT with detailed instructions and the subsequent RLVR stage, relies on generating and processing long, complex CoT reasoning sequences. The 7B model's larger context window allows it to fully accommodate and learn from these extensive reasoning paths. In contrast, the 3B model's limited context window acts as a critical bottleneck, preventing it from fully leveraging the rich information contained in the long-form training data. Consequently, its ability to master the reflective reasoning process is fundamentally constrained.

\begin{table}[ht]
    \centering
    \resizebox{\columnwidth}{!}{\begin{tabular}{c|cccccc}
    \toprule
    \multirow{2}{*}{\textbf{Model Size}}& \multicolumn{3}{c}{\textbf{ID}}&\multicolumn{3}{c}{\textbf{OOD-C}} \\
    &\small ACC.&\small P.&\small R.&\small ACC.&\small P.&\small R.\\
    \midrule
    3B & 96.02 & 97.78 & 94.57 & 69.73 & 96.91 & 41.22\\
    7B & \textbf{97.26} & 96.45 & 98.41 & \textbf{78.46} & 96.30 & 59.87\\
    \bottomrule    
    \end{tabular}}
     \caption{Ablations on the size of the backbone model. }
    \label{table:ablation_size}
\end{table}

\paragraph{Ablation on SFT Temperature.}
We investigated the impact of the sampling temperature used during the SFT stage on the model's final performance. A series of models were trained with different temperature settings, and their performance was evaluated on the ID and OOD-C subsets post-SFT (i.e., before the RLVR stage).

The results, presented in Table~\ref{table:ablation_temp}, reveal a clear trend: a moderately low temperature of 0.3 yields the optimal performance, particularly for OOD generalization. Both very low (0.1) and higher temperatures ($\geq$0.5) lead to a noticeable degradation in performance.

This outcome can be explained by the fundamental trade-off governed by temperature in language generation. A very low temperature (e.g., 0.1) leads to highly deterministic and repetitive outputs, causing the model to learn a narrow, less diverse set of reasoning patterns from the training data. This lack of diversity hinders its ability to generalize. Conversely, high temperatures (e.g., $\geq$0.5) introduce excessive randomness, potentially causing the model to generate noisy or less coherent reasoning paths during training. While this promotes diversity, it can also disrupt the learning of structured, logical thinking.

The optimal temperature of 0.3 appears to strike the ideal balance. It allows for sufficient diversity in the generated reasoning to foster robust generalization, while maintaining enough structure and fidelity to the training data to ensure the model learns coherent and effective problem-solving strategies.

\begin{table}[ht]
    \centering
    \resizebox{\columnwidth}{!}{\begin{tabular}{c|cccccc}
    \toprule
    \multirow{2}{*}{\textbf{Proportion}}& \multicolumn{3}{c}{\textbf{ID}}&\multicolumn{3}{c}{\textbf{OOD-C}} \\
    &\small ACC.&\small P.&\small R.&\small ACC.&\small P.&\small R.\\
    \midrule
    0.1 &  96.36 &  96.06 & 97.05 & 73.63 & 95.34 & 50.41 \\
    0.3 & \textbf{96.84} & 97.37 & 96.60 &\textbf{ 76.02} & 97.62 & 53.95 \\
    0.5 & 95.78 & 96.67 & 95.25 & 69.15 & 97.40 & 41.05 \\
    0.7 & 96.14 &  98.46 & 94.12 & 70.06 & 97.68 & 41.54 \\
    1 & 95.90 & 97.44 & 94.68 & 71.63 & 98.19 & 44.50\\
    \bottomrule    
    \end{tabular}}
     \caption{Ablations on the model temperature in SFT. }
    \label{table:ablation_temp}
\end{table}

\paragraph{Ablations on Annotation Strategy. }We conducted an ablation study to compare the efficacy of our two proposed reasoning annotation strategies. The results, presented in Table~\ref{table:ablation_annotation}, unequivocally demonstrate the superiority of our single-turn simulated reflection approach (Strategy 2). Note that in the table we reported the model performance in the SFT stage without RLVR. 

While both strategies yield strong performance on the ID test set, Strategy 2 establishes a commanding lead on the more challenging OOD-C subset, improving accuracy by 3 percentage points and, most critically, boosting recall by around 7 points. This indicates that models trained on Strategy 2 data are significantly more capable of identifying out-of-distribution fakes.

We attribute this performance gap to the fundamental difference in how the reasoning data is generated. Strategy 1, the two-turn conversational approach, is constrained by the annotator VLM's initial, unguided judgment. If the VLM's initial reasoning is weak or flawed, the subsequent reflection is built upon a shaky foundation, limiting the quality of the final training sample. In contrast, Strategy 2 provides us with programmatic control over the learning signal. By explicitly forcing the model to generate reasoning paths from both correct and incorrect initial heuristics, we create a more diverse and targeted training dataset. This dataset systematically teaches the model how to recover from errors and how to verify initial intuitions—crucial skills for robust generalization. Therefore, the results confirm that the controlled generation of reflective reasoning paths in Strategy 2 is a more effective method for cultivating a model's ability to generalize to unseen, in-the-wild scenarios.

\begin{table}[ht]
    \centering
    \resizebox{\columnwidth}{!}{\begin{tabular}{c|cccccc}
    \toprule
    \multirow{2}{*}{\textbf{Proportion}}& \multicolumn{3}{c}{\textbf{ID}}&\multicolumn{3}{c}{\textbf{OOD-C}} \\
    &\small ACC.&\small P.&\small R.&\small ACC.&\small P.&\small R.\\
    \midrule
    Strategy 1 & 96.24 & 97.12 & 95.68 & 71.46 & 94.94 & 46.63 \\
    Strategy 2 & \textbf{96.84} & 97.37 & 96.60 & \textbf{76.02} & 97.62 & 53.95 \\
    \bottomrule    
    \end{tabular}}
     \caption{Ablations on the annotation strategy for annotating the VLM reasoning. }
    \label{table:ablation_annotation}
\end{table}

\paragraph{Ablations on Annotation VLM. }We conducted an ablation study to assess the impact of the choice of annotation VLM on our final model's performance. We trained two separate models using reasoning data generated by either Gemini-2.5 Pro or QvQ-Max and evaluated them on the ID and OOD-C subsets. We reported the model performance in the SFT stage. 

The results, presented in Table~\ref{table:ablation_annot_model}, reveal a surprising outcome: the model trained on annotations from QvQ-Max significantly outperforms the one trained on Gemini-2.5 Pro's annotation, especially in OOD generalization. This finding appears to contradict the zero-shot performance results in Table 2 in the main paper, where Gemini-2.5 Pro proved to be the stronger zero-shot detector. We hypothesize that this reversal is attributable to two primary factors:
\begin{itemize}
    \item \textit{Task Misalignment with Zero-Shot Capabilities}: Our annotation task is fundamentally different from a zero-shot detection task. During annotation, the VLM is provided with the ground-truth label and tasked with generating a plausible reasoning path to justify it. This is a task of post-hoc rationalization rather than genuine detection. It is plausible that while Gemini-2.5 Pro excels at detection, QvQ-Max may be more adept at generating high-quality, structured explanations when the outcome is already known.
    \item \textit{Architectural Homogeneity}: Our backbone model is Qwen-VL-7B, which shares almost the same architectural family as the QvQ-Max annotation model. This architectural alignment may create a form of ``self-alignment," where the reasoning style, tokenization patterns, and latent biases of the QvQ-Max annotator are more readily and effectively assimilated by the Qwen-VL-7B student model. This results in a more efficient knowledge transfer compared to training on data from a model with a disparate architecture like Gemini.
\end{itemize}
Therefore, we conclude that for our training methodology, the optimal annotation VLM is not necessarily the best zero-shot performer, but one that both excels at generating post-hoc rationales and aligns well with the student model's architecture.

\begin{table}[ht]
    \centering
    \resizebox{\columnwidth}{!}{\begin{tabular}{c|cccccc}
    \toprule
    \multirow{2}{*}{\textbf{Proportion}}& \multicolumn{3}{c}{\textbf{ID}}&\multicolumn{3}{c}{\textbf{OOD-C}} \\
    &\small ACC.&\small P.&\small R.&\small ACC.&\small P.&\small R.\\
    \midrule
    Gemini2.5-Pro & 96.08 & 97.12 & 95.36 & 71.38 & 95.50 & 45.32 \\
    QvQ-Max & \textbf{96.84} & 97.37 & 96.60 & \textbf{76.02} & 97.62 & 53.95 \\
    \bottomrule    
    \end{tabular}}
     \caption{Ablations on the annotation model for annotating the VLM reasoning. }
    \label{table:ablation_annot_model}
\end{table}

\paragraph{Ablations on HA-R Training Set Proportion. }
A cornerstone of our methodology is teaching the model to perform reflective reasoning by learning from its own simulated initial mistakes. We investigated the optimal proportion of training samples where the initial heuristic answer ($A_i$) is deliberately set to be incorrect. This ablation was conducted at the SFT stage, and the results are presented in Table~\ref{table:ablation_proportion}.

The data reveals a non-trivial relationship between this proportion and model performance. The optimal OOD generalization is achieved when 70\% of the training samples feature an incorrect initial heuristic. Performance degrades when this proportion is either too low or too high.

This finding highlights a critical aspect of learning reflective skills:
\begin{itemize}
    \item \textit{Low Proportion of Errors (e.g., $\leq$50\%)}: When the model is predominantly exposed to examples where its initial guess is correct, it receives insufficient training on the crucial skill of self-correction. It learns how to rationalize correct answers but does not adequately learn how to identify and recover from its own flawed initial judgments. This leads to brittle performance when faced with challenging OOD cases where its initial heuristics are more likely to fail.
    \item \textit{High Proportion of Errors (e.g., $\geq$90\%)}: Conversely, if the model is almost always told its initial guess is wrong, it may learn to systematically distrust its own heuristics. This can lead to ``over-thinking" even on simple cases, or it may fail to develop a reliable heuristic function in the first place, impairing both efficiency and accuracy.
    \item \textit{Optimal Proportion (70\%)}: A proportion of 70\% appears to provide the ideal curriculum. It ensures the model is rigorously trained on self-correction pathways, forcing it to develop robust analytical skills. At the same time, it is still exposed to enough ``correct first-guess" scenarios to learn to trust its heuristics when they are reliable. This balanced diet of success and failure is key to cultivating a genuinely adaptive and robust reasoning mechanism.
\end{itemize}

Therefore, we conclude that explicitly and frequently (but not too frequently) training the model to recover from initial errors is essential for building its generalization capabilities for in-the-wild scenarios.

\begin{table}[ht]
    \centering
    \resizebox{\columnwidth}{!}{\begin{tabular}{c|cccccc}
    \toprule
    \multirow{2}{*}{\textbf{Proportion}}& \multicolumn{3}{c}{\textbf{ID}}&\multicolumn{3}{c}{\textbf{OOD-C}} \\
    &\small ACC.&\small P.&\small R.&\small ACC.&\small P.&\small R.\\
    \midrule
    0 &  96.38 & 97.46 & 95.59 & 75.10 & 97.53 & 51.89 \\
    0.1 & 95.60 & 95.50  & 96.15 & 71.83 & 96.21 & 45.89 \\
    0.3 & 96.12 & 96.05 & 96.59 & 75.04 & 95.85 & 53.04\\
    0.5 & 96.79 & 96.42 & 97.51 & 75.27 & 97.26 & 52.38 \\
    0.7 & \textbf{96.84} & 97.37 & 96.60 & \textbf{76.02} & 97.62 & 53.95 \\
    0.9 & 95.60 & 95.50  & 96.15 & 71.83 & 96.21 & 45.89 \\
    1 & 96.07 & 98.81 & 93.66 & 72.87 & 96.07 & 48.11 \\
    \bottomrule    
    \end{tabular}}
     \caption{Ablations on the proportion of error impression in the SFT training set. }
    \label{table:ablation_proportion}
\end{table}

\paragraph{Ablations on Progressive Two-Stage Training. }To validate the effectiveness of our progressive two-stage training pipeline, we conducted a comprehensive ablation study, incrementally adding each key component. The performance of each configuration was evaluated on our ID and OOD-C test sets, with results summarized in Table~\ref{table:ablation_grpo}.

The results present a clear, stepwise improvement, underscoring the contribution of each stage:
\begin{itemize}
    \item \textit{GRPO-only v.s. SFT-only}: Training with either GRPO or SFT alone yields reasonable but suboptimal performance. Notably, the SFT-only model slightly outperforms the GRPO-only model on the OOD set, suggesting that a strong initial alignment with structured reasoning (provided by SFT) is a more crucial foundation than reinforcement learning from scratch.
    \item \textit{Adding RLVR (SFT+Hard GRPO)}: Building upon the SFT checkpoint with a standard GRPO objective (using a binary accuracy reward for $A_i$, denoted as ``Hard GRPO") significantly boosts performance, especially OOD accuracy, which rises from 76.02\% to 77.44\%. This confirms that the RL stage effectively enhances the model's reasoning capabilities beyond what supervised mimicry can achieve.
    \item \textit{Introducing the Confidence Reward (SFT+GRPO)}: The final and most critical step is replacing the ``hard" binary reward with our proposed ``soft" confidence-aware reward. This single change provides another substantial performance gain, pushing the OOD accuracy to a final 78.46\%. This demonstrates that explicitly rewarding the model for calibrated confidence in its initial heuristic ($A_i$) is key to unlocking its full generalization potential.
\end{itemize}

Furthermore, we observed that the RLVR stage consistently increased the average length of the model's reasoning from approximately 560 to over 640 tokens, indicating the generation of more detailed and thorough explanations. In conclusion, this ablation unequivocally validates our design: the SFT stage provides an essential foundation, the GRPO stage refines reasoning, and our novel confidence reward is the crucial ingredient that enables robust, adaptive performance in the wild.

\begin{table}[ht]
    \centering
    \resizebox{\columnwidth}{!}{\begin{tabular}{c|cccccc}
    \toprule
    \multirow{2}{*}{\textbf{Method}}& \multicolumn{3}{c}{\textbf{ID}}&\multicolumn{3}{c}{\textbf{OOD-C}} \\
    &\small ACC.&\small P.&\small R.&\small ACC.&\small P.&\small R.\\
    \midrule
    GRPO-only & 96.36 & 95.16 & 98.07 & 75.10 & 96.12 & 52.96\\
    SFT-only & 96.84 & 97.37 & 96.60 & 76.02 & 97.62 & 53.95 \\
    SFT+Hard GRPO & 97.02 & 96.64 & 97.73 & 77.44 & 95.91 & 57.89\\
    SFT+GRPO & \textbf{97.26} & 96.45 & 98.41 & \textbf{78.46 }& 96.30 & 59.87\\
    \bottomrule    
    \end{tabular}}
     \caption{Ablation study on the components of our progressive two-stage training pipeline. We compare SFT-only, GRPO-only, and their combination. ``Hard GRPO" uses a standard binary accuracy reward, while our full model (``SFT+GRPO") incorporates the proposed confidence-aware reward.}
    \label{table:ablation_grpo}
\end{table}

\section{Example Visualization}
This section provides qualitative examples to illustrate the reasoning processes of different models and validate our design choices.

\begin{figure*}[ht]
\centering
\begin{tcolorbox}[listing only, 
                  listing options={basicstyle=\ttfamily\footnotesize},
                  colback=gray!10, 
                  title=Reasoning Examples]
    \begin{center}
        \includegraphics[width=0.4\linewidth]{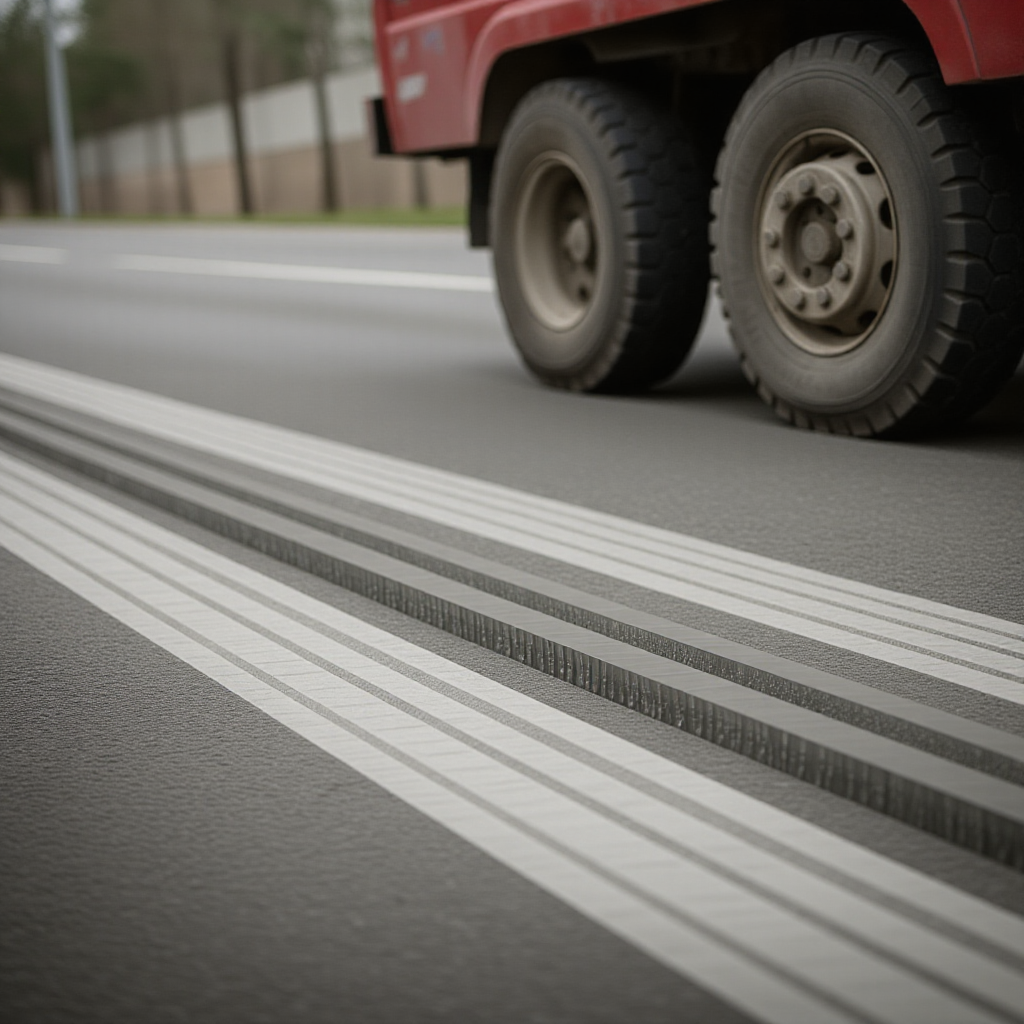}
    \end{center}
    
    \verb|<answer>|
    {\textcolor{red}{\textbf{real}}}
    \verb|</answer>|

    \verb|<reason>|
    
    - \textbf{Observation}: The image depicts a close-up view of a red truck's rear wheels parked on a paved road. The foreground features neatly painted white horizontal stripes on the asphalt, creating a textured pattern. The truck's tires appear robust, with visible treads and metallic hubcaps. The background shows a blurred urban landscape with sparse trees and a low wall, suggesting a shallow depth of field. The lighting seems even, likely from ambient daylight, casting subtle shadows beneath the wheels. No obvious distortions or anomalies are immediately apparent.  
    
    - \textbf{Key Areas Evaluation}: 
    \begin{itemize}
        \item Lighting and Geometry: The lighting appears consistent, with no conflicting shadow directions. The perspective of the truck aligns naturally with the road, indicating proper geometric alignment.
        \item Structure and Physics: The truck’s wheels are proportionally correct, seated realistically on the pavement. The spacing between the striped markings on the road follows a logical pattern, adhering to real-world construction standards.
        \item Focus and Depth of Field: The foreground (truck/wheels) is sharply focused, while the background is smoothly blurred, mimicking a camera’s shallow depth of field. Edges do not exhibit artificial cutoffs or digital artifacts.
        \item Material and Texture: The tire treads display detailed grooves, and the asphalt exhibits a gritty texture typical of concrete. The white paint on the stripes looks smooth but plausible for road marking.
        \item Color: The red hue of the truck is vibrant but not overly saturated, blending naturally with the neutral tones of the road and background.
    \end{itemize}
    
    While minor inconsistencies (e.g., perfectly parallel stripes despite the truck’s angle) could suggest AI generation, they might also result from careful photographic composition. Thus, I think this is a ``Real" image. 

    \verb|</reason>|
    \end{tcolorbox}
     \caption{Examples of response of the model trained with H-R. The image is fake while the model gives a ``real" prediction. }
    \label{fig:par_exp}
\end{figure*}

\begin{figure*}[ht]
\centering
\begin{tcolorbox}[listing only, 
                  listing options={basicstyle=\ttfamily\footnotesize},
                  colback=gray!10, 
                  title=Reasoning Examples]
    \begin{center}
        \includegraphics[width=0.4\linewidth]{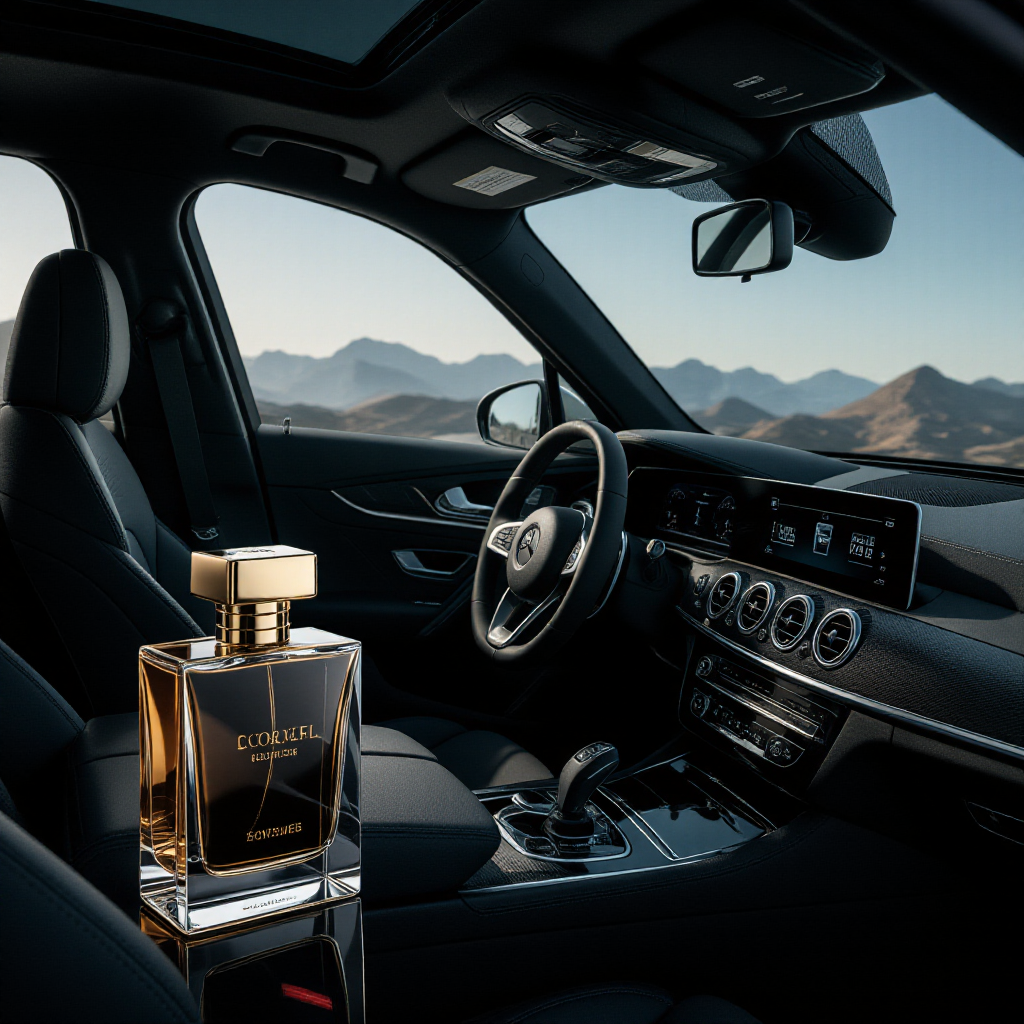}
    \end{center}
    
    \verb|<think>|

- \textbf{Observation}: The image depicts the interior of a luxury car, featuring a perfume bottle labeled ``DCORNFL FLORAL" positioned on the passenger seat. The car’s dashboard displays an Audi logo, with digital screens showing navigation and settings. The exterior view through the windows reveals layered mountain ranges under a clear sky, suggesting a desert-like landscape. The lighting appears natural, emanating from outside the vehicle, casting soft shadows inside. The perfumes’ glass material reflects light realistically, and the car’s leather seats exhibit fine-grained textures. No humans or animals are present.  

- \textbf{Key Areas Evaluation}:
\begin{itemize}
    \item Lighting and Geometry: The lighting seems consistent, with shadows aligning logically (e.g., the perfume’s shadow on the seat). The perspective of the car’s interior matches the viewer’s viewpoint, appearing undistorted.  
    \item Structure and Physics: The car’s components (steering wheel, dashboard, seats) are proportionally accurate. The perfume’s placement on the seat adheres to gravity. 
    \item Focus and Depth of Field: The foreground (perfume, dashboard) is sharply focused, while the background (mountains) is softly blurred, mimicking a natural depth of field effect. Edges of the blurred background do not appear artificially cutout.
    \item Human Pose and Looking: No humans are present, so this category is irrelevant.
    \item Material and Texture: The leather seats display intricate stitching and grain patterns. The perfume’s glass reflects light naturally, and the dashboard materials look touchable.
    \item Text and Logos: The label text on the perfume is legible, and the arring logo on the dashboard matches known Audi designs.
    \item Color: The mountains’ earth tones blend smoothly with the sky, creating a neutral color harmony inside the Car. 
\end{itemize} 

While minor quirks (e.g., the mountain layers appearing slightly flat) could suggest digital manipulation, they do not inherently prove the image is AI-generated. Therefore, this images is real.

    \verb|</think>|
    
    \verb|<answer>| {\textcolor{red}{\textbf{real}}}
    \verb|</answer>|
    \end{tcolorbox}
     \caption{Examples of response of the model trained with A-R. The image is fake while the model gives a ``real" prediction. }
    \label{fig:pra_exp}
\end{figure*}

\begin{figure*}[ht]
\centering
\begin{tcolorbox}[listing only, 
                  listing options={basicstyle=\ttfamily\footnotesize},
                  colback=gray!10, 
                  title=Reasoning Examples]
    \begin{center}
        \includegraphics[width=0.4\linewidth]{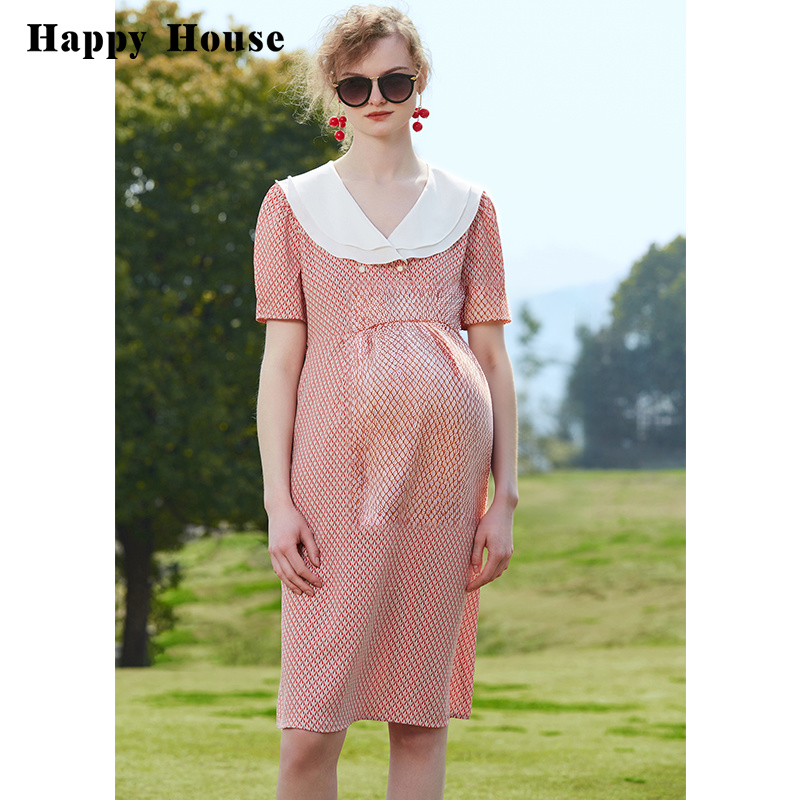}
    \end{center}
    
    \verb|<guess>|
    {\color{red}{\textbf{real}}}
    \verb|</guess>|
    
    \verb|<reason>|

- \textbf{Observation}: The image depicts a woman wearing a glossy white satin blouse paired with a black skirt. Her makeup is flawless, featuring defined eyebrows, eyeliner, and lipstick. She has long, straight dark hair and wears intricate diamond-shaped earrings. The background is uniformly gray, providing a neutral backdrop. Notably, her hands are not visible in the frame, and she appears to be standing upright with a slight smile. The lighting highlights the sheen of the blouse, creating reflections that suggest a professional photography setup. However, upon closer inspection, the area around her collarbone appears slightly distorted, and the earrings seem disproportionately large relative to her earlobes.  

- \textbf{Key Areas Evaluation}:
\begin{itemize}
    \item Lighting and Geometry: The lighting is even and directional, casting subtle shadows under her chin and along the neckline. The reflections on the blouse appear consistent with a single light source, though minor inconsistencies in shadow density hint at potential AI manipulation.
    \item Structure and Physics: The blouse’s fabric drapes naturally over her torso, with buttons aligned correctly. However, the waistline where the blouse meets the skirt seems unnaturally narrow, suggesting possible distortion. 
    \item Focus and Depth of Field: The subject is sharply focused against a softly blurred background, but the transition between her silhouette and the backdrop lacks organic blending, resembling a cut-out effect
    \item Human Pose and Looking: Her facial features are symmetrical and idealized, with no visible imperfections. While this could be attributed to heavy retouching, the complete absence of hand visibility raises suspicion (hands are often omitted in AI-generated images due to difficulty rendering them).
    \item Material and Texture: The satin blouse exhibits realistic glossiness, but the skin texture is unnaturally smooth, lacking pores or subtle imperfections typical of human skin.
    \item Text and Logo: No text or logos are present
    \item Color: The color palette is harmonious, with contrasting whites, blacks, and neutrals. However, the saturation of the white blouse feels slightly exaggerated.  
\end{itemize}

The image’s professionalism and coherence led me to conclude this image is real.

    \verb|</reason>|
    
    \verb|<answer>| {\color{red}{\textbf{real}}}
    \verb|</answer>|
    \end{tcolorbox}
     \caption{Examples of response of the model trained with our G-R-A. The image is fake while the model gives a ``real" prediction. }
    \label{fig:gra_exp}
\end{figure*}

\begin{figure*}[ht]
\centering
\begin{tcolorbox}[listing only, 
                  listing options={basicstyle=\ttfamily\footnotesize},
                  colback=gray!10, 
                  title=Reasoning Examples]
    \begin{center}
        \includegraphics[width=0.35\linewidth]{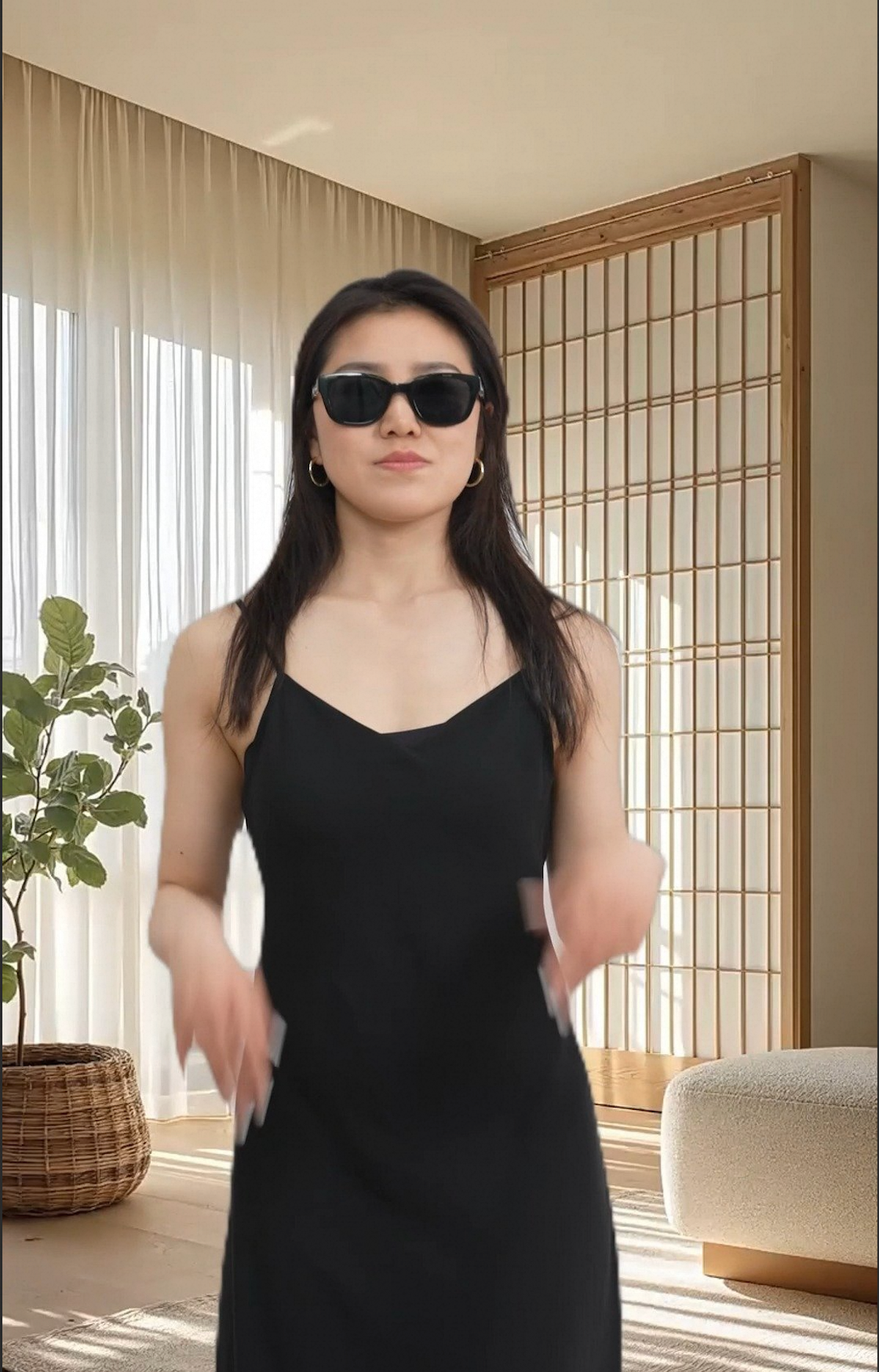}
    \end{center}
    
    \verb|<impression>|
    {\color{red}{\textbf{real}}}
    \verb|</impression>|
    
    \verb|<reason>|

    This image appears to be a photograph of a person standing indoors, possibly in a living room or a similar setting. The person is wearing a black dress and sunglasses, and there is a plant and a wicker basket in the background. The lighting seems to be natural, coming from a window on the left side of the image, which is consistent with the shadows cast on the floor and the person. The room has a modern design with a wooden floor, a white couch, and a shoji screen in the background. The overall scene looks realistic and well-lit, with no apparent signs of digital anomalies or inconsistencies that would suggest AI generation. The person's pose and the way they are holding an object (which appears to be a phone or a small item) look natural and plausible. The image does not show any unusual lighting effects, such as conflicting light sources or wrong shadows, which could be a sign of AI generation. The structure and physics of the scene are rational, with the object's placement and the person's anatomy looking correct. The focus is clear, and the depth of field appears natural, with the background slightly blurred, which is a common technique in photography to emphasize the subject. The material and texture of the person's dress and the surrounding objects also look realistic, with no signs of skin that is too smooth or too pale or skin or metal that looks like plastic. The color balance is natural, and there are no strange colors or oversaturated colors that would suggest AI editing. The image does not show any digital flaws, such as edges of the blurred background that seem to be ``cut out" or show digital artifacts. The human pose and lookings are natural, with no abnormal fingers, facial features, or joints, and the pose is natural and believable. There are no indications of AI generation in the image, and it appears to be a real photograph taken in a real setting.    
    \verb|</reason>|
    
    \verb|<reflection>|
    
    In the Key Areas Evaluation, the image passes all the checks for ``Real' Signs. The lighting and geometry are consistent and matching, the structure and physics are rational, the focus and depth of field look natural, the human pose and lookings are correct and natural, the material and texture appear realistic, the text and logos are correctly spelled and meaningful, and the color is balanced and natural. There are no ``Fake" Signs such as conflicting light sources, wrong shadows, distorted perspective, wrong proportions, placement that ignores gravity, structure that doesn't make sense, unclear focus, the product and background that don't blend naturally, abnormal fingers, facial features, or joints, skin that is too smooth or too pale, skin or metal that looks like plastic, easily spotted repeating textures, unnatural sharp outlines, gibberish, meaningless characters, text that doesn't wrap correctly on curved surfaces, strange color, oversaturated colors, or feeling of obvious unnaturalness from the image. Therefore, the image is likely real.
    
    \verb|</reflection>|
    
    \verb|<answer>| {\color{red}{\textbf{real}}}
    \verb|</answer>|
    \end{tcolorbox}
     \caption{Examples of response of the model trained with GRPO without cold start. The image is fake while the model gives a ``real" prediction. }
    \label{fig:grpo_exp}
\end{figure*}

\begin{figure*}[ht]
\centering
\begin{tcolorbox}[listing only, 
                  listing options={basicstyle=\ttfamily\footnotesize},
                  colback=gray!10, 
                  title=Reasoning Examples]
    \begin{center}
        \includegraphics[width=0.3\linewidth]{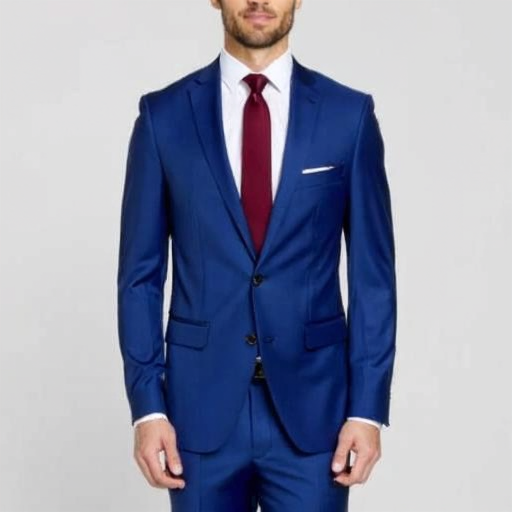}
    \end{center}
    
    \verb|<impression>|
    {\color{red}{\textbf{real}}}
    \verb|</impression>|
    
    \verb|<reason>|
    
- \textbf{Observation}: The image depicts a male model wearing a vibrant blue suit jacket paired with a crisp white shirt and a deep red tie. The background is a uniform, neutral gray, typical of professional photography setups. The lighting is even and frontal, casting soft shadows beneath the chin and around the collar area. The model’s hands are relaxed, with one hand slightly curled, and the suit features visible stitching, button alignment, and a neatly placed pocket square. No obvious distortions or inconsistencies are apparent at first glance.  

- \textbf{Key Areas Evaluation}: 
\begin{itemize}
    \item Lighting and Geometry: The lighting is consistent, with no conflicting shadows or unrealistic highlights. The perspective is straightforward, suggesting a standard photographic setup.
    \item Structure and Physics: The suit’s fit appears rational, with proper sleeve length and waistline positioning. The model’s posture is upright, adhering to natural human biomechanics.
    \item Focus and Depth of Field: The subject is sharply focused, while the background remains uniformly blurred without noticeable digital artifacts. The transition between foreground and background seems seamless.
    \item Human Pose and Looking: The model’s hands exhibit normal anatomical proportions, with visible finger joints and natural positioning. The beard stubble is detailed, adding realism.
    \item Material and Texture: The suit’s fabric displays subtle wrinkles and sheen, mimicking real wool or synthetic blends. The skin tone is smooth but retains minor imperfections, such as pores near the beard line.
    \item Text and Logo: No text or logos are present in the image.
    \item Color: The color palette is harmonious, with the blue suit contrasting effectively against the neutral background. The red tie adds vibrancy without appearing oversaturated. 
\end{itemize}

While minor quirks (e.g., slight asymmetry in the suit’s lapel) could be attributed to photography imperfections, the overall coherence supports a ``Real" impression.

    \verb|</reason>|
    
    \verb|<reflection>|

Upon revisiting the image, several overlooked anomalies emerge: 

- \textbf{Observation Corrections}:  
\begin{itemize}
    \item The suit’s right sleeve (viewer’s left) shows inconsistent draping, with folds that defy gravitational logic, suggesting algorithmic rendering errors.
    \item The pocket square’s diagonal fold is unnaturally precise, lacking the organic creases of real fabric.
\end{itemize}

- \textbf{Key Areas Reevaluation}:
\begin{itemize}
    \item Material and Texture: While the suit’s texture initially seemed realistic, closer inspection reveals repetitive patterns in the fabric’s weave—a hallmark of AI-generated textures.
    \item Color: The blue suit’s saturation is unnaturally uniform, lacking the subtle tonal variations seen in physical materials under natural light.
\end{itemize}

These corrections undermine the initial ``Real" assessment. The combination of artificial textures and the consistency of color breaks strongly indicate AI generation.

    \verb|</reflection>|
    
    \verb|<answer>| {\color{blue}{\textbf{fake}}}
    \verb|</answer>|
    \end{tcolorbox}
     \caption{Success reflection example of the response from our \model.}
    \label{fig:sft_success}
\end{figure*}

\begin{figure*}[ht]
\centering
\begin{tcolorbox}[listing only, 
                  listing options={basicstyle=\ttfamily\footnotesize},
                  colback=gray!10, 
                  title=Reasoning Examples]
    \begin{center}
        \includegraphics[width=0.1\linewidth]{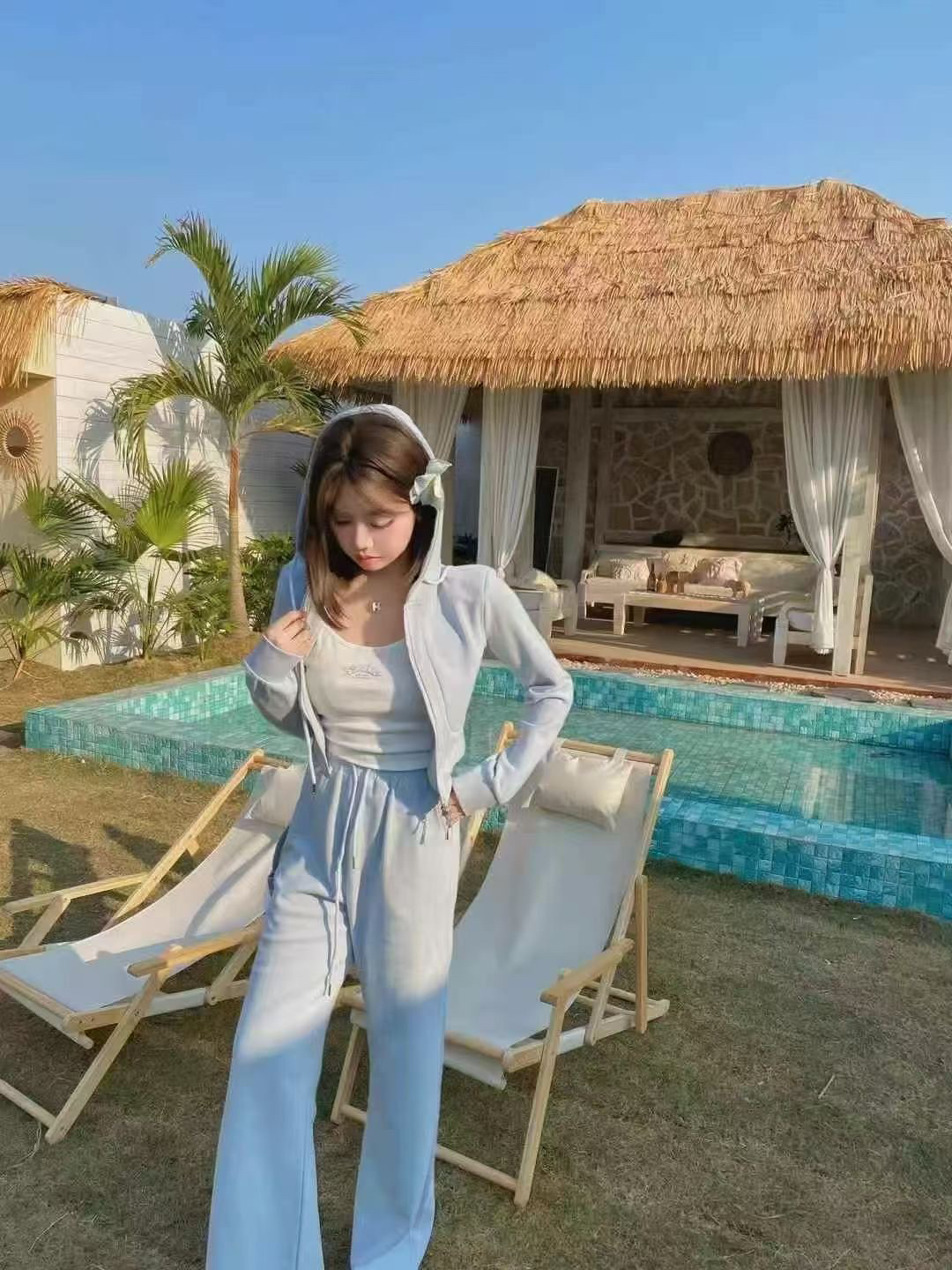}
    \end{center}
    
    \verb|<impression>|
    {\textcolor{blue}{\textbf{fake}}}
    \verb|</impression>|
    
    \verb|<reason>|
    
- \textbf{Observation}: The image depicts a young woman standing outdoors near a swimming pool, wearing a light blue tracksuit with a hood. She has her hair tied back and wears a white bow accessory. The background features a thatched-roof structure resembling a cabana, surrounded by palm trees and modern architectural elements. The lighting appears bright and natural, suggesting daytime. Notable details include the woman’s attire, the crisp edges of the pool’s tiled edge, and the textured grass foreground. 

- \textbf{Key Areas Evaluation}:  
\begin{itemize}
    \item Lighting and Geometry: The lighting seems consistent, with shadows cast toward the right side of the frame. However, the shadow under the woman’s feet appears slightly disconnected from her stance, raising suspicion about artificial manipulation.
    \item Structure and Physics: The thatched roof and cabana structure exhibit logical construction, but the integration of the cabana into the scene feels somewhat staged, as if placed digitally.
    \item Focus and Depth of Field**: The subject is sharply focused, while the background is softly blurred. However, the transition between the woman and the grassy foreground lacks natural blending, hinting at potential cut-out effects.
    \item Human Pose and Looking: The woman’s posture is relaxed, but her left hand (visible in the foreground) appears disproportionately small and lacks detailed finger definition, a common AI artifact.
    \item Material and Texture: The skin texture is unnaturally smooth, lacking subtle imperfections typical of real photographs. The fabric of the tracksuit reflects light uniformly, appearing almost synthetic.
    \item Text and Logo: A small logo on the woman’s chest is illegible, possibly due to low resolution or intentional obfuscation.
    \item Color: The overall color palette is vibrant but harmonious. However, the saturation of the blue tiles in the pool contrasts sharply with the muted tones of the grass, creating an uncanny disconnect.
\end{itemize}

These observations collectively suggest AI generation, leading to my initial ``Fake" impression.

    \verb|</reason>|
    
    \verb|<reflection>|

- \textbf{Observation Corrections}:  
\begin{itemize}
    \item Upon re-examination, the shadow under the woman’s feet aligns naturally with her stance, indicating proper lighting geometry.
    \item The cabana’s integration into the scene is contextually appropriate for a tropical setting, not necessarily indicative of digital placement. 
    \item The ``disconnected" hand was misinterpreted; closer inspection reveals it is partially obscured by the woman’s body, resolving the proportionality concern.
\end{itemize}

- \textbf{Key Areas Reevaluation}
\begin{itemize}
    \item Lighting and Geometry: The shadows are coherent with a single light source (sunlight), validating the scene’s realism.
    \item Focus and Depth of Field: The soft blur of the background smoothly transitions into the foreground, mimicking a shallow depth of field achieved via photography techniques.
    \item Human Pose and Looking: The hand’s detail is sufficient for a casual pose, and the skin texture, while smoothed, retains minor pores and natural variations, contradicting earlier assumptions of AI synthesis.
    \item Material and Texture: The fabric’s reflectivity is consistent with lightweight materials, and the pool’s tile pattern shows irregularities typical of physical constructions. 
\end{itemize}

My initial errors stemmed from overemphasizing minor inconsistencies without considering photographic editing or stylistic choices. The image’s coherence across multiple dimensions now supports authenticity.

    \verb|</reflection>|
    
    \verb|<answer>| {\textcolor{red}{\textbf{real}}}
    \verb|</answer>|
    \end{tcolorbox}
     \caption{Failed example of the response from our \model.}
    \label{fig:sft_fail}
\end{figure*}

\begin{itemize}
    \item \textbf{Baseline Models (Lacking Reflection)}: Figures~\ref{fig:par_exp} through \ref{fig:gra_exp} present reasoning examples from our baseline models (H-R, A-R, and G-R-A). As these models inherently lack a reflective mechanism, they are prone to making unrecoverable errors based on flawed initial assessments, demonstrating the limitations of single-pass reasoning.
    \item \textbf{Ablated Model (GRPO-only)}: Figure~\ref{fig:grpo_exp} showcases a critical failure case from a model trained solely with GRPO, bypassing our SFT cold-start stage. The model not only misclassifies a fake image as real but also generates reasoning that fails to adhere to the structured format defined in our instructions. This lack of structure severely undermines the model's explainability and indicates a failure to internalize the task's core requirements, highlighting the critical role of SFT in establishing a strong foundation.
    \item \textbf{Our Full Model (HA-R)}: We present both a success and an insightful failure case for our full HA-R model.
    \begin{itemize}
        \item Successful Reflection (Fig.~\ref{fig:sft_success}): This example demonstrates the intended behavior of our model. It initially makes an incorrect heuristic judgment but effectively self-corrects during the reflective reasoning phase to arrive at the correct final answer.
        \item ``Over-thinking" Failure (Fig.~\ref{fig:sft_fail}): This case exemplifies the ``over-thinking" phenomenon discussed in the main paper. Here, the model's initial ``fast" answer is correct, but the subsequent, unnecessary reflection process leads it to an incorrect final conclusion. This highlights the delicate balance our adaptive mechanism aims to strike and validates the need for a confidence-based trigger to avoid potentially counterproductive deep reasoning.
    \end{itemize}
\end{itemize}

\begin{figure*}[ht]
\centering
\begin{tcolorbox}[listing only, 
                  listing options={basicstyle=\ttfamily\footnotesize},
                  colback=gray!10, 
                  title=Instruction]
\# Role

- You are an expert on detecting AI-generated images by analyzing visual anomalies in the content and details of the image.

\# Evaluation Framework

1. Concepts
\begin{itemize}
    \item \textbf{Real images}: Images with clear content. Proper post-processing on the image (like professional photo editing or background changes) is allowed, as long as the final image looks natural and believable.
    \item \textbf{Fake images}: Images with signs of AI generation or AI editing. 
\end{itemize}

2. Key Areas to Evaluate (Neutral View)

The evaluation is based on the following key categories. Each category includes checkpoints to confirm for ``real" or ``fake".

\indent\textbf{Lighting and Geometry}: 
\begin{itemize}
    \item ``Real" Signs: consistent lighting, matching shadows, and correct perspective;
    \item ``Fake" Signs: Conflicting light sources, wrong shadows, distorted perspective.
\end{itemize}
\indent\textbf{Structure and Physics}
\begin{itemize}
    \item ``Real" Signs: The object's structure, proportions, and placement is rational;
    \item ``Fake" Signs: Wrong proportions, placement that ignores gravity, structure that doesn't make sense.
\end{itemize}
\indent\textbf{Focus and Depth of Field}
\begin{itemize}
    \item ``Real" Signs: The focus is clear and the depth of field looks natural;
    \item ``Fake" Signs: Unclear focus, the product and background don't blend naturally, the edges of the blurred background seem to be "cut out" or show digital flaws.
\end{itemize}
\indent\textbf{Human Pose and Looking}
\begin{itemize}
    \item ``Real" Signs: The anatomy of facial features, limbs, and fingers is correct and the pose is natural;
    \item ``Fake" Signs: Abnormal fingers, facial features, or joints; unnatural poses.
\end{itemize}
\indent\textbf{Material and Texture}
\begin{itemize}
    \item ``Real" Signs: The texture of skin and product materials looks realistic;
    \item ``Fake" Signs: Skin that is too smooth or too pale, skin or metal that looks like plastic, easily spotted repeating textures, unnatural sharp outlines.
\end{itemize}
\indent\textbf{Text and Logos}
\begin{itemize}
    \item ``Real" Signs: Correctly-spelled and meaningful texts and logos;
    \item ``Fake" Signs: Gibberish, meaningless characters, text that doesn't wrap correctly on curved surfaces.
\end{itemize}
\indent\textbf{Color}
\begin{itemize}
    \item ``Real" Signs: The overall color of the image looks balanced and natural;
    \item ``Fake" Signs: Strange hue, over-saturated colors, or feeling of obvious unnaturalness from the image.
\end{itemize}
\end{tcolorbox}
\caption{Instruction of our annotation. }
\label{list:instruction}
\end{figure*}

\begin{figure*}[ht]
\centering
\begin{tcolorbox}[listing only, 
                  listing options={basicstyle=\ttfamily\footnotesize},
                  colback=gray!10, 
                  title=Strategy-1-Round-1-Part-1]
    \verb|{Instruction}|
    
        \# Task
        
        Write an analysis report for AI-generated image detection. 

        - \textbf{Input image}: \verb|<image>|

        \#\# Task instruction
        
        Judge whether the input image is AI-generated and write detailed reasons. Your core task is to explain the basis for your judgment in the report.

        \textbf{Please strictly follow the output format below:}
        \begin{itemize}
            \item \textbf{Format}: Strictly output in the format of: \verb|[Answer]|Real/Fake \verb|[Reasons]| ...
            \item \textbf{Content of Reasons}: Your \verb|[Reasons]| should begin with description on the image content, followed by an analysis of key areas suggested by the user. Finally, it must conclude with the sentence: ``Therefore, I think this is a real/fake image."
            \item \textbf{Style}: Your \verb|[Reasons]| must be logically clear and well-argued. Its style and level of detail must strictly follow the example provided by the user.
        \end{itemize}

        ---
        
        \# Execute Task

        \#\# Example 1
        
        \verb|[Answer]| Real

        \verb|[Reasons]|
        
        \textbf{Observation}: The image shows a female model wearing a khaki casual sportswear set, posing in a semi-squatting position. She is wearing black sunglasses, has her hair in a neat high ponytail, and there is cursive text reading ``Ambie" or something similar on her chest.

        I will evaluate this image's performance on key areas from both positive and negative aspects to maintain an objective analysis.

        \begin{itemize}
            \item Lighting and Geometry: The lighting is handled very professionally. A single primary light source casts consistent, soft-edged shadows on the model and the ground. The shadow cast on the ground by the squatting model is particularly natural in its shape and density, firmly ``grounding" the figure in the space and completely avoiding the feel of a generated image.
            \item Structure and Physics: Although the model's pose is dynamic, it fully conforms to human biomechanics, appearing natural and stable. The wrinkles on the clothing are formed by the bending of the body, are distributed reasonably, and realistically reflect the physical properties of the fabric.
            \item Focus and Depth of Field: The background, consisting of a plain beige wall and a dark floor, has an unnatural cut-out feeling in relation to the model.
            \item Human Pose and Looking: The model's facial features and limbs are well-proportioned, her posture is graceful, and there are no anatomical abnormalities. The details of the hair are clear, with highlights and individual strands well-rendered.
            \item Material and Texture: The model's skin has undergone significant post-processing and is very smooth. However, it should be noted that this retouching remains within the scope of proper editing. While it smooths the skin, it preserves the natural light and shadow transitions of the face and body structure.
            \item Text and Logo: The cursive text on the chest ("Ambie" or similar) is successful as a design element, but its legibility is slightly low. If this were a brand logo needing significant promotion, its readability might be questionable. However, in this context, it functions more as a graphic decoration, so this does not pose a serious problem.
            \item Color: The overall color hue of the image is unified and harmonious. The khaki clothing, beige background, and dark brown floor are well-matched, creating a warm and high-end visual feel. The color reproduction is accurate, with no unnatural color casts.
        \end{itemize}
        Therefore, I think this is a real image. 

        ... (see part 2)
\end{tcolorbox}
\caption{Annotation prompt of round 1 in strategy 1 (part 1). The prompt is so long that we have to split it into two parts. Note that the two parts are combined to be feed to the annotation VLM. Here, ``Instruction" refers to Fig.~\ref{list:instruction}.}
\label{list:round1-1}
\end{figure*}

\begin{figure*}[ht]
\centering
\begin{tcolorbox}[listing only, 
                  listing options={basicstyle=\ttfamily\footnotesize},
                  colback=gray!10, 
                  title=Strategy-1-Round-1-Part-2]
        ... (see part 1)
        
        \#\# Example 2
        
        \verb|[Answer]| Fake

        \verb|[Reasons]|
        
        \textbf{Observation}: The image shows a young woman sitting on a chair with a rattan back, wearing a white round-neck short-sleeved T-shirt paired with blue jeans. On the chest of the T-shirt, there is a colorful phoenix embroidery pattern. The woman has neat hair and is wearing a necklace, earrings, and a watch. Her posture is natural, and she is looking directly at the camera. Her skin, especially on her face, neck, and arms, appears unusually smooth. There is some ghosting on her eyes, which appears to be a low-level flaw from AI generation.

        I will evaluate this image's performance on key areas from both positive and negative aspects to maintain an objective analysis.

        \begin{itemize}
            \item Lighting and Geometry: The lighting has obvious issues. The light is too flat, causing the person and clothing to lack the 3-D appearance. More seriously, there is a lack of realistic shadow between the model's body and the chair back, which defies the laws of physics. She looks as if she has been digitally ``pasted" in front of the chair rather than naturally leaning against it, which severely undermines the scene's realism.
            \item Structure and Physics: The image's composition is professional. The use of a simple, solid-color background effectively focuses the viewer's attention on the model and the main garment, avoiding unnecessary environmental distractions. This is a common and effective strategy for professional photos with white or solid-color backgrounds.
            \item Focus and Depth of Field: Upon close inspection of the hair edge on the model's right shoulder (left from the viewer's perspective), blurry and smeared marks are visible. This is not a natural blur caused by optical depth of field, but rather looks like an artifact from poor background replacement or AI generation.
            \item Human Pose and Looking: There is some ghosting on the model's eyes, which is strong evidence of AI generation — professional retouching would never overlook such a basic artifact.
            \item Material and Texture: The model's skin has a completely unrealistic ``plastic" or ``ceramic" feel. Professional retouching usually preserves basic skin texture to maintain realism, but this image has completely erased all physiological features like pores. This excessive smoothing makes the model look less like a real person and more like a digitally rendered virtual human or a wax figure. Besides the skin, the texture of other materials is also affected. Some areas of the model's hair have lost their strand-like details, becoming a blurry patch of color. Meanwhile, the T-shirt fabric also appears overly smooth, lacking the subtle texture that cotton fabric should have, which further exacerbates the overall "fake" feel of the picture.
            \item Text and Logo: There is no text in the image.
            \item Color: The image does not have any strange color casts. The overall hue is harmonious and bright.
        \end{itemize}
        Therefore, I think this is a fake image. 

        \#\# Now write your report: 
\end{tcolorbox}
\caption{Annotation prompt of round 1 in strategy 1 (part 2).}
\label{list:round1-2}
\end{figure*}
\begin{figure*}[ht]
\centering
\begin{tcolorbox}[listing only, 
                  listing options={basicstyle=\ttfamily\footnotesize},
                  colback=gray!10, 
                  title=Strategy-1-Round-2]
        \# Task

        The ground truth label of this image is \verb|{Ground Truth}|. Review your first impression and the reasons, write a complete analysis report for this image involving in-depth and reflective thinking. 

        \#\# Task instruction
        
        You should:
        
        Reflective review your first impression and reasons. Identify and correct the errors in \textbf{observation} and \textbf{Key Areas Evaluation}, respectively. In the end, conclude with ``I think this is a \verb|{Ground Truth}| image.".  

        \textbf{Please strictly follow the output format below:}
        \begin{itemize}
            \item \textbf{Format}: Strictly output in the format of: \verb|[Reflection]| ... \verb|[Answer]Real/Fake|.
            \item \textbf{Style}: Your \verb|[Reflection]| must be logically clear and well-argued. Its style and level of detail must strictly follow the example provided by the user.
        \end{itemize}
        ---

        \# Execute Task

        \#\# Example
        
        \verb|[Reflection]|
        
        My first impression on the image is ``Fake". Now, I will review the impression as well as its supports, reflectively think, correct the inaccurate parts, and draw a final conclusion. 

        \textbf{Observation}:
        \begin{itemize}
            \item In my initial observation, I forgot to describe the lighting in the photo. The lighting comes from the left-front side of the model. The light is soft and natural. 
            \item After careful reviewing, I find the texts on the chest of the model are not garbled -- they are actually cursive scripts reading ``Ambie". This might be the brand name of the sportswear. 
        \end{itemize}

        \textbf{Key Areas Evaluation}:
        \begin{itemize}
            \item Lighting and Geometry: After updating my observation, I should update the corresponding dimension. The lighting is handled very professionally. A single primary light source casts consistent, soft-edged shadows on the model and the ground. The shadow cast on the ground by the squatting model is particularly natural in its shape and density, firmly ``grounding" the figure in the space and completely avoiding the feel of a generated image.
            \item Focus and Depth of Field: At the first glance, I notice the unnatural cut-out feeling between the model and the background. I need to discriminate the feeling of ``concise" and that of ``fake": according to my updated evaluation on ``Lighting and Geometry", the natural shadow cast on the ground by the model proves that none of the model and the background is generated. 
            \item Human Pose and Looking: At the first, I think there is retouching on the model. I need to differentiate the ``retouching" and ``AI generation": in the image, although the skin is smooth, it still has muscle texture and visible (although not very obvious) pores. 
            \item Text and Logo: After updating my observation, the texts on the model's chest are cursive scripts instead of garbled texts. This dimension is no longer a support for ``fake". 
        \end{itemize}
        
        Based on the analysis above, although some of the points in my first analysis are correct, I misunderstand some image details and the major points supporting the first impression do not exist any more. 
        
        I think this is a real image. 

        \verb|[Answer]| Real

        \#\# Now write your report:

\end{tcolorbox}
\caption{Annotation prompt of round 2 in strategy 1. Note that we include the first round model response in the conversation history when we are fetching the round 2 answer. In the prompt, ``Ground Truth" means the ground-truth label of the image.}
\label{list:round2} 
\end{figure*}

\begin{figure*}[ht]
\centering
\begin{tcolorbox}[listing only, 
                  listing options={basicstyle=\ttfamily\footnotesize},
                  colback=gray!10, 
                  title=Strategy-2-Fake-Fake-Part-1]
        \verb|{Instruction}|
        
        \# Task
        \# Context
        You are viewing a suspected AI-generated image two times. For the first time, you give your fast impression on the image at the first glance as well as your reasons. For the second time, you review your first impression and reasons, reflectively think, correct the potential errors, and draw a final conclusion. 
        
        Now you have already finished your first view. You have known that:
        
        - \textbf{Input image}: \verb|<image>|
        
        - \textbf{The ground truth label for this image is}: Fake
        
        - \textbf{Your impression}: Fake

        Your task is to write a complete analysis report for this image involving in-depth and reflective thinking. 

        \#\# Task instruction
        
        You should:
        
        1. Based on your first impression, write the reasons supporting the impression. The reasons should involve the two step below sequentially:
        \begin{itemize}
            \item Observation: Captioning the image content, including the subject and object in the photo, the background, the lighting, texts and any unnatural details. 
            \item Key Areas Evaluation: Analyze the image from each dimension from the key areas suggested by the user. In the end, correlated the reasons with your impression. 
        \end{itemize}
        
        2. Reflective review your analysis above. Identify and correct the errors in \textbf{observation} and \textbf{Key Areas Evaluation}, respectively. In the end, conclude with ``I think this is a fake image.".  

        \textbf{Please strictly follow the output format below:}
        \begin{itemize}
            \item \textbf{Format}: Strictly output in the format of: \verb|[Impression]| Fake \verb|[Reasons]| ... \verb|[Reflection]| ... \verb|[Answer]| Fake.
            \item \textbf{Style}: Your \verb|[Reasons]| and \verb|[Reflection]| must be logically clear and well-argued. Its style and level of detail must strictly follow the example provided by the user.
            \item \textbf{Objective Judgment}: You should independently produce \verb|[Reasons]| and \verb|[Reflection]|. You are not allowed to make mistakes in \verb|[Reasons]| and correct them in \verb|[Reflection]| on purpose. However, in \verb|[Reasons]|, you can leave some details that could be ignored by humans at the first glance. You should only write the most obvious reasons in \verb|[Reasons]|.
        \end{itemize}
        
        ---

        \# Execute Task

        \#\# Example
        
        \verb|[Impression]| Fake

        \verb|[Reasons]|
        
        - \textbf{Observation}: The image shows a young woman sitting on a chair with a rattan back, wearing a white round-neck short-sleeved T-shirt paired with blue jeans. On the chest of the T-shirt, there is a colorful phoenix embroidery pattern. The woman has neat hair and is wearing a necklace, earrings, and a watch. Her posture is natural, and she is looking directly at the camera. Her skin, especially on her face, neck, and arms, appears unusually smooth. In addition, the fingers of the model are missing. 

        - \textbf{Key Areas Evaluation}:
        
        I will evaluate this image's performance on key areas from both positive and negative aspects to maintain an objective analysis.
        
        ...(see part 2)
\end{tcolorbox}
\caption{Annotation prompt of strategy 2 when the ground-truth label of the sample is ``Fake" and the initial guess is controlled to be ``Fake" (part 1). }
\label{list:strategy2-1-1}
\end{figure*}

\begin{figure*}[ht]
\centering
\begin{tcolorbox}[listing only, 
                  listing options={basicstyle=\ttfamily\footnotesize},
                  colback=gray!10, 
                  title=Strategy-2-Fake-Fake-Part-2]
        ...(see part 1)
        
        \begin{itemize}
            \item Lighting and Geometry: The lighting has obvious issues. The light is too flat, causing the person and clothing to lack the 3-D appearance. More seriously, there is a lack of realistic shadow between the model's body and the chair back, which defies the laws of physics. She looks as if she has been digitally ``pasted" in front of the chair rather than naturally leaning against it, which severely undermines the scene's realism.
            \item Structure and Physics: The image's composition is professional. The use of a simple, solid-color background effectively focuses the viewer's attention on the model and the main garment, avoiding unnecessary environmental distractions. This is a common and effective strategy for professional photos with white or solid-color backgrounds.
            \item Focus and Depth of Field: The focus and depth of field of the photo is natural. 
            \item Human Pose and Looking: The fingers of the model are missing -- this is a strong sign of AIGC. 
            \item Material and Texture: The model's skin has a completely unrealistic ``plastic" or ``ceramic" feel. Professional retouching usually preserves basic skin texture to maintain realism, but this image has completely erased all physiological features like pores. This excessive smoothing makes the model look less like a real person and more like a digitally rendered virtual human or a wax figure. Besides the skin, the texture of other materials is also affected. Some areas of the model's hair have lost their strand-like details, becoming a blurry patch of color. Meanwhile, the T-shirt fabric also appears overly smooth, lacking the subtle texture that cotton fabric should have, which further exacerbates the overall ``fake" feel of the picture.
            \item Text and Logo: There is no text in the image.
            \item Color: The image does not have any strange color casts. The overall hue is harmonious and bright.
        \end{itemize}
        
        The flat light, missing fingers of the model, as well as the over-smoothed skin are all clear evidence supporting my first impression. For the reasons above, my first impression on the image is ``Fake".
        
        \verb|[Reflection]|
        
        My first impression on the image is ``Fake". Now, I will review the impression as well as its supports, reflectively think, correct the inaccurate parts, and draw a final conclusion. 

        - \textbf{Observation}:
        
            - In my initial observation, I forgot to check the eyes of the model -- there is some ghosting on her eyes, which appears to be a low-level flaw from AI generation. 
            - My fast observation on the fingers is wrong. Actually the fingers are hidden inside the pockets of the model's jeans. This is natural. 

        - \textbf{Key Areas Evaluation}:
        \begin{itemize}
            \item Focus and Depth of Field: At the first glance, I think the focus of the image is natural. When I inspect the details carefully, I find that upon close inspection of the hair edge on the model's right shoulder (left from the viewer's perspective), blurry and smeared marks are visible. This is not a natural blur caused by optical depth of field, but rather looks like an artifact from poor background replacement or AI generation.
            \item Human Pose and Looking: After updating my observation, I find my first impression on the fingers are wrong. They are not missing but inside the jeans pockets. I also ignore the flaw on the eyes of the model, which is a strong sign of AIGC. 
            \item Material and Texture: At the first, I think there is retouching on the model. I need to differentiate the ``retouching" and ``AI generation": in the image, the skin of the model is too smooth that I can not see any pores or muscle texture. My first impression on this dimension is correct.
            \item Color: In my fast impression, the image color is harmonious. However, at the second glance, I feel a strong ``unnaturalness". This could be the color casts brought by AI models. 
        \end{itemize}
        
        Based on the analysis above, although there are some mistakes and missing details in my first analysis, the major points supporting the first impression still exist. 

        To conclude, I think this is a fake image. 

        \verb|[Answer]| Fake

        \#\# Now write your report:
\end{tcolorbox}
\caption{Annotation prompt of strategy 2 when the ground-truth label of the sample is ``Fake" and the initial guess is controlled to be ``Fake" (part 2). }
\label{list:strategy2-1-2}
\end{figure*}

\begin{figure*}[ht]
\centering
\begin{tcolorbox}[listing only, 
                  listing options={basicstyle=\ttfamily\footnotesize},
                  colback=gray!10, 
                  title=Strategy-2-Real-Fake-Part-1]
        \verb|{Instruction}|
        
        \# Task
        
        \# Context
        
        You are viewing a suspected AI-generated image two times. For the first time, you give your fast impression on the image at the first glance as well as your reasons. For the second time, you review your first impression and reasons, reflectively think, correct the potential errors, and draw a final conclusion. 
        Now you have already finished your first view. You have known that:
        - \textbf{Input image}: \verb|<image>|
        - \textbf{The ground truth label for this image is}: Fake
        - \textbf{Your impression}: Real

        Your task is to write a complete analysis report for this image involving in-depth and reflective thinking. 

        \#\# Task instruction
        
        You should:
        
        1. Based on your first impression, write the reasons supporting the impression. The reasons should involve the two step below sequentially:
        \begin{itemize}
            \item Observation: Captioning the image content, including the subject and object in the photo, the background, the lighting, texts and any unnatural details. 
            \item Key Areas Evaluation: Analyze the image from each dimension from the key areas suggested by the user. In the end, correlated the reasons with your impression. 
        \end{itemize}
        
        2. Reflective review your analysis above. Identify and correct the errors in \textbf{observation} and \textbf{Key Areas Evaluation}, respectively. In the end, conclude with ``I think this is a fake image.".  

        \textbf{Please strictly follow the output format below:}
        \begin{itemize}
            \item \textbf{Format}: Strictly output in the format of: \verb|[Impression]| Real \verb|[Reasons]| ... \verb|[Reflection]| ... \verb|[Answer]| Fake.
            \item \textbf{Style}: Your \verb|[Reasons]| and \verb|[Reflection]| must be logically clear and well-argued. Its style and level of detail must strictly follow the example provided by the user.
            \item \textbf{Objective Judgment}: You should independently produce \verb|[Reasons]| and \verb|[Reflection]|. You are not allowed to make mistakes in \verb|[Reasons]| and correct them in \verb|[Reflection]| on purpose. However, in \verb|[Reasons]|, you can leave some details that could be ignored by humans at the first glance. You should only write the most obvious reasons in \verb|[Reasons]|.
        \end{itemize}
        
        ---

        \# Execute Task

        \#\# Example
        
        \verb|[Impression]| Real

        \verb|[Reasons]|
        
        - \textbf{Observation}: The image shows a young woman sitting on a chair with a rattan back, wearing a white round-neck short-sleeved T-shirt paired with blue jeans. On the chest of the T-shirt, there is a colorful phoenix embroidery pattern. The woman has neat hair and is wearing a necklace, earrings, and a watch. Her posture is natural, and she is looking directly at the camera. Her skin, especially on her face, neck, and arms, appears unusually smooth. In addition, the fingers of the model are missing. 

        - \textbf{Key Areas Evaluation}:
        
        I will evaluate this image's performance on key areas from both positive and negative aspects to maintain an objective analysis.
        
        ...(see part 2)
        
\end{tcolorbox}
\caption{Annotation prompt of strategy 2 when the ground-truth label of the sample is ``Fake" and the initial guess is controlled to be ``Real" (part 1). }
\label{list:strategy2-2-1}
\end{figure*}

\begin{figure*}[ht]
\centering
\begin{tcolorbox}[listing only, 
                  listing options={basicstyle=\ttfamily\footnotesize},
                  colback=gray!10, 
                  title=Strategy-2-Real-Fake-Part-2]
        ...(see part 1)

        \begin{itemize}
            \item Lighting and Geometry: The lighting has obvious issues. The light is too flat, causing the person and clothing to lack the 3-D appearance. More seriously, there is a lack of realistic shadow between the model's body and the chair back, which defies the laws of physics. She looks as if she has been digitally ``pasted" in front of the chair rather than naturally leaning against it, which severely undermines the scene's realism.
            \item Structure and Physics: The image's composition is professional. The use of a simple, solid-color background effectively focuses the viewer's attention on the model and the main garment, avoiding unnecessary environmental distractions. This is a common and effective strategy for professional photos with white or solid-color backgrounds.
            \item Focus and Depth of Field: The focus and depth of field of the photo is natural. 
            \item Human Pose and Looking: The fingers of the model are missing -- this is a strong sign of AIGC. 
            \item Material and Texture: The model's skin has a completely unrealistic ``plastic" or ``ceramic" feel. Professional retouching usually preserves basic skin texture to maintain realism, but this image has completely erased all physiological features like pores. This excessive smoothing makes the model look less like a real person and more like a digitally rendered virtual human or a wax figure. Besides the skin, the texture of other materials is also affected. Some areas of the model's hair have lost their strand-like details, becoming a blurry patch of color. Meanwhile, the T-shirt fabric also appears overly smooth, lacking the subtle texture that cotton fabric should have, which further exacerbates the overall ``fake" feel of the picture.
            \item Text and Logo: There is no text in the image.
            \item Color: The image does not have any strange color casts. The overall hue is harmonious and bright.
        \end{itemize}
        
        The image has some suspectful issues (e.g., flat light, missing fingers, over-smoothed skin). However, these issues could also be brought by proper post-processing and bad photographing. For the reasons above, my first impression on the image is ``Real".
        
        \verb|[Reflection]|
        
        My first impression on the image is ``Real". Now, I will review the impression as well as its supports, reflectively think, correct the inaccurate parts, and draw a final conclusion. 

        - \textbf{Observation}:
        
            - In my initial observation, I forgot to check the eyes of the model -- there is some ghosting on her eyes, which appears to be a low-level flaw from AI generation. 
            - My fast observation on the fingers is wrong. Actually the fingers are hidden inside the pockets of the model's jeans. This is natural. 

        - \textbf{Key Areas Evaluation}:
        \begin{itemize}
            \item Focus and Depth of Field: At the first glance, I think the focus of the image is natural. When I inspect the details carefully, I find that upon close inspection of the hair edge on the model's right shoulder (left from the viewer's perspective), blurry and smeared marks are visible. This is not a natural blur caused by optical depth of field, but rather looks like an artifact from poor background replacement or AI generation.
            \item Human Pose and Looking: After updating my observation, I find my first impression on the fingers are wrong. They are not missing but inside the jeans pockets. I also ignore the flaw on the eyes of the model, which is a strong sign of AIGC. 
            \item Material and Texture: At the first, I think there is retouching on the model. I need to differentiate the ``retouching" and ``AI generation": in the image, the skin of the model is too smooth that I can not see any pores or muscle texture. My first impression on this dimension is correct.
            \item Color: In my fast impression, the image color is harmonious. However, at the second glance, I feel a strong ``unnaturalness". This could be the color casts brought by AI models. 
        \end{itemize}
        
        Based on the analysis above, although some of the points in my first analysis are correct, I misunderstand some image details and the major points supporting the first impression do not exist any more.  

        To conclude, I think this is a fake image. 

        \verb|[Answer]| Fake

        \#\# Now write your report:
\end{tcolorbox}
\caption{Annotation prompt of strategy 2 when the ground-truth label of the sample is ``Fake" and the initial guess is controlled to be ``Real" (part 2). }
\label{list:strategy2-2-2}
\end{figure*}

\begin{figure*}[ht]
\centering
\begin{tcolorbox}[listing only, 
                  listing options={basicstyle=\ttfamily\footnotesize},
                  colback=gray!10, 
                  title=Strategy-2-Real-Real-Part-1]
        \verb|{Instruction}|
        
        \# Task
        
        \# Context
        
        You are viewing a suspected AI-generated image two times. For the first time, you give your fast impression on the image at the first glance as well as your reasons. For the second time, you review your first impression and reasons, reflectively think, correct the potential errors, and draw a final conclusion. 
        
        Now you have already finished your first view. You have known that:
        
        - \textbf{Input image}: \verb|<image>|
        
        - \textbf{The ground truth label for this image is}: Real
        
        - \textbf{Your impression}: Real

        Your task is to write a complete analysis report for this image involving in-depth and reflective thinking. 

        \#\# Task instruction
        
        You should:
        
        1. Based on your first impression, write the reasons supporting the impression. The reasons should involve the two step below sequentially:
        \begin{itemize}
            \item Observation: Captioning the image content, including the subject and object in the photo, the background, the lighting, texts and any unnatural details. 
            \item Key Areas Evaluation: Analyze the image from each dimension from the key areas suggested by the user. In the end, correlated the reasons with your impression. 
        \end{itemize}
        
        2. Reflective review your analysis above. Identify and correct the errors in \textbf{observation} and \textbf{Key Areas Evaluation}, respectively. In the end, conclude with ``I think this is a real image.".  

       \textbf{ Please strictly follow the output format below:}
       \begin{itemize}
           \item \textbf{Format}: Strictly output in the format of: \verb|[Impression]| Real \verb|[Reasons]| ...  \verb|[Reflection]| ... \verb|[Answer]| Real.
           \item \textbf{Style}: Your \verb|[Reasons]| and \verb|[Reflection]| must be logically clear and well-argued. Its style and level of detail must strictly follow the example provided by the user.
           \item \textbf{Objective Judgment}: You should independently produce \verb|[Reasons]| and \verb|[Reflection]|. You are not allowed to make mistakes in \verb|[Reasons]| and correct them in \verb|[Reflection]| on purpose. However, in \verb|[Reasons]|, you can leave some details that could be ignored by humans at the first glance. You should only write the most obvious reasons in \verb|[Reasons]|.
       \end{itemize}
       
        ---

        \# Execute Task

        \#\# Example
        
        \verb|[Impression]| Real

        \verb|[Reasons]|
        
        - \textbf{Observation}: The image shows a female model wearing a khaki casual sportswear set, posing in a semi-squatting position. She is wearing black sunglasses, has her hair in a neat high ponytail, and there is garbled text or something similar on her chest.

        - \textbf{Key Areas Evaluation}:
        
        I will evaluate this image's performance on key areas from both positive and negative aspects to maintain an objective analysis.

        ... (see part 2)
\end{tcolorbox}
\caption{Annotation prompt of strategy 2 when the ground-truth label of the sample is ``Real" and the initial guess is controlled to be ``Real" (part 1).}
\label{list:strategy2-3-1}
\end{figure*}

\begin{figure*}[ht]
\centering
\begin{tcolorbox}[listing only, 
                  listing options={basicstyle=\ttfamily\footnotesize},
                  colback=gray!10, 
                  title=Strategy-2-Real-Real-Part-2]
        ... (see part 1)
        \begin{itemize}
            \item Lighting and Geometry: The geometry of the objects in the photo does not have problem. 
            \item Structure and Physics: Although the model's pose is dynamic, it fully conforms to human biomechanics, appearing natural and stable. The wrinkles on the clothing are formed by the bending of the body, are distributed reasonably, and realistically reflect the physical properties of the fabric.
            \item Focus and Depth of Field: The background, consisting of a plain beige wall and a dark floor, has an unnatural cut-out feeling in relation to the model.
            \item Human Pose and Looking: The model's facial features and limbs are well-proportioned, her posture is graceful, and there are no anatomical abnormalities. The details of the hair are clear, with highlights and individual strands well-rendered.
            \item Material and Texture: The model's skin has undergone significant post-processing and is very smooth. This is very similar to the features of AI-generated skin. 
            \item Text and Logo: The texts on the chest of the model are garbled. This could be a sign of AI generation. 
            \item Color: The overall color hue of the image is unified and harmonious. The khaki clothing, beige background, and dark brown floor are well-matched, creating a warm and high-end visual feel. The color reproduction is accurate, with no unnatural color casts.
        \end{itemize}
        
        The image has some suspicious issues (e.g., retouching, unnatural cut-out feeling, garbled texts). However, these issues could also be brought by proper post-processing. For the reasons above, my first impression on the image is ``Real". 

        \verb|[Reflection]|
        
        My first impression on the image is ``Real". Now, I will review the impression as well as its supports, reflectively think, correct the inaccurate parts, , and draw a final conclusion. 

        - \textbf{Observation}:
            - In my initial observation, I forgot to describe the lightings in the photo. The lighting comes from the left-front side of the model. The light is soft and natural. 
            - After careful reviewing, I find the texts on the chest of the model are not garbled -- they are actually cursive scripts reading "Ambie". This might be the brand name of the sportswear. 

        - \textbf{Key Areas Evaluation}:
        \begin{itemize}
            \item Lighting and Geometry: After updating my observation, I should update the corresponding dimension. The lighting is handled very professionally. A single primary light source casts consistent, soft-edged shadows on the model and the ground. The shadow cast on the ground by the squatting model is particularly natural in its shape and density, firmly ``grounding" the figure in the space and completely avoiding the feel of a generated image.
            \item Focus and Depth of Field: At the first glance, I notice the unnatural cut-out feeling between the model and the background. I need to discriminate the feeling of ``concise" and that of ``fake": according to my updated evaluation on ``Lighting and Geometry", the natural shadow cast on the ground by the model proves that none of the model and the background is generated. 
            \item Human Pose and Looking: At the first, I think there is retouching on the model. I need to differentiate the ``retouching" and ``AI generation": in the image, although the skin is smooth, it still has muscle texture and visible (although not very obvious) pores. Therefore, my first impression on this dimension is correct. 
            \item Text and Logo: After updating my observation, the texts on the model's chest are cursive scripts instead of garbled texts. This dimension is no longer a support for ``fake". 
        \end{itemize}

        Based on the analysis above, although there are some mistakes and missing details in my first analysis, the major points supporting the first impression still exist. 
        
        To conclude, I think this is a real image. 

        \verb|[Answer]| Real

        \#\# Now write your report: 
\end{tcolorbox}
\caption{Annotation prompt of strategy 2 when the ground-truth label of the sample is ``Real" and the initial guess is controlled to be ``Real" (part 2).}
\label{list:strategy2-3-2}
\end{figure*}

\begin{figure*}[ht]
\centering
\begin{tcolorbox}[listing only, 
                  listing options={basicstyle=\ttfamily\footnotesize},
                  colback=gray!10, 
                  title=Strategy-2-Fake-Real-Part-1]
        \verb|{Instruction}|
        
        \# Task
        
        \# Context
        
        You are viewing a suspected AI-generated image two times. For the first time, you give your fast impression on the image at the first glance as well as your reasons. For the second time, you review your first impression and reasons, reflectively think, correct the potential errors, and draw a final conclusion. 
        
        Now you have already finished your first view. You have known that:
        
        - \textbf{Input image}: \verb|<image>|
        
        - \textbf{The ground truth label for this image is}: Real
        
        - \textbf{Your impression}: Fake

        Your task is to write a complete analysis report for this image involving in-depth and reflective thinking. 

        \#\# Task instruction
        
        You should:
        
        1. Based on your first impression, write the reasons supporting the impression. The reasons should involve the two step below sequentially:
        \begin{itemize}
            \item Observation: Captioning the image content, including the subject and object in the photo, the background, the lighting, texts and any unnatural details. 
            \item Key Areas Evaluation: Analyze the image from each dimension from the key areas suggested by the user. In the end, correlated the reasons with your impression. 
        \end{itemize}
        
        2. Reflective review your analysis above. Identify and correct the errors in \textbf{observation} and \textbf{Key Areas Evaluation}, respectively. In the end, conclude with ``I think this is a real image.".  

       \textbf{ Please strictly follow the output format below:}
       \begin{itemize}
           \item \textbf{Format}: Strictly output in the format of: \verb|[Impression]| Fake \verb|[Reasons]| ...  \verb|[Reflection]| ... \verb|[Answer]| Real.
           \item \textbf{Style}: Your \verb|[Reasons]| and \verb|[Reflection]| must be logically clear and well-argued. Its style and level of detail must strictly follow the example provided by the user.
           \item \textbf{Objective Judgment}: You should independently produce \verb|[Reasons]| and \verb|[Reflection]|. You are not allowed to make mistakes in \verb|[Reasons]| and correct them in \verb|[Reflection]| on purpose. However, in \verb|[Reasons]|, you can leave some details that could be ignored by humans at the first glance. You should only write the most obvious reasons in \verb|[Reasons]|.
       \end{itemize}
       
        ---

        \# Execute Task

        \#\# Example
        
        \verb|[Impression]| Fake

        \verb|[Reasons]|
        
        - \textbf{Observation}: The image shows a female model wearing a khaki casual sportswear set, posing in a semi-squatting position. She is wearing black sunglasses, has her hair in a neat high ponytail, and there is garbled text or something similar on her chest.

        - \textbf{Key Areas Evaluation}:
        
        I will evaluate this image's performance on key areas from both positive and negative aspects to maintain an objective analysis.

        ... (see part 2)
\end{tcolorbox}
\caption{Annotation prompt of strategy 2 when the ground-truth label of the sample is ``Real" and the initial guess is controlled to be ``Fake" (part 1).}
\label{list:strategy2-4-1}
\end{figure*}

\begin{figure*}[ht]
\centering
\begin{tcolorbox}[listing only, 
                  listing options={basicstyle=\ttfamily\footnotesize},
                  colback=gray!10, 
                  title=Strategy-2-Fake-Real-Part-2]
        ... (see part 1)
        \begin{itemize}
            \item Lighting and Geometry: The geometry of the objects in the photo does not have problem. 
            \item Structure and Physics: Although the model's pose is dynamic, it fully conforms to human biomechanics, appearing natural and stable. The wrinkles on the clothing are formed by the bending of the body, are distributed reasonably, and realistically reflect the physical properties of the fabric.
            \item Focus and Depth of Field: The background, consisting of a plain beige wall and a dark floor, has an unnatural cut-out feeling in relation to the model.
            \item Human Pose and Looking: The model's facial features and limbs are well-proportioned, her posture is graceful, and there are no anatomical abnormalities. The details of the hair are clear, with highlights and individual strands well-rendered.
            \item Material and Texture: The model's skin has undergone significant post-processing and is very smooth. This is very similar to the features of AI-generated skin. 
            \item Text and Logo: The texts on the chest of the model are garbled. This could be a sign of AI generation. 
            \item Color: The overall color hue of the image is unified and harmonious. The khaki clothing, beige background, and dark brown floor are well-matched, creating a warm and high-end visual feel. The color reproduction is accurate, with no unnatural color casts.
        \end{itemize}
        
        The image has some suspicious issues (e.g., retouching, unnatural cut-out feeling, garbled texts). For the reasons above, my first impression on the image is ``Fake".

        \verb|[Reflection]|
        
        My first impression on the image is ``Fake". Now, I will review the impression as well as its supports, reflectively think, correct the inaccurate parts, , and draw a final conclusion. 

        - \textbf{Observation}:
            - In my initial observation, I forgot to describe the lightings in the photo. The lighting comes from the left-front side of the model. The light is soft and natural. 
            - After careful reviewing, I find the texts on the chest of the model are not garbled -- they are actually cursive scripts reading "Ambie". This might be the brand name of the sportswear. 

        - \textbf{Key Areas Evaluation}:
        \begin{itemize}
            \item Lighting and Geometry: After updating my observation, I should update the corresponding dimension. The lighting is handled very professionally. A single primary light source casts consistent, soft-edged shadows on the model and the ground. The shadow cast on the ground by the squatting model is particularly natural in its shape and density, firmly ``grounding" the figure in the space and completely avoiding the feel of a generated image.
            \item Focus and Depth of Field: At the first glance, I notice the unnatural cut-out feeling between the model and the background. I need to discriminate the feeling of ``concise" and that of ``fake": according to my updated evaluation on ``Lighting and Geometry", the natural shadow cast on the ground by the model proves that none of the model and the background is generated. 
            \item Human Pose and Looking: At the first, I think there is retouching on the model. I need to differentiate the ``retouching" and ``AI generation": in the image, although the skin is smooth, it still has muscle texture and visible (although not very obvious) pores. Therefore, my first impression on this dimension is correct. 
            \item Text and Logo: After updating my observation, the texts on the model's chest are cursive scripts instead of garbled texts. This dimension is no longer a support for ``fake". 
        \end{itemize}

        Based on the analysis above, although some of the points in my first analysis are correct, I misunderstand some image details and the major points supporting the first impression do not exist any more. 
        
        To conclude, I think this is a real image. 

        \verb|[Answer]| Real

        \#\# Now write your report: 
\end{tcolorbox}
\caption{Annotation prompt of strategy 2 when the ground-truth label of the sample is ``Real" and the initial guess is controlled to be ``Fake" (part 2).}
\label{list:strategy2-4-2}
\end{figure*}

\end{document}